\documentclass[journal]{IEEEtran}

\usepackage{amsmath}
\usepackage{color}
\usepackage{graphicx}
\usepackage{amssymb}
\usepackage{lipsum}
\usepackage{widetext}
\usepackage{subfigure}
\usepackage{booktabs}
\usepackage{tabularx}
\usepackage{multicol}
\usepackage{comment}

\providecommand{\ve}[1]{\boldsymbol{\mathrm{#1}}}
\providecommand{\mat}[1]{\boldsymbol{\mathbf{#1}}}

\providecommand{\y}{\boldsymbol{y}}

\newtheorem{theorem}{Theorem}

\newcommand{\Rmnum}[1]{\uppercase\expandafter{\romannumeral #1}}

\usepackage{cite}

\ifCLASSINFOpdf
\else
\fi

\usepackage{url}

\begin{document}

\graphicspath{{.}}
\title{Robust Visual Tracking using Multi-Frame Multi-Feature Joint Modeling}

\author{Peng~Zhang$^\ast$,
         Shujian~Yu$^\ast$,~\IEEEmembership{Student Member,~IEEE,}
         Jiamiao~Xu,
         Xinge~You$^\dagger$,~\IEEEmembership{Senior Member,~IEEE,}\\
         Xiubao~Jiang,
         Xiao-Yuan~Jing,
        and~Dacheng~Tao,~\IEEEmembership{Fellow,~IEEE.}% <-this % stops a space
\thanks{$^\ast$The first two authors contributed equally to this work and should be regarded as co-first authors.}
\thanks{$^\dagger$Corresponding author.}
\thanks{P. Zhang, J. Xu, X. You and X. Jiang are with the School of Electronic Information and Communications, Huazhong University of Science and Technology, Wuhan, Hubei, China. e-mail: youxg@mail.hust.edu.cn.}% <-this % stops a space
\thanks{S. Yu is with the Department of Electrical and Computer Engineering, University of Florida, Gainesville, FL 32611, USA. e-mail: yusjlcy9011@ufl.edu.}% <-this % stops a space
\thanks{X.-Y. Jing is with the State Key Laboratory of Software Engineering, School of Computer, Wuhan University, Wuhan 430072, China and also with the School of Automation, Nanjing University of Posts and Telecommunications, Nanjing 210023, China. e-mail: jingxy$\_$2000@126.com.}
%\thanks{Manuscript received April 19, 2005; revised September 17, 2014.}}
\thanks{D. Tao is with the UBTech Sydney Artificial Intelligence Institute and the School of Information Technologies, in the Faculty of Engineering and Information Technologies at The University of Sydney, Darlington, NSW 2008, Australia. e-mail: dacheng.tao@sydney.edu.au.}}

\maketitle

\begin{abstract}
It remains a huge challenge to design effective and efficient trackers under complex scenarios, including occlusions, illumination changes and pose variations. To cope with this problem, a promising solution is to integrate the temporal consistency across consecutive frames and multiple feature cues in a unified model. Motivated by this idea, we propose a novel correlation filter-based tracker in this work, in which the temporal relatedness is reconciled under a multi-task learning framework and the multiple feature cues are modeled using a multi-view learning approach. We demonstrate the resulting regression model can be efficiently learned by exploiting the structure of $\emph{blockwise diagonal matrix}$. A fast $\emph{blockwise diagonal matrix}$ inversion algorithm is developed thereafter for efficient online tracking. Meanwhile, we incorporate an adaptive scale estimation mechanism to strengthen the stability of scale variation tracking. We implement our tracker using two types of features and test it on two benchmark datasets. Experimental results demonstrate the superiority of our proposed approach when compared with other state-of-the-art trackers. MATLAB code is available from our
project homepage \url{http://bmal.hust.edu.cn/project/KMF2JMTtracking.html}.

%We extend $(\text{MF})^2\text{JMT}$ to its kernelized version, namely $\text{K}(\text{MF})^2\text{JMT}$, and show that the main computational cost of $\text{K}(\text{MF})^2\text{JMT}$ in the training phase comes from the inversion of a blockwise diagonal matrix when using the dense sampling strategy.
\end{abstract}


% \fi
\IEEEpeerreviewmaketitle

\section{Introduction}

%\IEEEPARstart{T}{racking}

\IEEEPARstart{V}{isual} tracking is one of the most important components in computer vision system. It has been widely used in visual surveillance, human computer interaction, and robotics \cite{smeulders2014visual,ding2016severely}. Given an annotation of the object (bounding box) in the first frame, the task of visual tracking is to estimate the target locations and scales in subsequent video frames. Though much progress has been made in recent years, robust visual tracking, which can reconcile different varying circumstances, still remains a challenging problem \cite{smeulders2014visual}. On the one hand, it is expected that the designed trackers can compensate for large appearance changes caused by illuminations, occlusions, etc. On the other hand, the real-time requirement in real applications impedes the usage of overcomplicated models.

Briefly speaking, there are two major categories of modern trackers: generative trackers and discriminative trackers \cite{li2013survey,smeulders2014visual}. Generative trackers typically assume a generative process of the target appearance and search for the regions most similar to the target model, while discriminative trackers usually train a classifier to distinguish the target from the background. Among discriminative trackers, correlation filter-based trackers (CFTs) drawn an increasing number of attentions since the development of Kernel Correlation Filter (KCF) tracker \cite{henriques2015high}. As has been demonstrated, KCF can achieve impressive performance on accuracy, robustness and speed on both the Online Tracking Benchmark (OTB)~\cite{wu2015object} and the Visual Object Tracking (VOT) challenges~\cite{kristan2016novel}.

Despite the overwhelming evidence of success achieved by CFTs, two observations prompt us to come up with our tracking approach. First, almost all the CFTs ignore the temporal consistency or invariance among consecutive frames, which has been demonstrated to be effective in augmenting tracking performance \cite{yao2012robust,wang2016visual}. Second, there is still a lack of theoretical sound yet computational efficient model to integrate multiple feature cues. Admittedly, integrating different channel features is not new under the CFTs umbrella. However, previous work either straightforwardly concatenating various feature vectors \cite{li2014scale,zhang2014fast} (i.e., assuming mutual independence of feature channels) or inheriting high computational burden which severely compromises the tracking speed \cite{tang2015multi,li2016multi}.

In this paper, to circumvent these two drawbacks simultaneously, we embark on the basic KCF tracker and present a multi-frame multi-feature joint modeling tracker ($(\text{MF})^2\text{JMT}$) to strike a good trade-off between robustness and speed. In $(\text{MF})^2\text{JMT}$, the interdependencies between different feature cues are modeled using a multi-view learning (MVL) approach to enhance the discriminative power of target appearance undergoing various changes. Specifically, we use the view consistency principle to regularize the objective function learned from each view agree on the labels of most training samples \cite{blum1998combining}. On the other hand, the temporal consistency is exploited under a multi-task learning (MTL) framework \cite{evgeniou2004regularized}, i.e., we model $M$ ($M\geq2$) consecutive frames simultaneously by constraining the learned objectives from each frame close to their mean. We extend $(\text{MF})^2\text{JMT}$ to its kernelized version (i.e., $\text{K}(\text{MF})^2\text{JMT}$) and develop a fast $\emph{blockwise diagonal matrix}$ inversion algorithm to accelerate model training. Finally, we adopt a two-stage filtering pipeline \cite{li2014scale,danelljan2014accurate,hong2015multi} to cope with the problem of scale variations.

%We extend $(\text{MF})^2\text{JMT}$ to its kernelized version (i.e., $\text{K}(\text{MF})^2\text{JMT}$) and demonstrate the resulting regression model can be efficiently learned by exploiting the structure of $\emph{blockwise diagonal matrix}$. A fast $\emph{blockwise diagonal matrix}$ inversion algorithm is developed thereafter for efficient online tracking. Finally, we adopt a two-stage filtering pipeline \cite{li2014scale,danelljan2014accurate,hong2015multi} to cope with the problem of scale variations, where the first filter (the basic $\text{K}(\text{MF})^2\text{JMT}$) is used for translation estimation and the second filter (the Discriminative Scale Space Correlation Filter (DSSCT) \cite{danelljan2014accurate}) serves as a scale estimator.

%To summarize, the main contributions of our work are threefold. First, a novel tracker, which can joint model temporal consistency and multiple feature cues, is present for robust visual tracking. Second, a fast $\emph{blockwise diagonal matrix}$ inversion algorithm is developed to speed up training and detection, and the incorporation of adaptive scale estimation mechanism can significantly enhance the stability of scale variation tracking. Third, extensive experiments are conducted to compare with other CFTs published in recent years. These results reveal the underline clues on the importance of ``intelligent" feature integration and temporal modeling, and also illustrate the future directions for the design of modern discriminative trackers.

To summarize, the main contributions of our work are twofold. First, a novel tracker, which can integrate multiple feature cues and temporal consistency in a unified model, is developed for robust visual tracking. Specifically, instead of simply concatenating multiple features into a single vector, we demonstrate how to reorganize these features by taking into consideration their intercorrelations. Moreover, we also present an advanced way to reconcile temporal relatedness amongst multiple consecutive frames, rather than naively using a forgetting factor~\cite{zhang2014fast,zhou2017multi}. Second, a fast $\emph{blockwise diagonal matrix}$ inversion algorithm is developed to speed up training and detection. Experiments against state-of-the-art trackers reveal the underline clues on the importance of ``intelligent" feature integration and temporal modeling, and also illustrate future directions for the design of modern discriminative trackers.

The rest of this paper is organized as follows. In section II, we introduce the background knowledge. Then, in section III, we discuss the detailed methodology of $(\text{MF})^2\text{JMT}$ and extend it under kernel setting. A fast algorithm for $\emph{blockwise diagonal matrix}$ inversion is also developed to speed up training and detection. Following this, an adaptive scale estimation mechanism is incorporated into our tracker in section IV. The performance of our proposed approach against other state-of-the-art trackers is evaluated in section V. This paper is concluded in section VI.

$Notation$: Scalars are denoted by lowercase letters (e.g., $x$), vectors appear as lowercase boldface letters (e.g., $\ve{x}$), and matrices are indicated by uppercase letters (e.g., $X$ or $\ve{X}$). The $(i,j)$-th element of $X$ (or $\ve{X}$) is represented by $X_{ij}$ (or $\ve{X}_{ij}$), while $[\cdot]^T$ denotes transpose and $[\cdot]^H$ denotes Hermitian transpose. If $X$ (or $\ve{X}$) is a square matrix, then $X^{-1}$ (or $\ve{X}^{-1}$) denotes its inverse. $\ve{I}$ stands for the identity matrix with compatible dimensions, $diag(\ve{x})$ denotes a square diagonal matrix with the elements of vector $\ve{x}$ on the main diagonal. The $i$-th row of a matrix $X$ (or $\ve{X}$) is declared by the row vector $\ve{x}^i$, while the $j$-th column is indicated with the column vector $\ve{x}_j$. If $|\cdot|$ denotes the absolute value operator, then, for $\ve{x}\in\mathbb{R}^n$, the $\ell_1$-norm and $\ell_2$-norm of $\ve{x}$ are defined as $\|\ve{x}\|_1\triangleq\sum_{i=1}^n|x_i|$ and $\|\ve{x}\|_2\triangleq\sqrt{\sum_{i=1}^nx_i^2}$, respectively.

\section{Related work}
%An extensive review on visual tracking is beyond the scope of this paper. We refer readers to some recently published surveys \cite{li2013survey,smeulders2014visual} for more details about existing trackers, and an extensive survey on multi-view learning (MVL) and multi-task learning (MTL) can be found in \cite{xu2013survey} and \cite{zhang2017survey}.

In this section, we briefly review previous work of most relevance to our approach, including popular conventional trackers, the basic KCF tracker and its extensions. %We also give a general formulation of blockwise circular matrix.

\subsection{Popular conventional trackers}
Success in visual tracking relies heavily on how discriminative and robust the representation of target appearance is against varying circumstances \cite{fan2017robust,zhou2017locality}. This is especially important for discriminative trackers, in which a binary classifier is required to distinguish target from background. Numerous types of visual features have been successfully applied for discriminative trackers in the last decades, including color histograms \cite{collins2005online,he2017robust}, texture features \cite{nguyen2004tracking}, Haar-like features \cite{grabner2006line}, etc. Unfortunately, none of them can handle all kinds of varying circumstances individually and the discriminant capability of a unique type of feature is not stable across the video sequence \cite{collins2005online}. As a result, it becomes prevalent to take advantage of multiple features (i.e., multi-view representations) to enable more robust tracking. For example, \cite{avidan2007ensemble,tang2007co} used the AdaBoost to combine an ensemble of weak classifiers to form a powerful classifier, where each weak classifier is trained online on a different training set using pixel colors and a local orientation histogram. Note that, several generative trackers also have improved performance by incorporating multi-view representations. \cite{mei2015robust} employed a group sparsity technique to integrate color histograms, intensity, histograms of oriented gradients (HOGs) \cite{dalal2005histograms} and local binary patterns (LBPs) \cite{ojala2002multiresolution} via requiring these features to share the same subset of the templates, whereas \cite{yoon2016interacting} proposed a probabilistic approach to integrate HOGs, intensity and Haar-like features for robust tracking. These trackers perform well normally, but they are far from satisfactory when being tested on challenging videos.

%\cite{kwon2010visual} developed a visual tracking decomposition (VTD) system that consists of a fixed number of basic trackers based on multiple types of features (including hue, saturation, intensity, edge, etc.) to deal with various scenarios. \cite{du2008probabilistic} proposed a probabilistic approach to integrate four features, i.e., color, edges, motion and contours, for robust tracking. \cite{hong2013tracking} employed a group sparsity technique to integrate different features via requiring these features share the same subset of the templates. Similar technique has been applied in \cite{hu2015single}, where the templates for sparse representation extend to pixel values, textures and edges. \cite{moreno2008dependent} proposed a probabilistic framework allowing the integration of multiple features for tracking by considering cue dependencies.

\subsection{The basic KCF tracker and its extensions}

Much like other discriminative trackers, the KCF tracker needs a set of training examples to learn a classifier. The key idea of KCF tracker is that the augmentation of negative samples is employed to enhance the discriminative ability of the tracking-by-detection scheme while exploring the structure of the circulant matrix to speed up training and detection.

%\subsection{Extensions of KCF tracker}

Following the basic KCF tracker, numerous extensions %\footnote{\textcolor{red}{Note that, influenced by the successful application of correlation filters on visual tracking, there have been developed several thermal infrared trackers based on correlation filters (e.g.,~\cite{gundogdu2015comparison,liu2017deep}). This work only focuses on visual tracking on colored video sequences. We leave extension and investigation on thermal infrared tracking or detection~\cite{reddy2013improved,yadav2016combined} as future work.}}
have been conducted to boost its performance, which generally fall into two categories: application of improved features and conceptual improvements in filter learning \cite{lukevzivc2016discriminative}. The first category lies in designing more discriminative features to deal with challenging environments \cite{ma2015learning,hu2017manifold} or straightforwardly concatenating multiple feature cues into a single feature vector to boost representation power \cite{li2014scale,hu2017object}. A recent trend is to integrate features extracted from convolutional neural networks (CNNs) trained on large image datasets to replace the traditional hand-crafted features \cite{ma2015hierarchical,danelljan2015convolutional}. The design or integration of more powerful features often suffers from high computational burden, whereas blindly concatenating multiple features assumes the mutual independence of these features and neglects their interrelationships. Therefore, it is desirable to have a more ``intelligent" manner to reorganize these features such that their interrelationships are fully exploited.

Conceptually, the theoretical extension on the filter learning also drawn lots of attentions. Early work focused on accurate and efficient scale estimation as the basic KCF assumes a fixed target size~\cite{danelljan2014accurate,tang2015multi}.
%\cite{danelljan2014accurate} proposed a scale pyramid representation by learning discriminative correlation filters to adaptive estimate scale. This method was further improved by \cite{tang2015multi} via applying a fast feature pyramid.
The most recent work concentrated on developing more advanced convolutional operators. For example, \cite{danelljan2016beyond} employed an interpolation model to train convolution filters in continuous spatial domain. The so-called Continuous Convolution Operator Tracker (C-COT) is further improved in \cite{danelljan2016eco}, in which the authors introduce factorized convolution operators to drastically reduce the number of parameters C-COT as well as the number of filters. %We refer interested readers to \cite{chen2015experimental} for an early experimental survey on KCF trackers and its various extensions.

Apart from these two extensions, efforts have been made to embed the conventional tracking strategies (like part-based tracking \cite{liu2016part}) on KCF framework. For example, \cite{li2015reliable} developed a probabilistic tracking reliability metric to measure how reliable a patch can be tracked. On the other hand, \cite{ma2015long} employed an online random fern classifier as a re-detection component for long-term tracking, whereas \cite{hong2015multi} presented a biology-inspired framework where short-term processing and long-term processing are cooperated with each other under a correlation filter framework. Finally, it is worth noting that, with the rapid development of correlation filters on visual tracking, several correlation filter-based thermal infrared trackers (e.g.,~\cite{gundogdu2015comparison,liu2017deep,yadav2016combined}) have been developed in recent years. This work only focuses on visual tracking on color video sequences. We leave extension to thermal infrared video sequences as future work.

%Despite the overwhelming evidence of success achieved by KCF and its extensions, two observations impede the potentials of these trackers on real applications and prompt us to propose our approach. First, all these methods ignore the temporal consistency or invariance information among consecutive frames, which may benefit the tracking performance \cite{wang2016visual,ma2015learning,zhao2015metric,hare2012efficient}. Second, although prevalent works already demonstrated the superiority of integrating multiple feature cues to augment appearance representation power, they either straightforwardly concatenating different feature vectors into a single one which ignores their heterogeneity \cite{li2014scale} or inherit high computational complexity which severely compromises the real time capability \cite{tang2015multi,li2016multi,zhang2017robust}. In virtue of this, we develop the multi-frame multi-feature joint modeling tracker ($(\text{MF})^2\text{JMT}$), which will be discussed in the next section.

\vspace{-0.3cm}
\section{The Multi-Frame Multi-Feature Joint Modeling Tracker ($(\text{MF})^2\text{JMT}$)}

%In this section, we present our developed Multi-Frame Multi-Feature Joint Modeling Tracker ($(\text{MF})^2\text{JMT}$). We start from a formal formulation of $(\text{MF})^2\text{JMT}$ in section \ref{IV-A} and give its primal solution in section \ref{IV-B}. Following this, we extend $(\text{MF})^2\text{JMT}$  to its kernelized version and provide a fast implementation by exploiting the structure of blockwise diagonal matrix.

The $(\text{MF})^2\text{JMT}$ made two strategic extensions to the basic KCF tracker. Our idea is to integrate the temporal information and multiple feature cues in a unified model, thus providing a theoretical sound and computational-efficient solution for robust visual tracking. To this end, instead of simply concatenating different features, the $(\text{MF})^2\text{JMT}$ integrates multiple feature cues using a MVL approach to better exploit their interrelationships, thus forming a more informative representation to target appearance. Moreover, the temporal consistency is taken into account under a MTL framework. Specifically, different from prevalent CFTs that learn filter taps using a ridge regression function which only makes use of template (i.e. circulant matrix $\ve{X}$) from the current frame, we show that it is possible to use examples from $M$ ($M\geq2$) consecutive frames to learn the filter taps very efficiently by exploiting the structure of $\emph{blockwise diagonal matrix}$.

Before our work, there are two ways attempting to incorporate temporal consistency into KCF framework. The Spatio-Temporal Context (STC) tracker~\cite{zhang2014fast} and its extensions (e.g.,~\cite{zhou2017multi}) formulate the spatial relationships between the target and its surrounding dense contexts in a Bayesian framework and then use a temporal filtering procedure together with a forgetting factor to update the spatio-temporal model. Despite its simplicity, the tracking accuracy of STC tracker is poor compared with other state-of-the-art KCF trackers (see Section~\ref{experiments_section3}). On the other hand, another trend is to learn a temporally invariant feature representation trained on natural video repository for visual tracking~\cite{ma2015learning}. Although the learned feature can accommodate partial occlusion or slight illumination variation, it cannot handle the scenarios where there are large appearance changes. Moreover, the effectiveness of trained features depends largely on the selected repository which suffers from limited generalization capability. Different from these two kinds of methods, we explicitly model multiple consecutive frames in a joint cost function to circumvent abrupt drift in a single frame and to reconcile temporal relatedness amongst frames in a short period of time. Experiments demonstrate the superiority of our method.

\vspace{-0.3cm}
\subsection{Formulation of $(\text{MF})^2\text{JMT}$} \label{IV-A}
We start from the formulation of the basic $(\text{MF})^2\text{JMT}$. %We will show in Section \ref{IV-C} how the corresponding objective can be efficiently solved using the dense sampling strategy under kernel setting.
Given $M$ training frames, we assume the number of candidate patches in the $t$-th ($t=1,2,\cdots,M$) frame is $n$. Suppose the dimensions for the first and second types of features are $p$ and $q$ respectively\footnote{This paper only considers two types of features, but the developed objective function (\ref{MTMVmodel2}) and associated solution can be straightforwardly extended to three or more feature cues.}, we denote $X_t\in\mathbb{R}^{n\times p}$ ($Z_t\in\mathbb{R}^{n\times q}$) the matrix consists of the first (second) type of feature in the $t$-th frame (each row represents one sample). Also, let $\ve{y}_t\in\mathbb{R}^{n\times1}$ represent sample labels. Then, the objective of $(\text{MF})^2\text{JMT}$ can thus be formulated as:\\
\vspace{-0.3cm}
\begin{eqnarray}
\min_{\ve{w}_0,\ve{p}_t,\ve{v}_0,\ve{q}_t}\ \mathcal{J}&=& \sum_{t=1}^M\Big(\left\|\ve{y}_t-X_t\ve{w}_t\right\|_2^2+\lambda_1\left\|\ve{y}_t-Z_t\ve{v}_t\right\|_2^2\nonumber\\
&&+\lambda_2\|X_t\ve{w}_t-Z_t\ve{v}_t\|_2^2\Big)+\frac{\gamma_1}{M}\sum_{t=1}^M\left\|\ve{p}_t\right\|_2^2\nonumber\\
&&+\gamma_2\|\ve{w}_0\|_2^2+\frac{\eta_1}{M}\sum_{t=1}^M\left\|\ve{q}_t\right\|_2^2+\eta_2\|\ve{v}_0\|_2^2,\nonumber\\
s.t. && \ve{w}_t=\ve{w}_0+\ve{p}_t\nonumber\\
&& \ve{v}_t=\ve{v}_0+\ve{q}_t
\label{MTMVmodel2}
\end{eqnarray}
where $\lambda_1,\lambda_2,\gamma_1,\gamma_2,\eta_1,\eta_2$ are the non-negative regularization parameters controlling model complexity. $\ve{w}_0$ is the regression coefficients shared by $M$ frames for the first type of feature and $\ve{p}_t$ denotes the deviation term from $\ve{w}_0$ in the $t$-th frame. The same definition goes for $\ve{v}_0$ and $\ve{q}_t$ for the second type of feature.

The problem (\ref{MTMVmodel2}) contains three different items with distinct objectives, namely the MTL item, the MVL item and the regularization item. The MTL item, i.e., $\sum_{t=1}^M\left\|\y_t-X_t\ve{w}_t\right\|_2^2+\frac{\gamma_1}{M}\sum_{t=1}^M\left\|\ve{p}_t\right\|_2^2$ (or $\sum_{t=1}^M\left\|\y_t-Z_t\ve{v}_t\right\|_2^2+\frac{\eta_1}{M}\sum_{t=1}^M\left\|\ve{q}_t\right\|_2^2$), is analogous to the formulation of regularized multi-task learning (RMTL) \cite{evgeniou2004regularized}, as it encourages $\ve{w}_t$ (or $\ve{v}_t$) close to the mean value $\ve{w}_0$ (or $\ve{v}_0$) with a small deviation $\ve{p}_t$ (or $\ve{q}_t$). The MVL item $\left\|\ve{y}_t-X_t\ve{w}_t\right\|_2^2+\lambda_1\left\|\ve{y}_t-Z_t\ve{v}_t\right\|_2^2+\lambda_2\|X_t\ve{w}_t-Z_t\ve{v}_t\|_2^2$ employs the view consistency principle that widely exists in MVL approaches (e.g., \cite{liu2013multi}) to constrain the objectives learned from two views agree on most training samples. Finally, the regularization term $\|\ve{w}_0\|_2^2$ (or $\|\ve{v}_0\|_2^2$) serves to prevent the ill-posed solution and enhance the robustness of selected features to noises or outliers.

%\subsection{Generalization and special cases}
%Before presenting the optimization method of (8), we would like to have a brief discussion about the proposed problem (6) in this section. The proposed optimization problem (6) can be generalized as

\vspace{-0.3cm}
\subsection{Solution to $(\text{MF})^2\text{JMT}$} \label{IV-B}

%The objective function $\mathcal{J}$ of $(\text{MF})^2\text{JMT}$ is quadratic and convex w.r.t. $\ve{w}_0,\ve{p}_t,\ve{v}_0$ and $\ve{q}_t$. Thus, a straightforward solution is to calculate the derivative of $\mathcal{J}$ with respect to all these variables and then solve a linear equation array. However, it is difficult to reorganize linear equation array into standard form herein.

Minimization of $\mathcal{J}$ has a closed-form solution by equating the gradients of (\ref{MTMVmodel2}) w.r.t. $\ve{w}_0,\ve{p}_t,\ve{v}_0$ and $\ve{q}_t$ to zero. Albeit its simplicity, reorganizing linear equation array into standard form is intractable and computational expensive herein. As an alternative, we present an equivalent yet simpler form of (\ref{MTMVmodel2}), which can be solved with matrix inversion in one step.

Denote
\begin{eqnarray}
\mu_1 &=& \frac{M\gamma_2}{\gamma_1},\\
\mu_2 &=& \frac{M\eta_2}{\eta_1},\\
\overline{X}_t &=& \Big(\frac{X_t}{\sqrt{\mu_1}},
\underbrace{\ve{0},\cdots,\ve{0}}_{t-1},X_t,
\underbrace{\ve{0},\cdots,\ve{0}}_{M-t}\Big),\label{Eq9}\\
\overline{Z}_t &=& \Big(\frac{Z_t}{\sqrt{\mu_2}},\underbrace{\ve{0},\cdots,\ve{0}}_{t-1},Z_t,
\underbrace{\ve{0},\cdots,\ve{0}}_{M-t}\Big),
\end{eqnarray}
\begin{eqnarray}
\ve{w} &=& \left(\begin{array}{c}\sqrt{\mu_1}\ve{w}_0\\ \ve{p}_1 \\ \vdots \\ \ve{p}_M\end{array}\right),\quad
\ve{v} = \left(\begin{array}{c}\sqrt{\mu_2}\ve{v}_0\\ \ve{q}_1 \\ \vdots \\ \ve{q}_M\end{array}\right),
\end{eqnarray}
where $\ve{0}$ denotes a zero matrix of the same size as $X_t$ in $\overline{X}_t$ (or $Z_t$ in $\overline{Z}_t$), we have:\\
\vspace{-0.3cm}
\begin{eqnarray}
\overline{X}_t\ve{w} &=& X_t\ve{w}_0+X_t\ve{p}_t = X_t\ve{w}_t,\label{multprop1}\\
\overline{Z}_t\ve{v} &=& Z_t\ve{v}_0+Z_t\ve{q}_t = Z_t\ve{v}_t,\\
\|\ve{w}\|_2^2 &=& \sum_{t=1}^M\|\ve{p}_t\|_2^2+\mu_1\|\ve{w}_0\|_2^2 \nonumber\\
&=& \sum_{t=1}^M\|\ve{p}_t\|_2^2+\frac{M\gamma_2}{\gamma_1}\|\ve{w}_0\|_2^2,\\
\|\ve{v}\|_2^2 &=& \sum_{t=1}^M\|\ve{q}_t\|_2^2+\mu_2\|\ve{v}_0\|_2^2 \nonumber\\
&=&\sum_{t=1}^M\|\ve{q}_t\|_2^2+\frac{M\eta_2}{\eta_1}\|\ve{v}_0\|_2^2.\label{multprop4}
\end{eqnarray}

Substituting (\ref{multprop1})-(\ref{multprop4}) into (\ref{MTMVmodel2}) yields:
%\begin{small}
\begin{eqnarray}
\min_{\ve{w},\ve{v}}\ \mathcal{J} &=& \sum_{t=1}^M\Big(\left\|\ve{y}_t-\overline{X_t}\ve{w}\right\|_2^2+\lambda_1\left\|\y_t-\overline{Z_t}\ve{v}\right\|_2^2\nonumber\\
&+& \lambda_2\|\overline{X_t}\ve{w}-\overline{Z_t}\ve{v}\|_2^2\Big)
+\frac{\gamma_1}{M}\left\|\ve{w}\right\|_2^2
+\frac{\eta_1}{M}\left\|\ve{v}\right\|_2^2.\nonumber\\
\label{MTMVmodel2form1}
\end{eqnarray}
%\end{small}

Denote
\begin{eqnarray}
\ve{y} = \left(\begin{array}{c}
\ve{y}_1\\
\vdots\\
\ve{y}_M
\end{array}
\right),\quad
\ve{X} = \left(\begin{array}{c}
\overline{X}_1\\
\vdots\\
\overline{X}_M
\end{array}
\right),\quad
\ve{Z} = \left(\begin{array}{c}
\overline{Z}_1\\
\vdots\\
\overline{Z}_M
\end{array}
\right),
\label{EqlargeXZ}
\end{eqnarray}
\begin{comment}
we have:
\begin{eqnarray}
\|\ve{y}-\ve{X}\ve{w}\|_2^2 &=& \sum_{t=1}^T\|\ve{y}-\overline{X}_t\ve{w}\|_2^2\label{Eqeqiv21}\\
\|\ve{y}-\ve{Z}\ve{v}\|_2^2 &=& \sum_{t=1}^T\|\ve{y}-\overline{Z}_t\ve{v}\|_2^2\\
\|\ve{X}\ve{w}-\ve{Z}\ve{v}\|_2^2 &=& \sum_{t=1}^T\|\overline{X}_t\ve{w}-\overline{Z}_t\ve{v}\|_2^2.\label{Eqeqiv22}
\end{eqnarray}
\end{comment}
then (\ref{MTMVmodel2form1}) becomes:
%\begin{small}
\begin{eqnarray}
\min_{\ve{w},\ve{v}}\ \mathcal{J} &=& \left\|\ve{y}-\ve{X}\ve{w}\right\|_2^2+\lambda_1\left\|\ve{y}-\ve{Z}\ve{v}\right\|_2^2+
\lambda_2\|\ve{X}\ve{w}-\ve{Z}\ve{v}\|_2^2\nonumber\\
&&+\frac{\gamma_1}{M}\left\|\ve{w}\right\|_2^2
+\frac{\eta_1}{M}\left\|\ve{v}\right\|_2^2.\label{MTMVmodel2form2}
\end{eqnarray}
%\end{small}

Equating the gradients of $\mathcal{J}$ w.r.t. $\ve{w}$ and $\ve{v}$ to zero. With straightforward derivation, we have:
%\begin{eqnarray}
\begin{flalign}
\left(\begin{array}{cc}
(1+\lambda_2)\ve{X}^T\ve{X}+\frac{\gamma_1}{M}\ve{I} & -\lambda_2\ve{X}^T\ve{Z}\\
-\lambda_2\ve{Z}^T\ve{X} & (\lambda_1+\lambda_2)\ve{Z}^T\ve{Z}+\frac{\eta_1}{M}\ve{I}
\end{array}
\right)
\left(\begin{array}{c}
\ve{w}\\
\ve{v}
\end{array}
\right)\nonumber\\=
\left(\begin{array}{c}
\ve{X}^T\ve{y}\\
\lambda_1\ve{Z}^T\ve{y}
\end{array}
\right).
\end{flalign}
%\end{eqnarray}

Denote $\ve{\xi}\doteq (\ve{w},\ve{v})^T$, the solution of $\mathcal{J}$ is given by~\cite{bishop2006pattern}:
\begin{eqnarray}
\ve{\xi} &=& \Big(\ve{U}^T\ve{D}\ve{U}+\ve{R}\Big)^{-1}\ve{U}^T\ve{Y} \nonumber\\
&=& \ve{R}^{-1}\ve{U}^T\Big(\ve{U}\ve{R}^{-1}\ve{U}^T+\ve{D}^{-1}\Big)^{-1}\ve{D}^{-1}\ve{Y}, \label{LinearEqsSolu}
\end{eqnarray}
where
\begin{eqnarray}
\ve{D} = \left(\begin{array}{cc}
(1+\lambda_2)\ve{I} & -\lambda_2\ve{I}\\
-\lambda_2\ve{I} & (\lambda_1+\lambda_2)\ve{I}
\end{array}\right),
\ve{U} =
\left(\begin{array}{cc}
\ve{X} & 0\\
0 & \ve{Z}
\end{array}\right),\nonumber\\
%\ve{\xi}\doteq \left(\begin{array}{c}\ve{w}\\ \ve{v}\end{array}\right).
\ve{R} = \left(\begin{array}{cc}
\frac{\gamma_1}{M}\ve{I} & 0\\
0 & \frac{\eta_1}{M}\ve{I}
\end{array}\right),
\ve{Y} = \left(\begin{array}{c}
\ve{y}\\\lambda_1\ve{y}
\end{array}\right).
%\ve{\xi}\doteq \left(\begin{array}{c}\ve{w}\\ \ve{v}\end{array}\right).
\end{eqnarray}

Having computed $\ve{w}_t$ and $\ve{v}_t$, the responses of candidate samples $\ve{z}$ in the next frame for the trained $(\text{MF})^2\text{JMT}$ model can be computed as:
\begin{eqnarray}
f(\ve{z})&=& \ve{U}^{new}\ve{\xi}\nonumber\\
&=& \ve{U}^{new}\ve{R}^{-1}\ve{U}^T\Big(\ve{U}\ve{R}^{-1}\ve{U}^T+\ve{D}^{-1}\Big)^{-1}\ve{D}^{-1}\ve{Y},\nonumber\\
\label{kpred}
\end{eqnarray}
where
\begin{eqnarray}
\ve{U}^{new} = \left(\begin{array}{cc}
\ve{X}^{new} & 0\\
0 & \ve{Z}^{new}
\end{array}
\right),
\end{eqnarray}
in which $\ve{X}^{new}$ and $\ve{Z}^{new}$ are feature matrices constructed from features in the new frame. Specifically, let $\ve{X}^{new}=(\overline{X}^{new}_1,...,\overline{X}^{new}_M)^T\in\mathbb{R}^{Mn\times(M+1)p}$ consist of the first type of feature, we construct $\overline{X}^{new}_M$ with feature in the new frame as defined in (\ref{Eq9}) and set $\overline{X}^{new}_1,...,\overline{X}^{new}_{M-1}$ as zero matrices to coincide with the size of $\ve{X}$ in (\ref{EqlargeXZ}). In this sense, only $\ve{w}_M$ and $\ve{v}_M$ contribute to $f(\ve{z})$. The same goes for $\ve{Z}^{new}$ and $\overline{Z}^{new}_t$ ($t=1,...,M$).

\subsection{Kernel extension and fast implementation} \label{IV-C}
Although (\ref{LinearEqsSolu}) gives a tractable solution to (\ref{MTMVmodel2}), it contains the inversion of $\ve{U}^T\ve{D}\ve{U}+\ve{R}$ with the computational complexity $\mathcal{O}(n^3)$ when using the well acknowledged Gaussian elimination algorithm \cite{golub2012matrix}. In this section, for a more powerful regression function and a fast implementation, we demonstrate how to incorporate the dense sampling strategy in the basic $(\text{MF})^2\text{JMT}$ under the kernel setting to speed up its tracking and detection. A computational-efficient optimization method is also presented by exploiting the structure of $\emph{blockwise diagonal matrix}$.

%\subsection{Kernel extension of $(\text{MF})^2\text{JMT}$}
%The core task for our tracker depends on the calculation of (\ref{kpred}). A natural way to allow more powerful, non-linear regression functions $f(\ve{z})$ is with the ``kernel trick". The most attractive quality is that the optimization problem is still linear, albeit in a high (or infinite) dimensional space, i.e., Reproducing Kernel Hilbert Space (\emph{RKHS}).

The dense sampling was considered previously to be a drawback for discriminative trackers because of the large number of redundant samples that are required \cite{li2013survey}. However, when these samples are collected and organized properly, they form a circulant matrix that can be diagonalized efficiently using the DFT matrix, thereby making the dual rigid regression problem can be solved entirely in the frequency domain \cite{henriques2015high}. Due to this attractive property, we first show that it is easy to embed the ``kernel trick" on $(\text{MF})^2\text{JMT}$, we will also show that it is possible to obtain non-linear filters as fast as linear correlation filters using the dense sampling, both to train and evaluate. We term this improvement kernel multi-frame multi-feature joint modeling tracker ($\text{K}(\text{MF})^2\text{JMT}$).

Denote
\begin{eqnarray}
\ve{K}^{xx^{new}}&\doteq&\ve{X}^{new}\ve{X}^T,\nonumber\\
\ve{K}^{zz^{new}}&\doteq&\ve{Z}^{new}\ve{Z}^T,\nonumber\\
\ve{K}^{xx}&\doteq&\ve{X}\ve{X}^T,\nonumber\\
\ve{K}^{zz}&\doteq&\ve{Z}\ve{Z}^T,
\end{eqnarray}
we have
\begin{eqnarray}
\ve{U}^{new}\ve{R}^{-1}\ve{U}^T
&=&
\left(\begin{array}{cc}
\frac{M}{\gamma_1}\ve{K}^{xx^{new}} & 0\\
0 & \frac{M}{\eta_1}\ve{K}^{zz^{new}}
\end{array}
\right),
\label{Eq28}
\end{eqnarray}
and
\vspace{-0.3cm}
%\begin{small}
\begin{flalign}
&\ve{U}\ve{R}^{-1}\ve{U}^T+\ve{D}^{-1} = \nonumber\\
&\left(\begin{array}{cc}
\Big(\frac{M}{\gamma_1}\ve{K}^{xx} + \tau^{-1}(\lambda_1+\lambda_2)\ve{I}\Big) & \Big(\tau^{-1}\lambda_2\ve{I}\Big)\\
\Big(\tau^{-1}\lambda_2\ve{I}\Big) & \Big(\frac{M}{\eta_1}\ve{K}^{zz}+\tau^{-1}(1+\lambda_2)\ve{I}\Big)
\end{array}
\right),\nonumber\\
\end{flalign}
%\end{small}
where $\tau = (1+\lambda_2)(\lambda_1+\lambda_2)-\lambda_2^2=\lambda_1+\lambda_2+\lambda_1\lambda_2$.

According to (\ref{EqlargeXZ}), $\ve{K}^{xx^{new}}$ consists of $M\times M$ block matrices and the $(i,j)$-th ($i,j=1,...,M$) block matrix can be represented as: %\footnote{We term $\ve{K}^{xx^{new}}$ $\emph{blockwise diagonal matrix}$. More properties of $\emph{blockwise diagonal matrix}$ are presented in Appendix A.}
\begin{eqnarray}
\ve{K}^{x x^{new}}_{ij} = \overline{X}_i^{new}\overline{X}_j^T
=(\frac{1}{\mu_1}+\delta_{ij})X_i^{new}X_j^T,
\label{Eq30}
\end{eqnarray}
where $\delta_{ij}=\mathbf{1}_{\{i=j\}}$ with $\mathbf{1}_{\{\cdot\}}$ denotes the indicator function.

If we project $\ve{X}$ and $\ve{Z}$ onto the Reproducing Kernel Hilbert Space (RKHS), i.e., applying a non-linear transform $\phi$ to both $\ve{X}$ and $\ve{Z}$, we can obtain the kernelized version of (\ref{MTMVmodel2}), i.e., $\text{K}(\text{MF})^2\text{JMT}$. According to \cite{henriques2015high}, (\ref{Eq30}) can be represented as:
\begin{eqnarray}
\ve{K}^{x x^{new}}_{ij} = \left(\frac{1}{\mu_1}+\delta_{ij}\right) C(\ve{k}^{\ve{x}_j\ve{x}^{new}_i}),
\end{eqnarray}
where $C(\ve{x})$ denotes a circular matrix generated by $\ve{x}$ (see Appendix A for more details). Similarly,
\begin{eqnarray}
\ve{K}^{x x}_{ij} &=& \left(\frac{1}{\mu_1}+\delta_{ij}\right) C(\ve{k}^{\ve{x}_j\ve{x}_i}),\\
\ve{K}^{z z^{new}}_{ij} &=& \left(\frac{1}{\mu_2}+\delta_{ij}\right) C(\ve{k}^{\ve{z}_j\ve{z}^{new}_i}),\\
\ve{K}^{z z}_{ij} &=& \left(\frac{1}{\mu_2}+\delta_{ij}\right) C(\ve{k}^{\ve{z}_j\ve{z}_i}).
\end{eqnarray}

Note that $C(\ve{k}^{\ve{x}_j\ve{x}^{new}_i})$, $C(\ve{k}^{\ve{x}_j\ve{x}_i})$, $C(\ve{k}^{\ve{z}_j\ve{z}^{new}_i})$ and $C(\ve{k}^{\ve{z}_j\ve{z}_i})$ are circular matrices, thus can be made diagonal as expressed below \cite{gray2006toeplitz}:
\begin{eqnarray}
C(\ve{k}^{\ve{x}_j\ve{x}^{new}_i}) &=& F diag(\hat{\ve{k}}^{\ve{x}_i\ve{x}^{new}_j})F^H,\nonumber\\
C(\ve{k}^{\ve{x}_j\ve{x}_i}) &=& F diag(\hat{\ve{k}}^{\ve{x}_i\ve{x}_j})F^H,\nonumber\\
C(\ve{k}^{\ve{z}_j\ve{z}^{new}_i}) &=& F diag(\hat{\ve{k}}^{\ve{z}_j\ve{z}^{new}_i})F^H,\nonumber\\
C(\ve{k}^{\ve{z}_j\ve{z}_i}) &=& F diag(\hat{\ve{k}}^{\ve{z}_i\ve{z}_j})F^H.
\label{Eq35}
\end{eqnarray}
where $F$ is the DFT matrix. Combining (\ref{Eq28})-(\ref{Eq35}), according to Appendix A, the $f(\ve{z})$ (defined in (\ref{kpred})) under kernel setting can be computed as:
\begin{eqnarray}
f(\ve{z}) &=& \ve{U}^{new}\ve{R}^{-1}\ve{U}^T\Big(\ve{U}\ve{R}^{-1}\ve{U}^T+\ve{D}^{-1}\Big)^{-1}\ve{D}^{-1}\ve{Y}\nonumber\\
&=&\ve{F}\Omega_1\ve{F}^H\left(\ve{F}\Omega_2\ve{F}^H\right)^{-1}\ve{D}^{-1}\ve{Y}\nonumber\\
&=&\ve{F}\Omega_1\Omega_2^{-1}\ve{F}^H\ve{D}^{-1}\ve{Y},
\label{Eq36}
\end{eqnarray}
where $\Omega_1$ and $\Omega_2$ are given in ($\mathrm{29}$) and $\ve{F}$ is a $\emph{block diagonal matrix}$ with blocks $F$ on its main diagonal.
%the number of blocks $F$ can be inferred from the context, thus omitted here.

\newcounter{storeeqcounter_two}
\newcounter{tempeqcounter}
%\addtocounter{equation}{1}%
\setcounter{storeeqcounter_two}{\value{equation}}%
\begin{figure*}[!t]
\scriptsize
\setcounter{tempeqcounter}{\value{equation}}
\begin{eqnarray}
\Omega_1 &=& \left(\begin{array}{*{2}{c}}
\frac{M}{\gamma_1}\left(\left(\frac{1}{\mu_1}+\delta_{ij}\right)diag(\ve{k}^{\ve{x}_j\ve{x}^{new}_i})\right)_{i,j=1}^M & 0\\
0 & \frac{M}{\eta_1}\left(\left(\frac{1}{\mu_2}+\delta_{ij}\right)diag(\ve{k}^{\ve{z}_j\ve{z}^{new}_i})\right)_{i,j=1}^M
\end{array}
\right)\nonumber\\
&=& \left(\begin{array}{*{2}{c}}
\left(\left(\frac{1}{\gamma_2}+\frac{M}{\gamma_1}\delta_{ij}\right)diag(\ve{k}^{\ve{x}_j\ve{x}^{new}_i})\right)_{i,j=1}^M & 0\\
0 & \left(\left(\frac{1}{\eta_2}+\frac{M}{\eta_1}\delta_{ij}\right)diag(\ve{k}^{\ve{z}_j\ve{z}^{new}_i})\right)_{i,j=1}^M
\end{array}
\right),\nonumber\\
\Omega_2 &=& \left(\begin{array}{*{2}{c}}
\frac{M}{\gamma_1}\left(\left(\frac{1}{\mu_1}+\delta_{ij}\right)diag(\ve{k}^{\ve{x}_j\ve{x}_i})\right)_{i,j=1}^M+\tau^{-1}(\lambda_1+\lambda_2)\ve{I} & \Big(\tau^{-1}\lambda_2\ve{I}\Big)\\
\Big(\tau^{-1}\lambda_2\ve{I}\Big) &
\frac{M}{\eta_1}\left(\left(\frac{1}{\mu_2}+\delta_{ij}\right)diag(\ve{k}^{\ve{z}_j\ve{z}_i})\right)_{i,j=1}^M+\tau^{-1}(1+\lambda_2)\ve{I}
\end{array}
\right)\nonumber\\
&=& \left(\begin{array}{*{2}{c}}
\left(\left(\frac{1}{\gamma_2}+\frac{M}{\gamma_1}\delta_{ij}\right)diag(\ve{k}^{\ve{x}_j\ve{x}_i})\right)_{i,j=1}^M+\tau^{-1}(\lambda_1+\lambda_2)\ve{I} & \Big(\tau^{-1}\lambda_2\ve{I}\Big)\\
\Big(\tau^{-1}\lambda_2\ve{I}\Big) &
\left(\left(\frac{1}{\eta_2}+\frac{M}{\eta_1}\delta_{ij}\right)diag(\ve{k}^{\ve{z}_j\ve{z}_i})\right)_{i,j=1}^M+\tau^{-1}(1+\lambda_2)\ve{I}
\end{array}
\right).\nonumber\\
\end{eqnarray}
\setcounter{equation}{\value{tempeqcounter}} % restore correct value
\hrulefill
% The spacer can be tweaked to stop underfull vboxes.
\vspace*{4pt}
\end{figure*}
\addtocounter{equation}{1}%

Denote
\begin{eqnarray}
L^x&=&
\frac{M}{\gamma_1}\left(\left(\frac{1}{\mu_1}+\delta_{ij}\right)diag(\ve{k}^{\ve{x}_j\ve{x}_i})\right)_{i,j=1}^M,\nonumber\\
L^z&=& \frac{M}{\eta_1}\left(\left(\frac{1}{\mu_2}+\delta_{ij}\right)diag(\ve{k}^{\ve{z}_j\ve{z}_i})\right)_{i,j=1}^M, \end{eqnarray}
then
\begin{eqnarray}
\Omega_2 = \left(\begin{array}{cc}
L^x+\tau^{-1}(\lambda_1+\lambda_2)\ve{I} & \Big(\tau^{-1}\lambda_2\ve{I}\Big)\\
\Big(\tau^{-1}\lambda_2\ve{I}\Big) & L^z+\tau^{-1}(1+\lambda_2)\ve{I}
\end{array}
\right).
\end{eqnarray}

According to the Equation ($\mathrm{2.76}$) of \cite{bishop2006pattern}, we have:
\begin{small}
\begin{flalign}
&\Omega_2^{-1} = \nonumber\\
&\left(\begin{array}{cc}
\Big(L^z+\tau^{-1}(1+\lambda_2)\ve{I}\Big)A^{-1} & \Big(-\tau^{-1}\lambda_2\ve{I}\Big)B^{-1}\\
\Big(-\tau^{-1}\lambda_2\ve{I}\Big)A^{-1} & \Big(L^x + \tau^{-1}(\lambda_1+\lambda_2)\ve{I}\Big)B^{-1}
\end{array}
\right),\nonumber\\
\end{flalign}
\end{small}
where
\begin{small}
\begin{eqnarray}
A &=& \left(L^x+ \tau^{-1}(\lambda_1+\lambda_2)\ve{I}\right)
\left(L^z+\tau^{-1}(1+\lambda_2)\ve{I}\right)-\tau^{-2}\lambda_2^2\ve{I},\nonumber\\
%&=& L^xL^z + \tau^{-1}(1+\lambda_2)L^x +
%\tau^{-1}(\lambda_1+\lambda_2)L^z+\tau^{-1}\ve{I}
%\\
B &=& \left(L^z+\tau^{-1}(1+\lambda_2)\ve{I}\right)\left(L^x+ \tau^{-1}(\lambda_1+\lambda_2)\ve{I}\right)
-\tau^{-2}\lambda_2^2\ve{I}\nonumber\\
&=& A^T.
\end{eqnarray}
\end{small}
It is obvious that $A$ and $B$ are $Mn\times Mn$ $\emph{blockwise diagonal matrix}$ that can be partitioned into $M\times M$ diagonal matrices of size $n\times n$. According to \emph{Theorem \ref{theo:ComInvBloDiag}}, the computational cost for $A^{-1}$ or $B^{-1}$ is $\mathcal{O}(nM^3)$. Besides, it takes $nM^3$ product operations to compute $L^x\times L^z$, $\Big(L^z+\tau^{-1}(1+\lambda_2)\ve{I}\Big)\times A^{-1}$ and $\Big(L^x + \tau^{-1}(\lambda_1+\lambda_2)\ve{I}\Big)\times B^{-1}$. In this sense, the computational cost for $\Omega_2^{-1}$ is still $\mathcal{O}(nM^3)$. On the other hand, the DFT bounds the cost at nearly $\mathcal{O}(n\log n)$ by exploiting the circulant structure \cite{henriques2015high}. Therefore, the overall cost for computing $f(\ve{z})$ is $\mathcal{O}(Mn\log n + nM^3)$ given that there are $2M$ inverse DFTs in (\ref{Eq36}).

\begin{theorem}\label{theo:ComInvBloDiag}
Given an invertible $\emph{blockwise diagonal matrix}$ $\ve{S}$ that can be partitioned into $M\times M$ diagonal matrices of size $n\times n$, the computational complexity of $\ve{S}^{-1}$ is $\mathcal{O}(nM^3)$.
\end{theorem}

\begin{IEEEproof}
Denote $S_{ij}$ ($i,j=1,...,M$) the $(i,j)$-th block of $\ve{S}$, where $S_{ij}$ is a diagonal matrix of size $n\times n$. After a series of elementary matrix operations, e.g., pre-multiply or post-multiply the matrix $\ve{S}$ by different elementary matrices,  we can interchange the rows and columns of $\ve{S}$ arbitrarily. Therefore, there exists an invertible matrix $\ve{P}$ with $\ve{P}\ve{P}^T=\ve{I}$, such that the matrix $\tilde{\ve{S}}=\ve{P}\ve{S}\ve{P}^T\triangleq (\tilde{S}_{\tilde{i}\tilde{j}})_{\tilde{i},\tilde{j}=1,\cdots,n}$ satisfies: the elements of $\tilde{S}_{\tilde{i}\tilde{j}}$ come from $\ve{S}$ with row indices $(\tilde{i},n+\tilde{i}, 2n+\tilde{i},\cdots,(M-1)n+\tilde{i})$ and column indices $(\tilde{j},n+\tilde{j},2n+\tilde{j},\cdots,(M-1)n+\tilde{j})$. Obviously, $\tilde{S}_{\tilde{i}\tilde{j}}=\ve{0}$ for $\tilde{i}\neq\tilde{j}$. Thus, $\tilde{\ve{S}}^{-1}$ can be represented as:
\begin{eqnarray}
\tilde{\ve{S}}^{-1}=\left(\begin{array}{cccc}
\tilde{S}_{11}^{-1} & \ve{0} & \cdots & \ve{0}\\
\ve{0} & \tilde{S}_{22}^{-1} & \cdots & \ve{0}\\
\vdots & \vdots & \ddots & \vdots\\
\ve{0} & \ve{0} & \cdots & \tilde{S}_{nn}^{-1}
\end{array}
\right).
\end{eqnarray}

Given that $\ve{S}^{-1}=\ve{P}^T\tilde{\ve{S}}^{-1}\ve{P}$, which means $\ve{S}^{-1}$ can be obtained by allocating the elements of $\tilde{S}_{\tilde{i}\tilde{i}}^{-1}$ to locations with row indices $(\tilde{i},n+\tilde{i}, 2n+\tilde{i},\cdots,(M-1)n+\tilde{i})$ and column indices $(\tilde{j},n+\tilde{j},2n+\tilde{j},\cdots,(M-1)n+\tilde{j})$. The main computational cost of $\ve{S}^{-1}$ comes from the calculation of $\tilde{S}_{\tilde{i}\tilde{i}}^{-1},i=1,\cdots,n$. The size of $\tilde{S}_{\tilde{i}\tilde{i}}$ is $M\times M$, thus the computational complexity of $\tilde{S}_{\tilde{i}\tilde{i}}$ is $\mathcal{O}(M^3)$. As a result, the computational complexity of $\ve{S}^{-1}$ is $\mathcal{O}(nM^3)$.
\end{IEEEproof}

%\textcolor{red}{Multiply two matrices of this kind, we need $nT^3$ multiplications. Thus to compute $\Big(\ve{U}^T\ve{R}^{-1}\ve{U}+\ve{D}^{-1}\Big)^{-1}$, the complexity is $4nT^3$. In this paper, we simply use the object template updating scheme in \cite{henriques2015high} for robust visual tracking.}

\section{Scale Adaptive $\text{K}(\text{MF})^2\text{JMT}$}
To further improve the overall performance of $\text{K}(\text{MF})^2\text{JMT}$, we follow the Integrated Correlation Filters (ICFs) framework in \cite{hong2015multi} to cope with scale variations. The ICF is a cascading-stage filtering process that performs translation estimation and scale estimation, respectively (same as the pipeline adopted in \cite{danelljan2014accurate} and \cite{li2014scale}). Unless otherwise specified, the $\text{K}(\text{MF})^2\text{JMT}$ mentioned in the following experimental parts refers to the scale adaptive one.

Specifically, in scale adaptive $\text{K}(\text{MF})^2\text{JMT}$, the training of basic $\text{K}(\text{MF})^2\text{JMT}$ is accompanied by the training of another $1$D Discriminative Scale Space Correlation Filter (DSSCF) \cite{danelljan2014accurate} and this new trained filter is performed for scale estimation. To evaluate the trained DSSCF, $S$ image patches centered around the location found by the $\text{K}(\text{MF})^2\text{JMT}$ are cropped from the image, each of size $a^sL\times a^sN$, where $L\times N$ is the target size in the current frame, $a$ is the scale factor, and $s\in\{-\frac{S-1}{2}, ...\frac{S-1}{2}\}$. All $S$ image patches are then resized to the template size for the feature extraction. Finally, the final output from the scale estimation is given as the image patch with the highest filtering response. Similar to $\text{K}(\text{MF})^2\text{JMT}$, the model parameters are also updated in an interpolating manner with learning rate $\eta$. We refer readers to \cite{danelljan2014accurate} for more details and the implementation of DSSCF.

\section{Experiments}
We conduct four groups of experiments to demonstrate the effectiveness and superiority of our proposed $\text{K}(\text{MF})^2\text{JMT}$. First, we implement $\text{K}(\text{MF})^2\text{JMT}$ and several of its baseline variants, including multi-feature-only tracker ($\text{MFT}$), multi-frame-only tracker ($\text{MFT-2}$), scale-adaptive-only tracker ($\text{SAT}$), scale-adaptive multi-feature tracker ($\text{SAMFT}$) and scale-adaptive multi-frame tracker ($\text{SAMFT-2}$), to analyze and evaluate the component-wise contributions to the performance gain. We then evaluate our tracker against $33$ state-of-the-art trackers on Object Tracking Benchmark (OTB) $2015$~\cite{wu2015object}. Following this, we present results on Visual Object Tracking (VOT) $2015$ challenge~\cite{kristan2015visual}. Finally, we compare $\text{K}(\text{MF})^2\text{JMT}$ with several other representative CFTs, including MOSSE \cite{bolme2010visual}, SAMF \cite{li2014scale}, MUSTer \cite{hong2015multi} and the recently published C-COT\cite{danelljan2016beyond}, to reveal the properties of our proposed tracker among CFTs family and also illustrate future research directions for modern discriminative tracker design.

\subsection{Experimental setup}
%\paragraph{Datasets and evaluation methods}
To make a comprehensive evaluation, we use all the color video sequences in OTB 2015 dataset ($77$ in total). These sequences are captured in various conditions, including occlusion, deformation, illumination variation, scale variation, out-of-plane rotation, fast motion, background clutter, etc. %In VOT 2014 dataset, the accuracy is measured by the VOR score. The robustness indicates the failing time for a tracker on the sequence.
We use the success plot to evaluate all trackers on OTB dataset. The success rate counts the percentage of the successfully tracked frames by measuring the overlap score $S$ for trackers on each frame. The average overlap measure is the most appropriate for tracker comparison, which accounts for both position and size. Let $B_T$ denote the tracking bounding box and $B_G$ denote the ground truth bounding box, the overlap score is defined as $S=\frac{|B_T\cap B_G|}{|B_T\cup B_G |}$, where $\cap$ and $\cup$ represent the intersection and union of two regions, and $|\cdot|$ denotes the number of pixels in the region. In success plot, $S$ is varied from $0$ to $1$, and the ranking of trackers is based on the Area Under Curve (AUC) score. We also report the speed of trackers in terms of average frames per second (FPS) over all testing sequences.

We also test tracker performance on VOT $2015$ dataset containing $60$ video sequences. The VOT challenge is a popular competition for single object tracking. Different from OTB 2015 dataset, a tracker is restarted in the case of a failure (i.e., there is no overlap between the detected bounding box and ground truth) in the VOT 2015 data set. For the VOT 2015 dataset, tracking performance is evaluated in terms of accuracy (overlap with the ground-truth), robustness (failure rate) and the expected average overlap (EAO). EAO is obtained by combining the raw values of accuracy and failures. Its score represents the average overlap a tracker would obtain on a typical short-term sequence without reset. For a full treatment of these metrics, interested readers can refer to \cite{kristan2015visual}.

\subsection{Implementation details of the proposed tracker}
We set the regularization parameters to $\lambda_1=0.5$, $\lambda_2=1.32$, $\gamma_1=0.0006$, $\gamma_2=0.005$, $\eta_1=0.001$, $\eta_2=0.005$. These parameters are tuned with a coarse-to-fine procedure. In the coarse module, we roughly determine a satisfactory range for each parameter (for example, the range of $\gamma_1$ is $[0.0001~0.001]$). Here, the word ``satisfactory" means that the value in the range can achieve higher mean success rate at the threshold $0.5$ in the OTB 2015 dataset. Then, in the fine-tuning module, we divide these parameters into three groups based on their correlations: (1) $\{\lambda_1,\lambda_2\}$; (2) $\{\gamma_1,\gamma_2\}$; and (3) $\{\eta_1,\eta_2\}$. When we test the value of one group of parameters, other groups are set to default values, i.e., the mean value of the optimal range given by the coarse module. In each group, the parameters are tuned with grid search (for example, $\gamma_1$ in the first group is tuned at the range $[0.0001~0.001]$ with an interval $0.0001$). We finally pinpointed the specific value as the one that can achieve the highest mean success rate (for example, the final value of $\gamma_1$ is $0.0006$).
Besides, we set the learning rate to $\eta = 0.025$ as previous work \cite{henriques2015high,li2014scale,danelljan2014accurate} and model three consecutive frames in the MTL module, i.e., $M=3$. We select HOGs \cite{dalal2005histograms} and color names \cite{van2009learning} for image representation in the MVL module. The HOGs and color names are complementary to each other, as the former puts emphasis on the image gradient which is robust to illumination and deformation while the latter focuses on the color information which is robust to motion blur. For HOGs, we employ a variant in \cite{felzenszwalb2010object}, with the implementation provided by \cite{PMT}. More particular, the cell size is $4\times4$ and number of orientations is set to $9$. %To coincident with HOG feature, we also resize RGB candidate samples to $1/4$ of their original size. %Finally, the extracted features are always multiplied by a Hanning window, as described in \cite{felzenszwalb2010object}.
For color names, we map the RGB values to a probabilistic $11$ dimensional color representation which sums up to $1$. All the experiments mentioned in this work are conducted using MATLAB on an Intel i$7$-$4790$ $3.6$GHz Quad-Core PC with $8$GB RAM. MATLAB code is available from our project homepage \url{http://bmal.hust.edu.cn/project/KMF2JMTtracking.html}.

\subsection{Evaluation on component-wise contributions} \label{experiment_section1}
Before systematically evaluating the performance of our proposed tracker, we first compare $\text{K}(\text{MF})^2\text{JMT}$ with its baseline variants to demonstrate the component-wise contributions to the performance gain. To this end, we implement seven trackers with various degraded settings, including multi-feature-only tracker (MFT) which just uses multiple feature cues, multi-frame-only tracker (MFT-2) which just uses the temporal relatedness across consecutive frames, scale-adaptive-only tracker (SAT) which just concerns scale variations, scale-adaptive multi-feature tracker (SAMFT) and scale-adaptive multi-frame tracker (SAMFT-2). Besides, to validate the efficiency of modeling multiple feature cues under a MVL framework rather than simply concatenating them, we also implement Multiple Feature tracker with feature concatenation (MFT-C). Note that, the KCF \cite{henriques2015high}, which only uses the HOG feature without temporal modeling and scale searching (i.e., $\lambda_1=\lambda_2=0$, $\eta_1=\gamma_1=+\infty$, $M=1$), serves as a baseline in this section.

Table \ref{lab:component_difference} summarized the differences between these trackers, where $\text{K}(\text{MF})^2\text{JMT-2}$ denotes the basic kernel multi-frame multi-feature joint modeling tracker without scale estimation. In this table, we report the mean success rate at the threshold of $0.5$, which corresponds to the PASCAL evaluation criterion~\cite{everingham2010pascal}. Although these trackers share one or more common components, their tracking performances differ significantly. This indicates that the visual features, temporal relatedness and scale searching strategy all are essentially important to the visual tracking tasks. KCF ranks the lowest among the compared trackers as expected. MFT (or MFT-C), MFT-2 and SAT extend KCF by augmenting the feature space with color information, taking advantage of temporal information with frame relatedness and introducing scale adaptive searching strategy respectively, thus achieving a few improvements. MFT outperforms MFT-C with a more ``intelligent" feature fusion strategy. Besides, it is obvious that the scale adaptive searching can effectively handle scale variations, thus obtaining a large improvement in success rate (see SAT, SAMFT, SAMFT-2 as against KCF, MFT and MFT-2 in Table \ref{lab:component_difference}). Moreover, by comparing MFT-2 with MFT and comparing SAMFT-2 with SAMFT, one can see that the integration of multiple frames plays a more positive role to improve tracking performance than the integration of multiple features. Finally, it is interesting to find that the success rate gains of MFT, MFT-2 and SAT are $3.2\%$, $3.3\%$ and $5.4\%$ respectively compared with KCF, while $\text{K}(\text{MF})^2\text{JMT}$ gets a $13.3\%$ improvement. This indicates that the $\text{K}(\text{MF})^2\text{JMT}$ is not just the simple combination of the $\text{MFT}$, $\text{MFT-2}$ and $\text{SAT}$.

\begin{table}
\centering
\caption{A comparison of our $\text{K}(\text{MF})^2\text{JMT}$ with different baseline variants. The mean overlap precision (OP) score ($\%$) at threshold $0.5$ over all the $77$ color videos in the OTB dataset are presented. The best two results are marked with red and blue respectively. MF, TM and SA are the abbreviation of Multiple Features, Temporal Modeling and Scale Adaptive, respectively.}\label{lab:component_difference}
\begin{tabular}{ccccc}\hline
 & MF & TM & SA & mean success rate \\\hline
$\text{KCF}$ & no & $\text{no}$ & $\text{no}$ & $51.0$ \\
$\text{MFT}$ & yes & no & no & $54.2$  \\
$\text{MFT-C}$ & yes & no & no & $53.7$  \\
$\text{MFT-2}$ & no & yes & no & $54.3$ \\
$\text{SAT}$ & no & no & yes & $56.4$ \\
$\text{SAMFT}$ & yes & no &  yes & $60.1$ \\
$\text{SAMFT-2}$ & no & yes & yes & $\color{blue}{61.7}$ \\
$\text{K}(\text{MF})^2\text{JMT-2}$ & yes & yes & no & $58.1$ \\
$\text{K}(\text{MF})^2\text{JMT}$ & yes & yes & yes & $\color{red}{64.3}$ \\\hline
\end{tabular}
\end{table}

%\begin{figure}[!t]
%\centering
%\includegraphics[width=.45\textwidth]{success_plot_KCF_family}
%\caption{Precision plots and success plots of $\text{K}(\text{MF})^2\text{JMT}$ and its baseline variants on OTB $2013$ dataset.}
%\label{Comparison_KCF_family}
%\end{figure}

\begin{figure*}[!t]
\centering
\subfigure[One pass evaluation (OPE)] {\includegraphics[width=.30\textwidth]{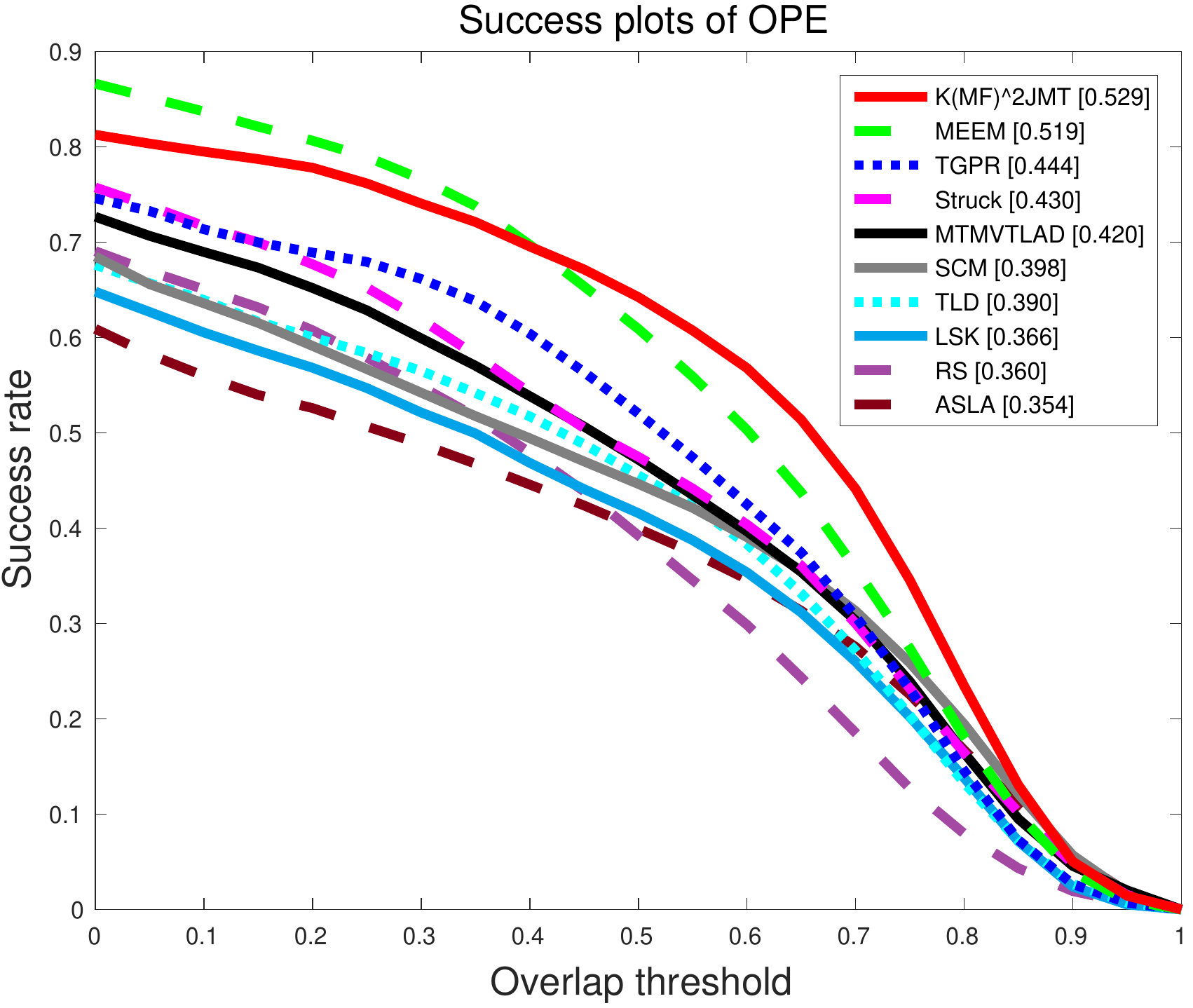}}
\subfigure[Temporal robustness evaluation (TRE)] {\includegraphics[width=.30\textwidth]{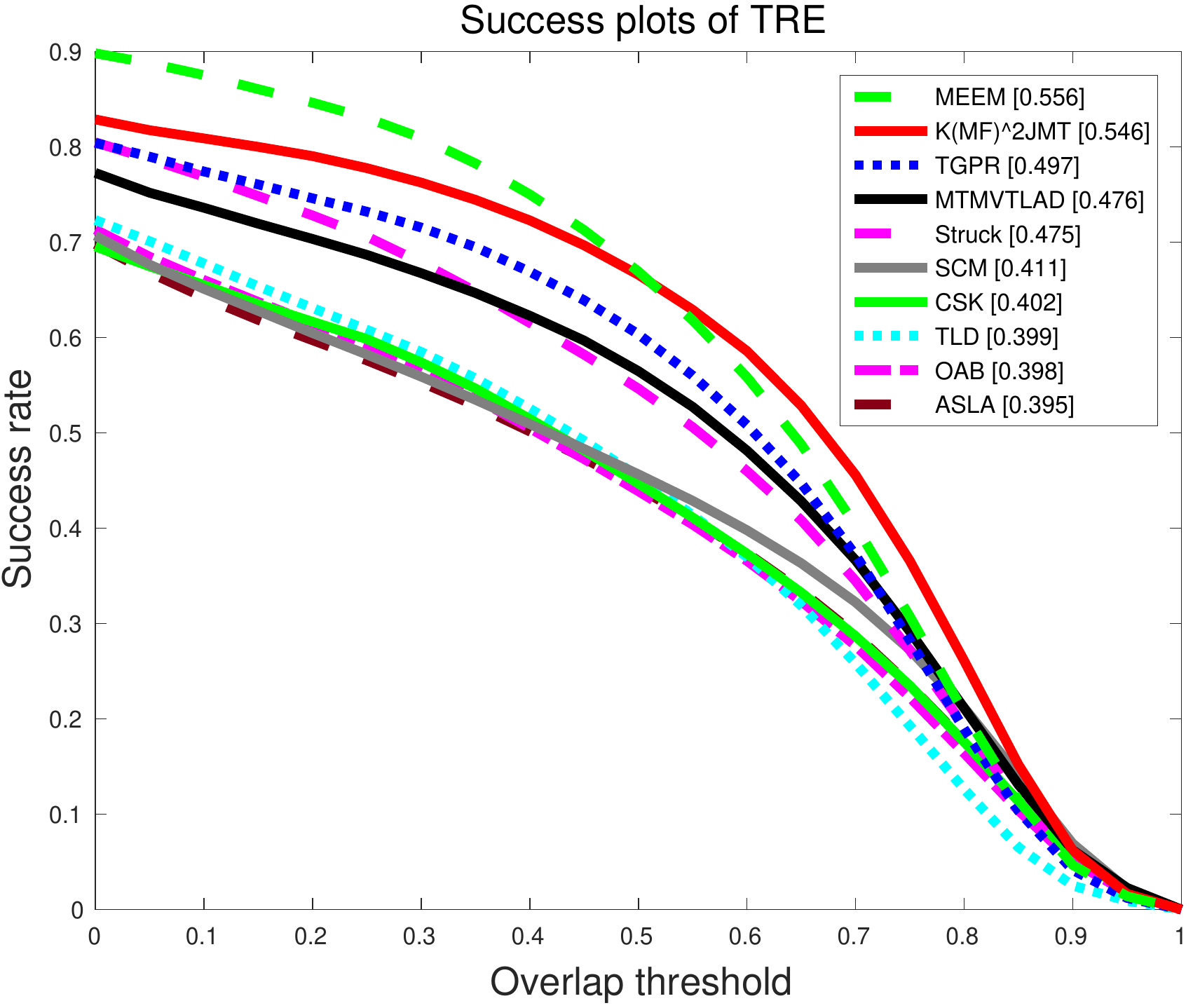}}
\subfigure[Spatial robustness evaluation (SRE)] {\includegraphics[width=.30\textwidth]{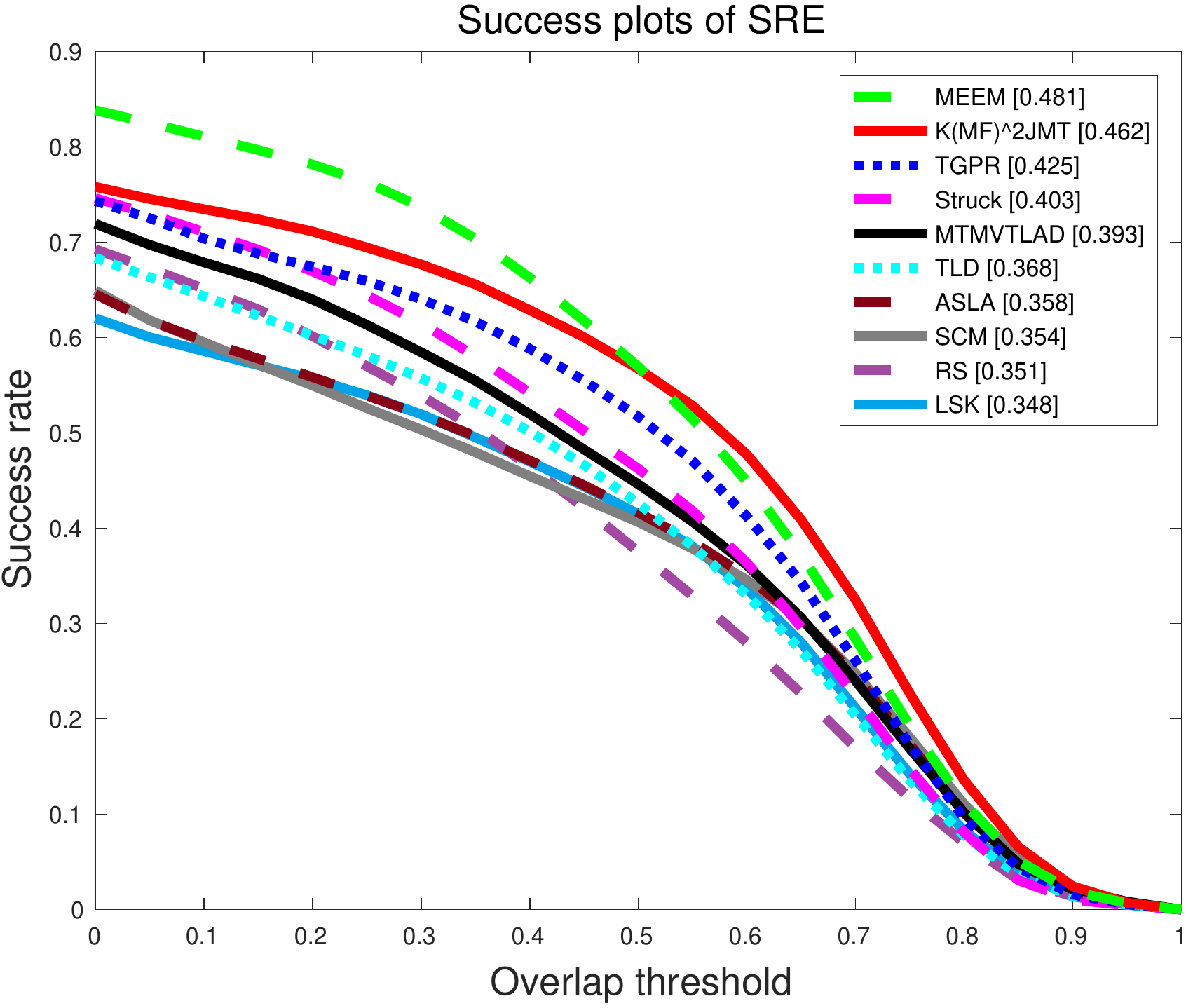}}
\caption{Success plot showing the performance of our $\text{K}(\text{MF})^2\text{JMT}$ compared to $33$ state-of-the-art methods on the OTB dataset. The AUC score for each tracker is reported in the legend. Only the top ten trackers are displayed in the legend for clarity in (a) OPE, (b) TRE and (c) SRE. Our approach provides the best performance in OPE and the second best performance in TRE and SRE.}
\label{OTBB_comparison plot}
\end{figure*}

\begin{figure*}[!htbp]
\centering
\begin{tabular}{cccc}
\includegraphics[width=.22\textwidth]{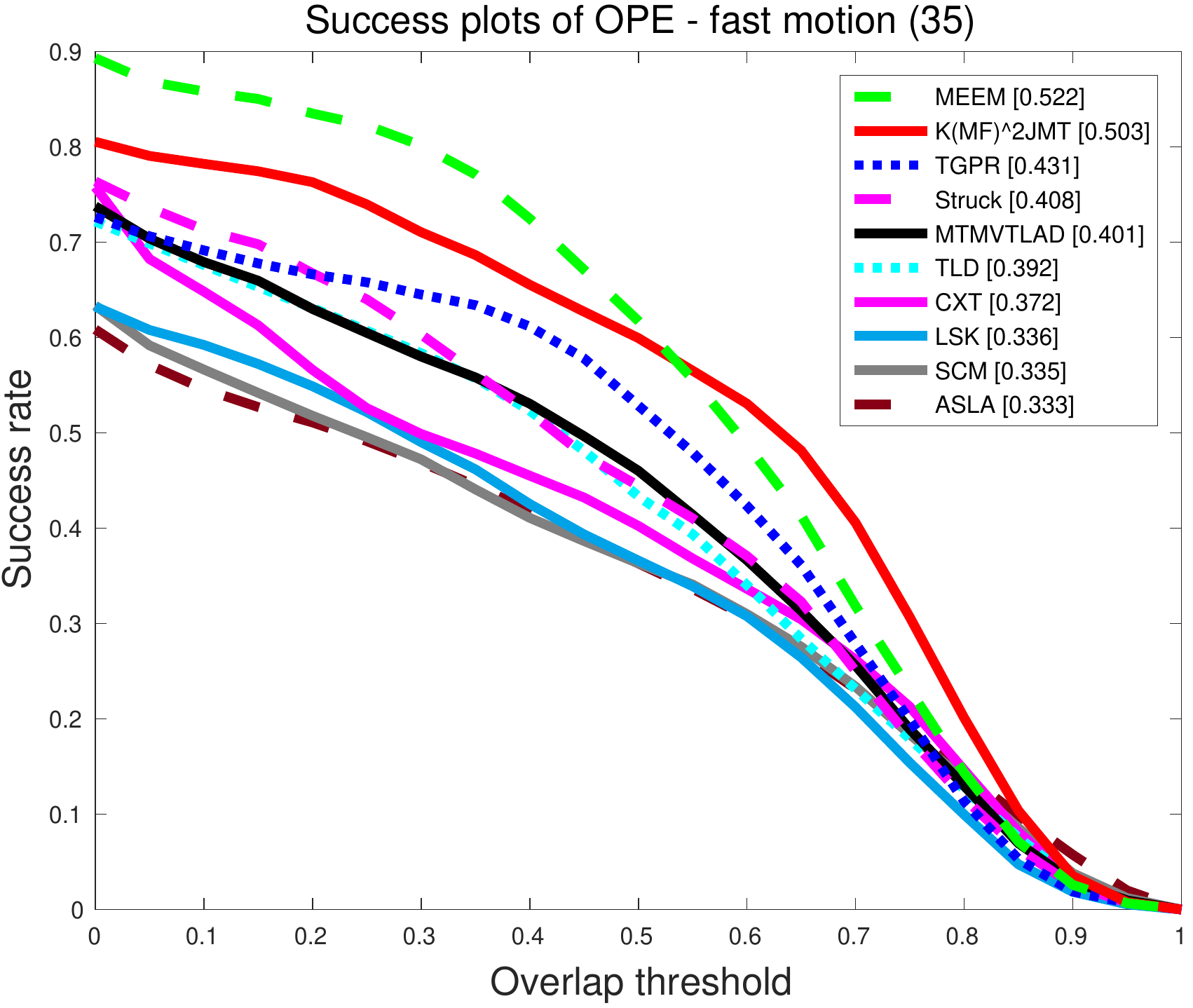} &
\includegraphics[width=.22\textwidth]{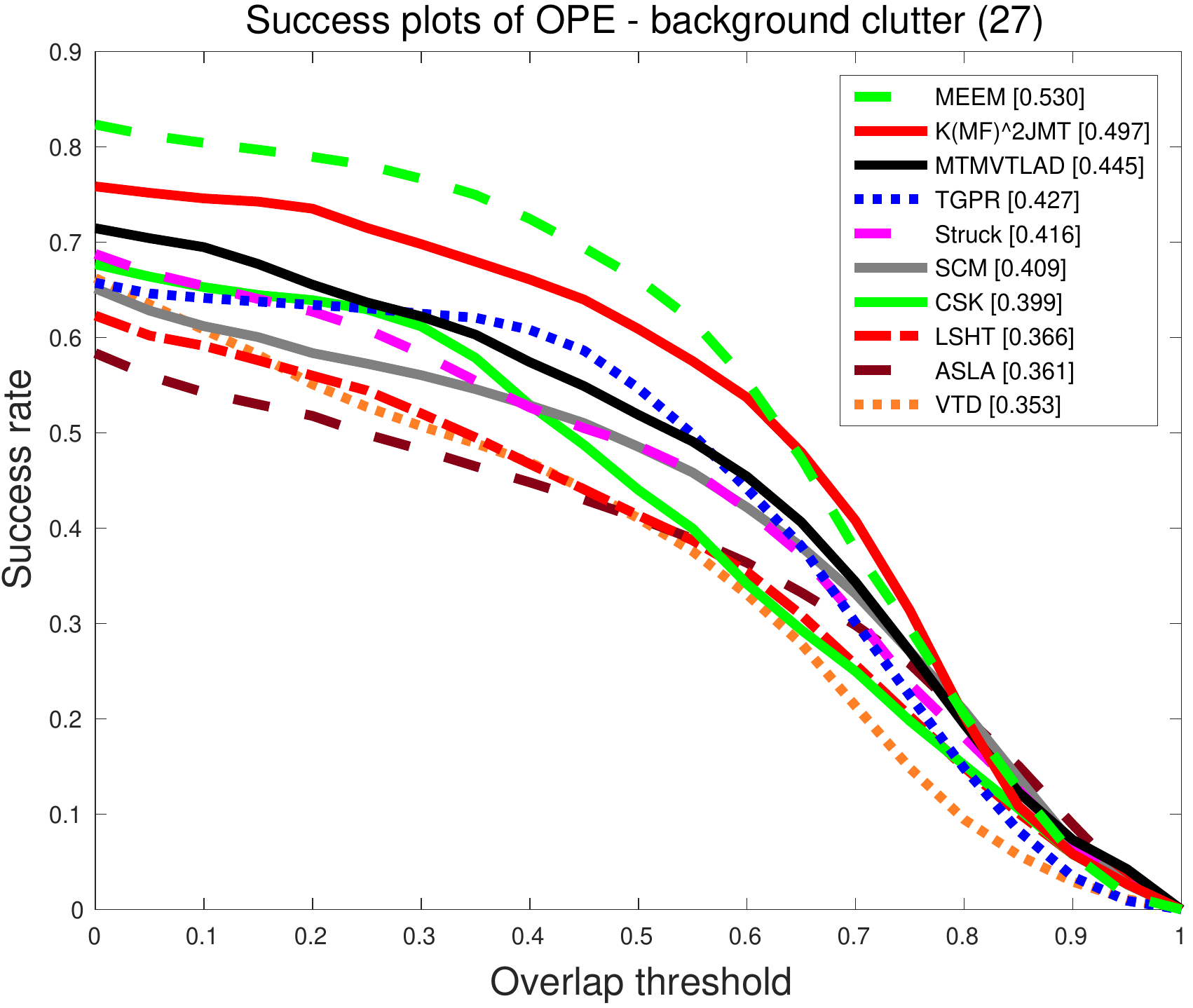} &
\includegraphics[width=.22\textwidth]{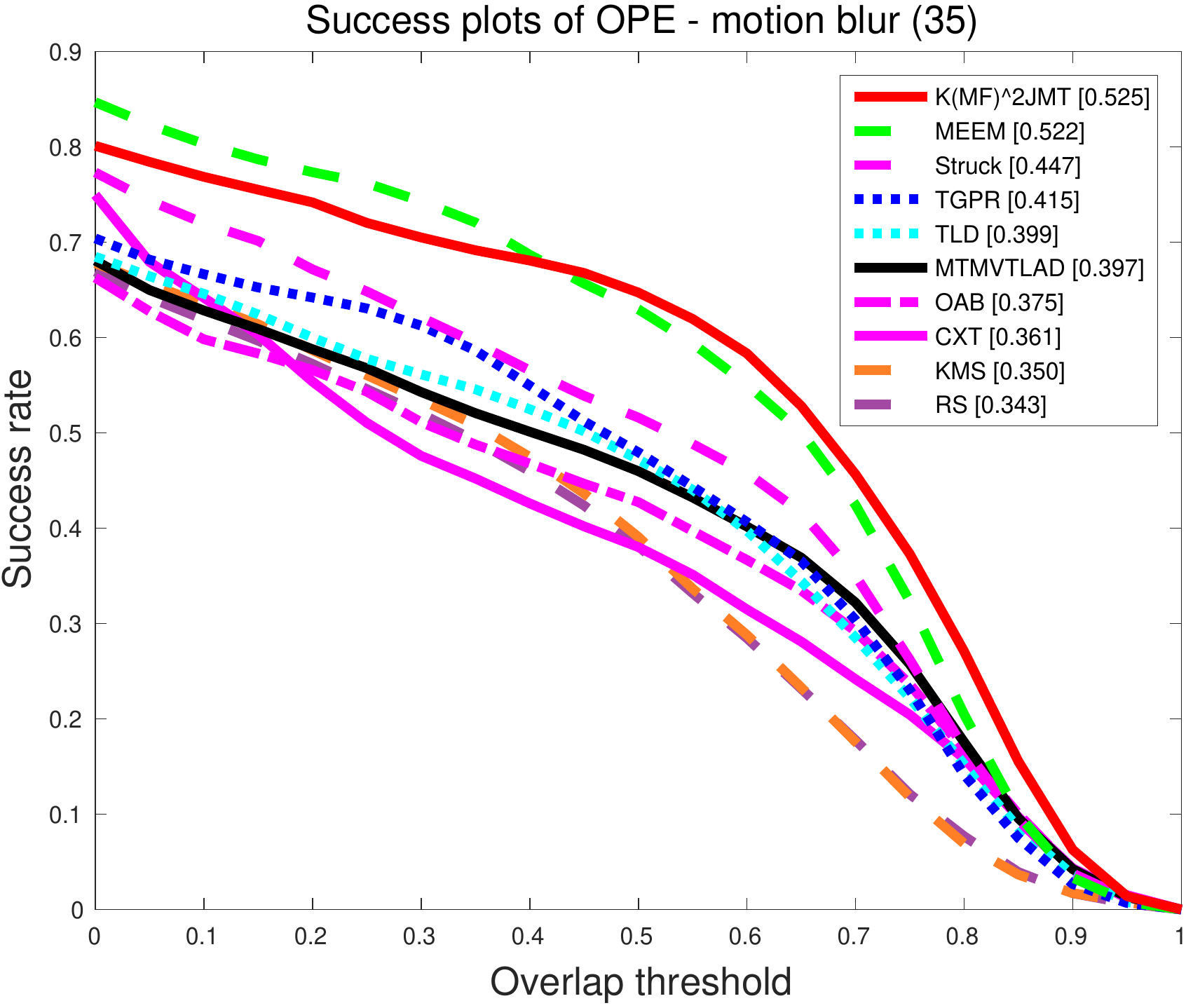}&
\includegraphics[width=.22\textwidth]{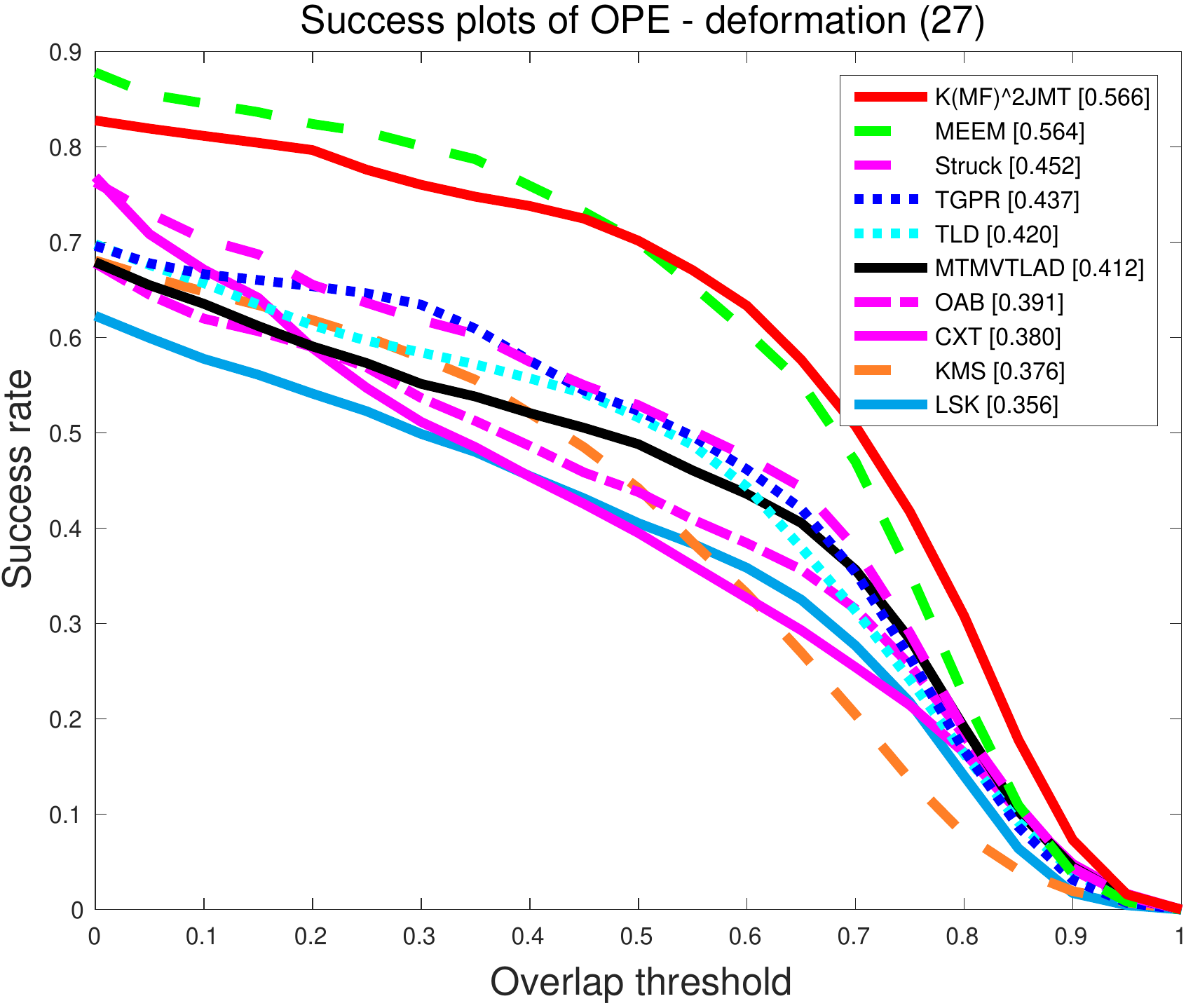} \\
\includegraphics[width=.22\textwidth]{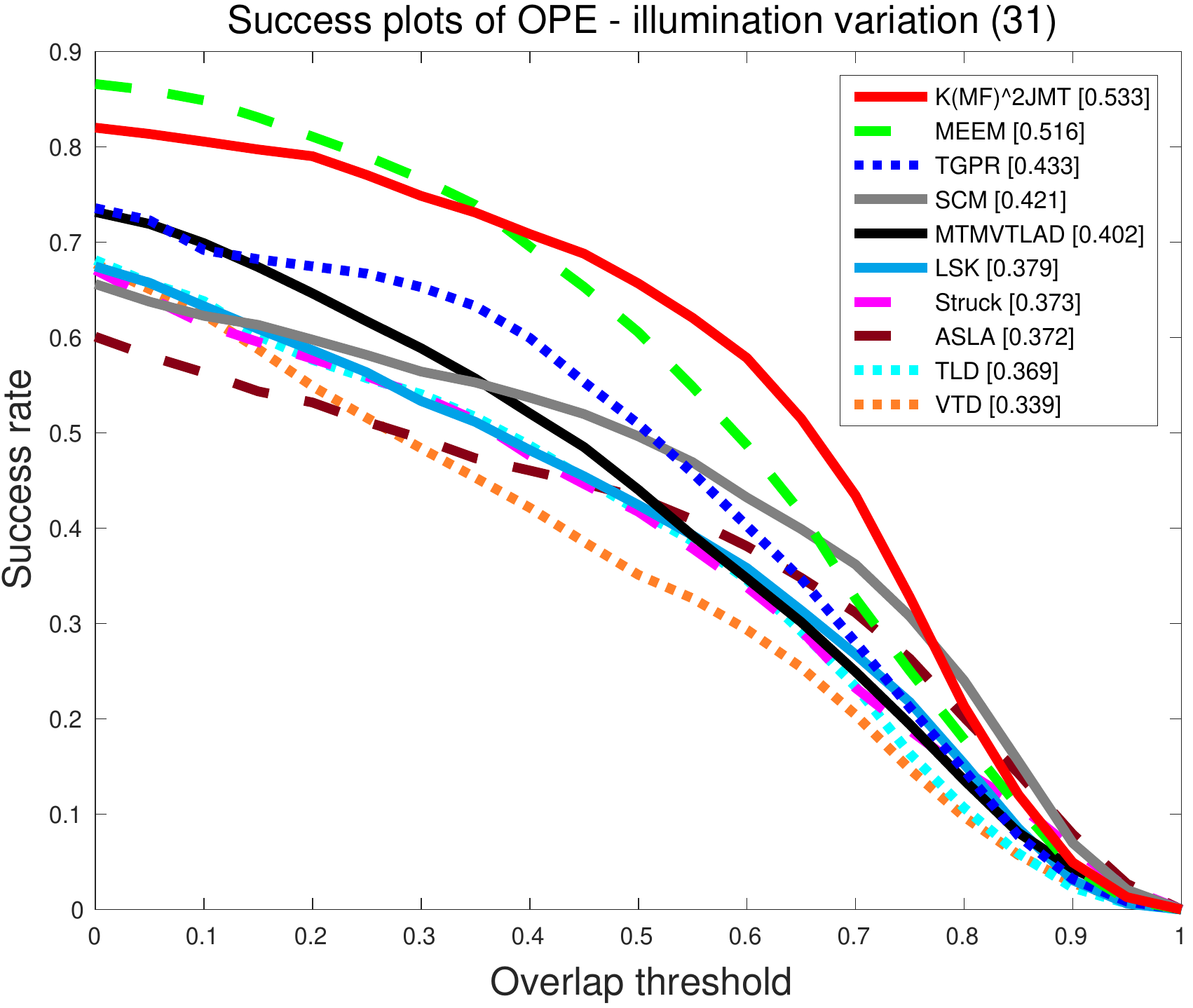} &
\includegraphics[width=.22\textwidth]{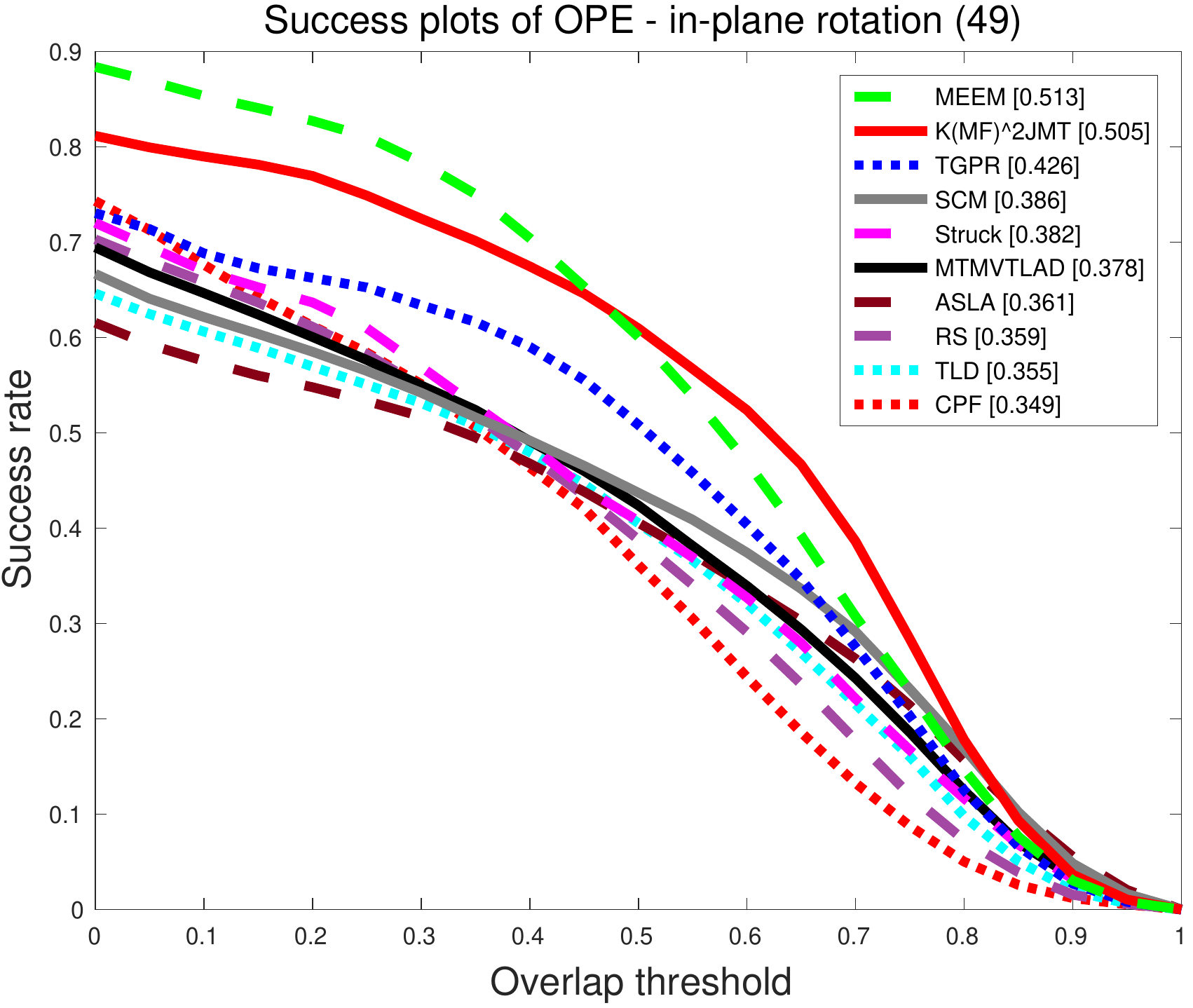} &
\includegraphics[width=.22\textwidth]{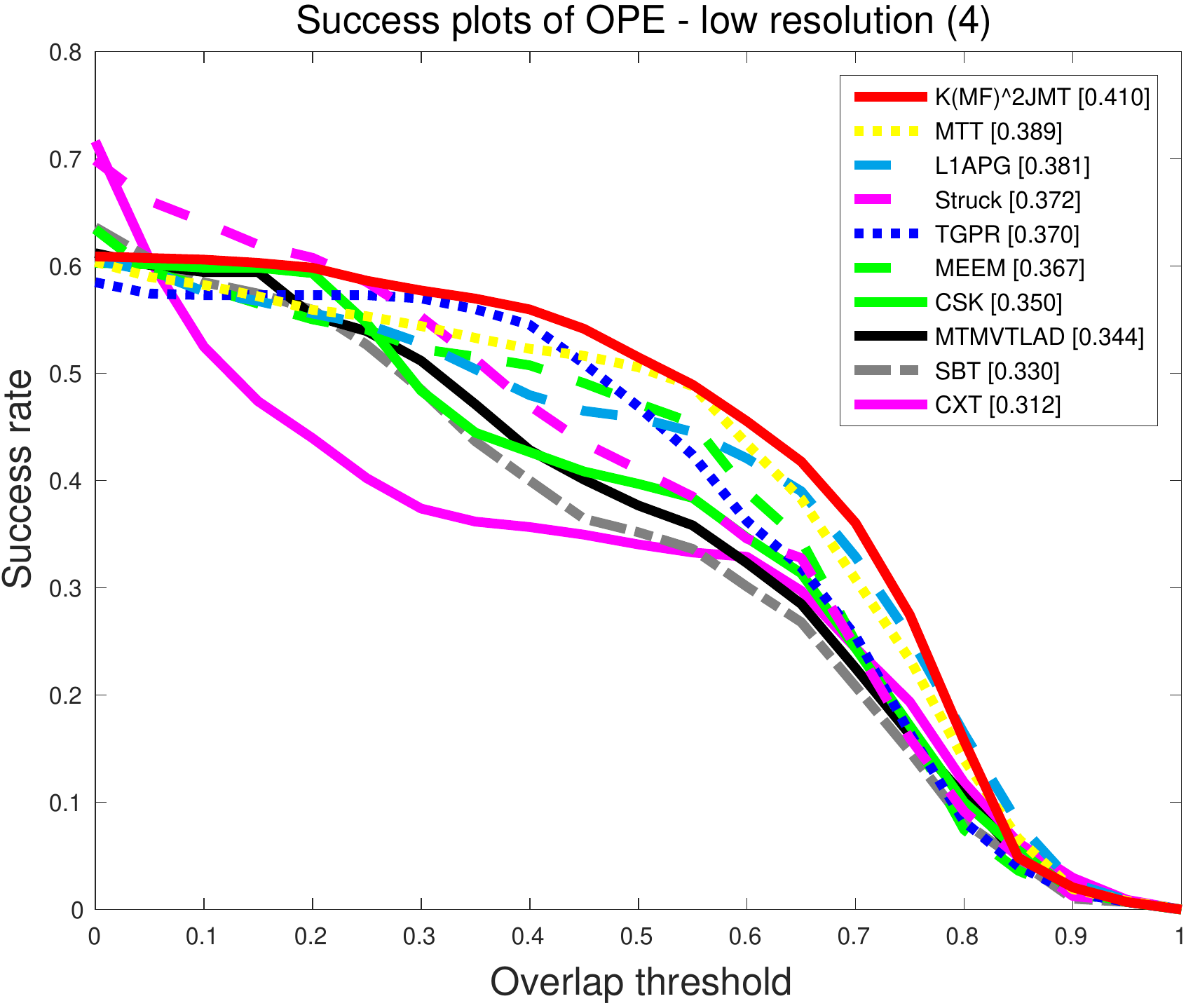} &
\includegraphics[width=.22\textwidth]{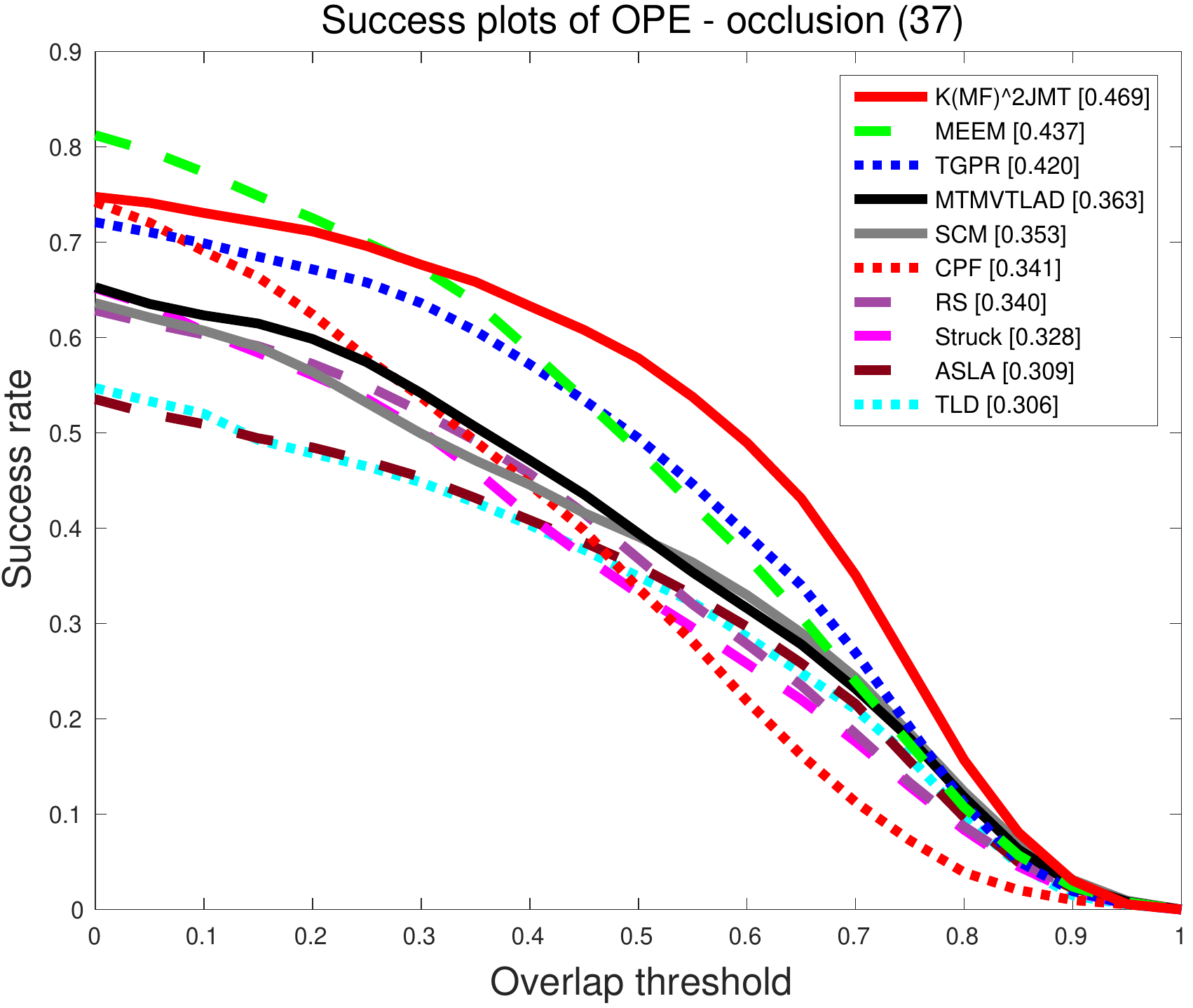} \\
\includegraphics[width=.22\textwidth]{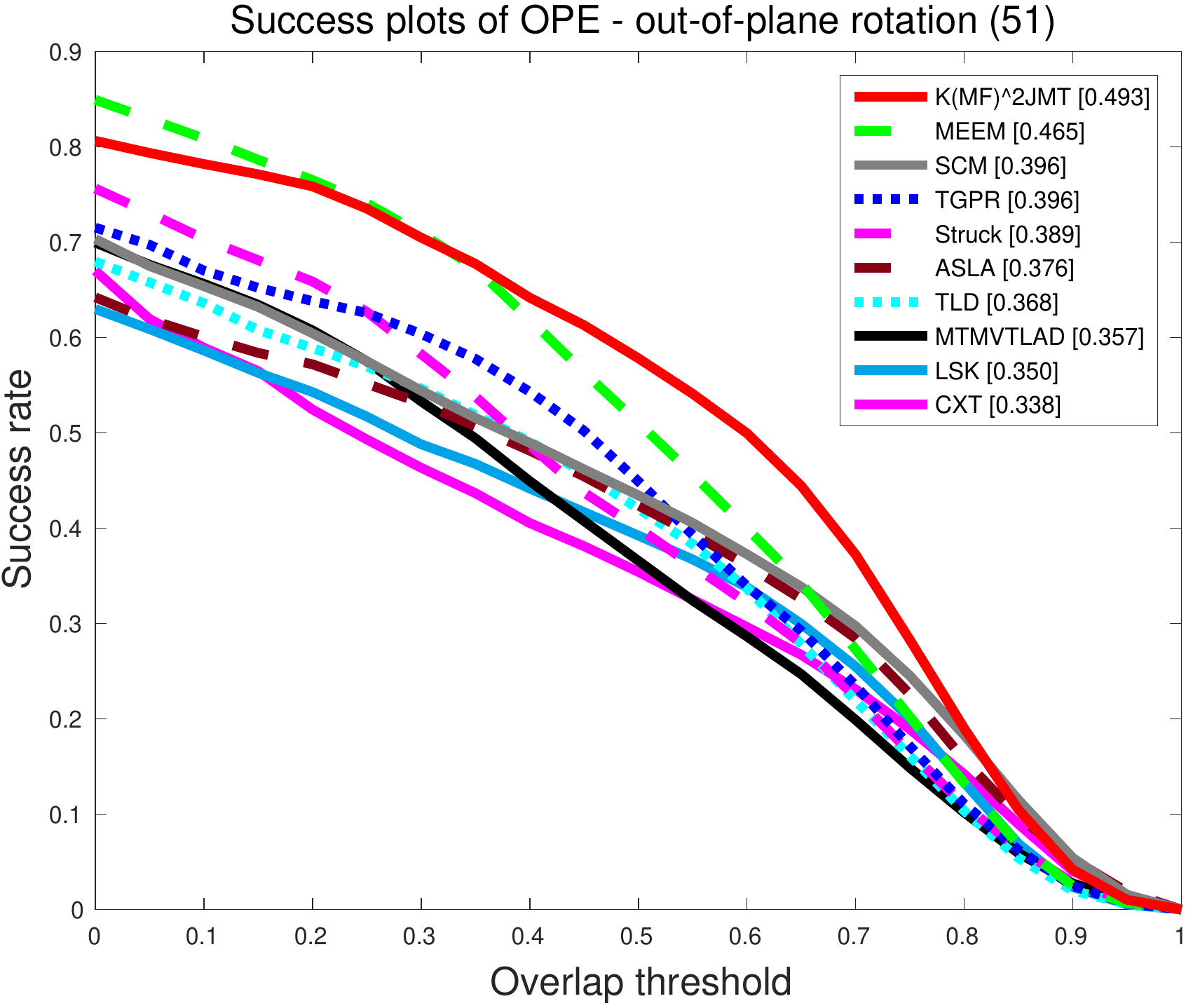}&
\includegraphics[width=.22\textwidth]{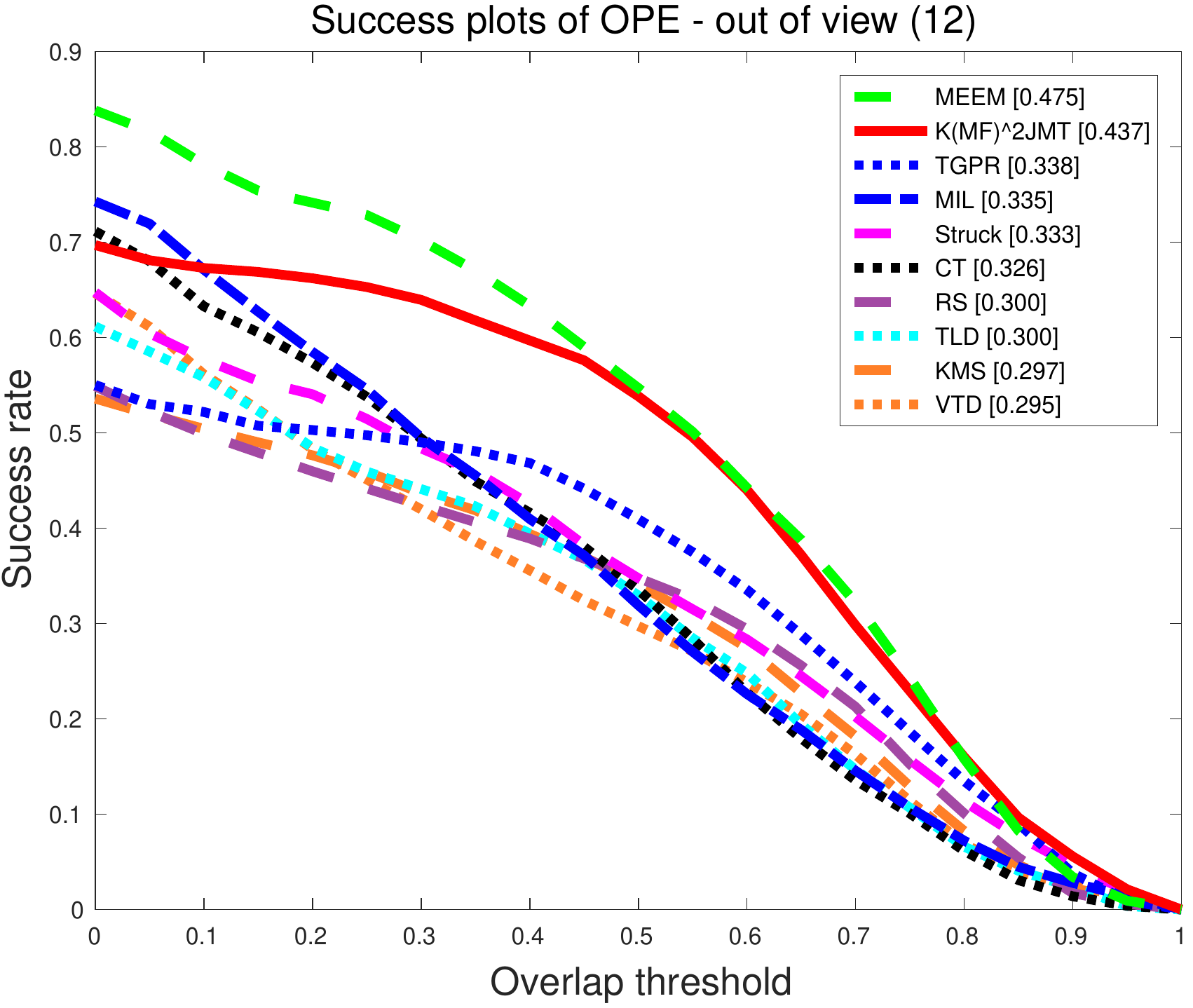} &
\includegraphics[width=.22\textwidth]{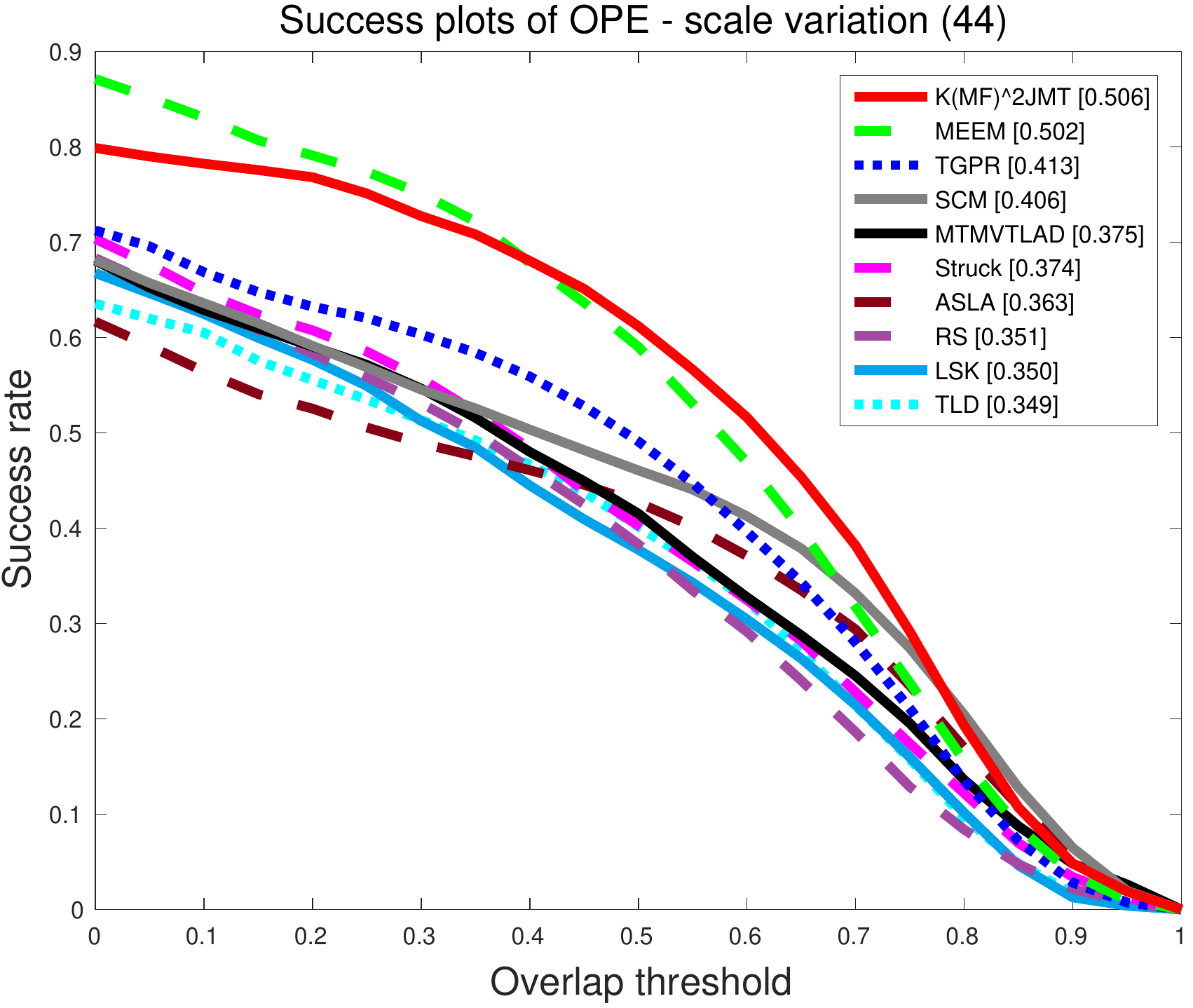}\\
\end{tabular}
\caption{Success plots for each attribute on OTB dataset. The value presented in the title represents the number of videos corresponding to the attributes. The success score is
shown in the legend for each tracker. Our approach provides the best performance on $7$ out of $11$ attributes, namely motion blur, deformation, illumination variation, low resolution, occlusion, out-of-plane rotation, and scale variation.}
\label{OTBB_attributes_success_plot}
\end{figure*}

\subsection{Comparison with state-of-the-art trackers}

In this section, we provide a comprehensive comparison of our proposed $\text{K}(\text{MF})^2\text{JMT}$ with $33$ state-of-the-art trackers on OTB 2015 dataset: 1) $4$ state-of-the-art trackers that do not follow CFT framework yet achieved remarkable performance on OTB 2015, that is MEEM \cite{zhang2014meem}, TGPR \cite{gao2014transfer}, LSST \cite{wang2013least} and MTMVTLAD \cite{mei2015robust}; 2) the $29$ popular trackers provided in \cite{wu2015object}, such as Struck \cite{hare2011struck}, ASLA \cite{jia2012visual}, SCM \cite{zhong2012robust}, TLD \cite{kalal2012tracking}, MTT\cite{zhang2012robust}, VTD \cite{kwon2010visual}, VTS \cite{kwon2011tracking} and MIL \cite{babenko2011robust}. Note that, as a generative tracker, MTMVTLAD has similar motivations as our approach, as it also attempts to integrate multi-task multi-view learning to improve tracking performance. However, different from $\text{K}(\text{MF})^2\text{JMT}$, MTMVTLAD casts tracking as a sparse representation problem in a particle filter framework. Moreover, MTMVTLAD does not reconcile the temporal coherence in consecutive frames explicitly.

The success plot of all the $34$ competing trackers using One Pass Evaluation (OPE) is shown in Fig. \ref{OTBB_comparison plot}(a). For clarity, we only show the top $10$ trackers in this comparison. As can be seen, our proposed $\text{K}(\text{MF})^2\text{JMT}$ achieves overall the best performance, which persistently outperforms the overall second and third best trackers, i.e., MEEM \cite{zhang2014meem} and TGPR \cite{gao2014transfer}. If we look deeper (see a detailed summarization in Table~\ref{lab:detail_difference}), the MEEM tracker, which uses an entropy-minimization-based ensemble learning strategy to avert bad model update, obtains a mean success rate of $60.9\%$. The transfer learning based TGPR tracker achieves a mean success rate of $52.0\%$. By contrast, our approach follows a CFT framework, while using an explicit temporal consistency regularization to enhance robustness. The MTMVTLAD tracker provides a mean success rate of $47.2\%$. Our tracker outperforms MTMVTLAD by $17.1\%$ in mean success rate. Finally, it is worth noting that our approach achieves superior performance while operating at real time, while the mean FPS for MEEM, TGPR and MTMVTLAD are approximately $13.53$, $0.70$ and $0.30$, respectively.

\begin{figure*}[!t]
\centering
\subfigure[Occlusion on \textit{Clifbar}.] {
\begin{tabular}{ccc}
\includegraphics[width=2.5cm,height=2cm]{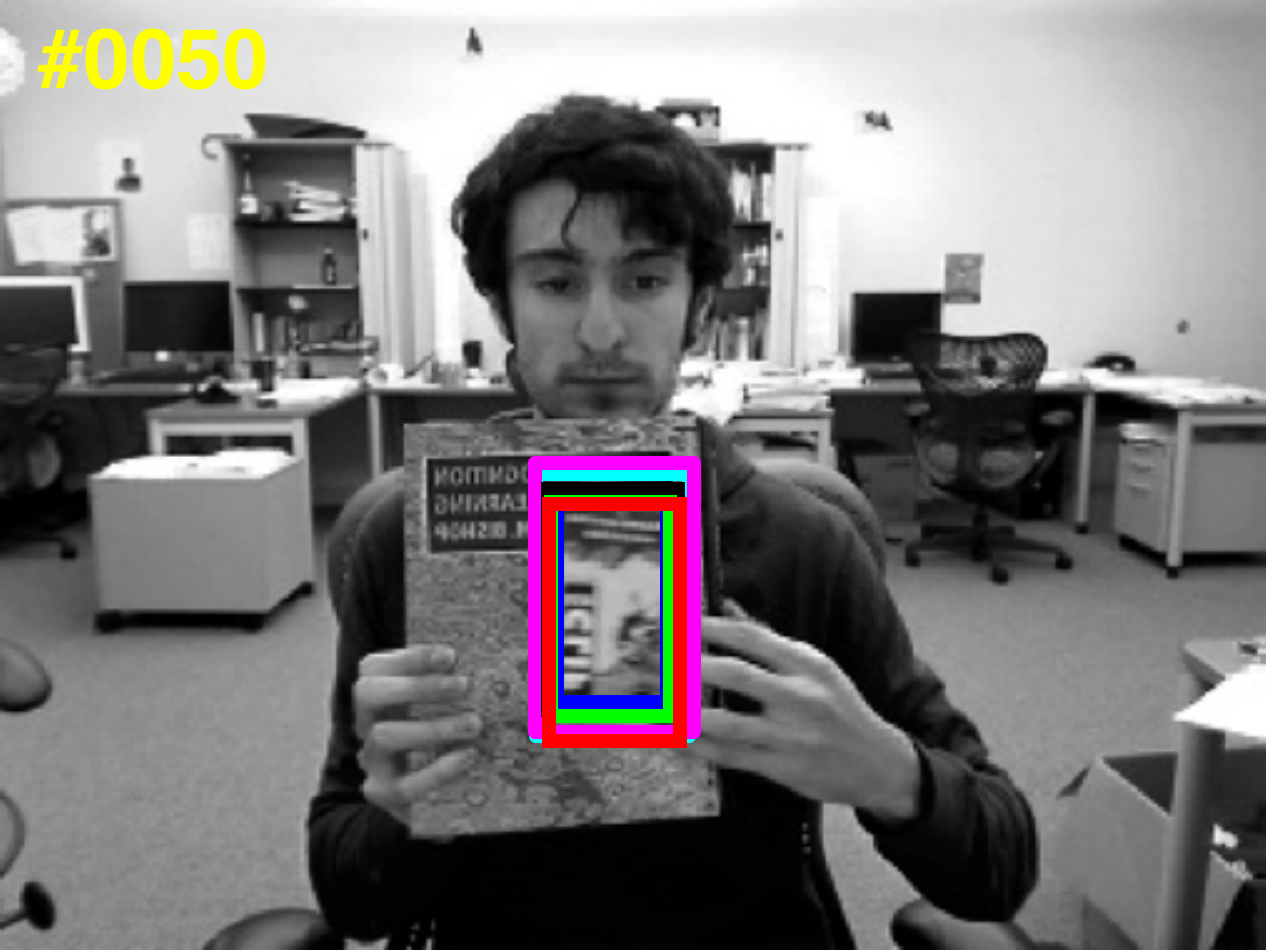}
\includegraphics[width=2.5cm,height=2cm]{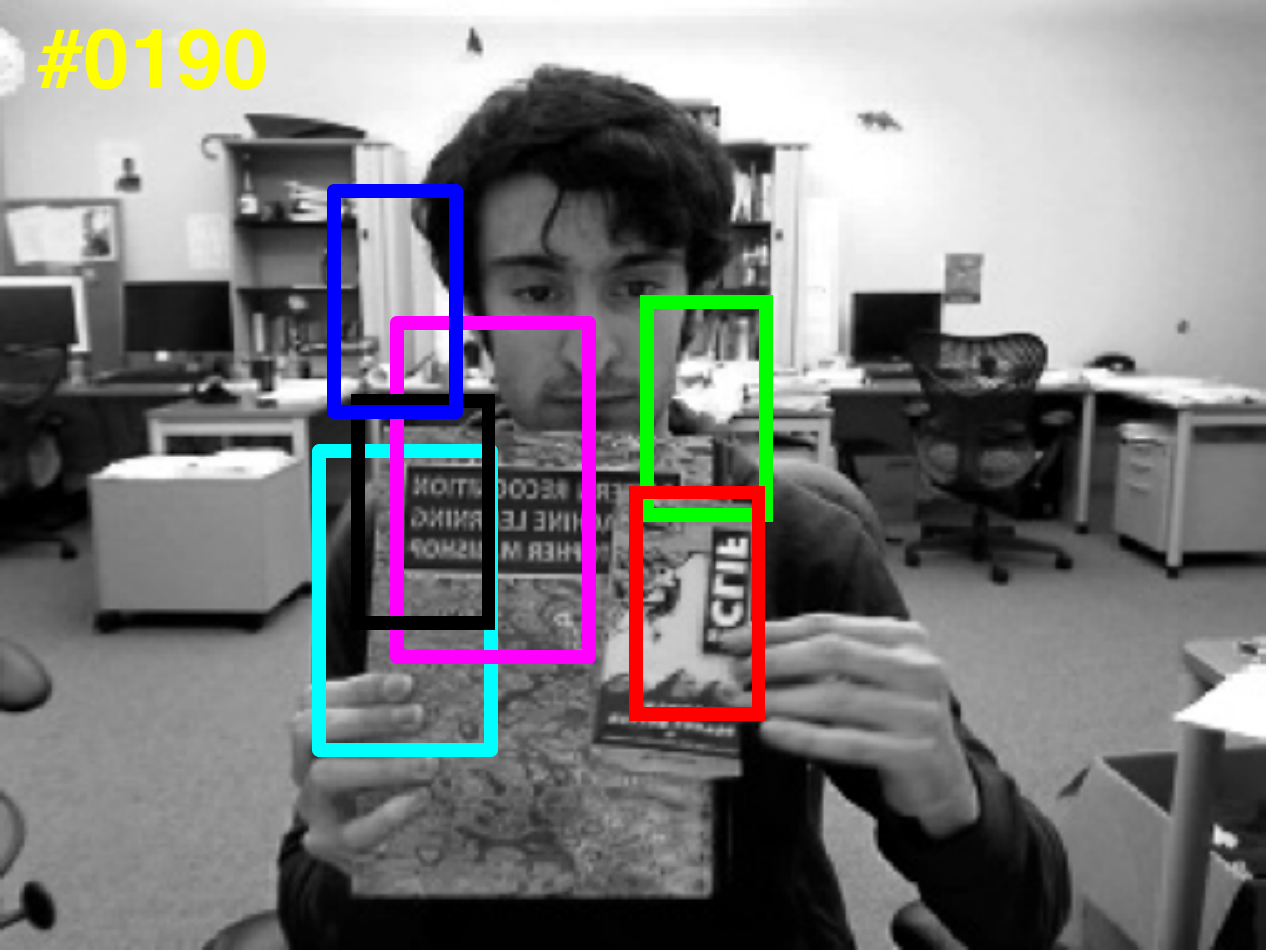}
\includegraphics[width=2.5cm,height=2cm]{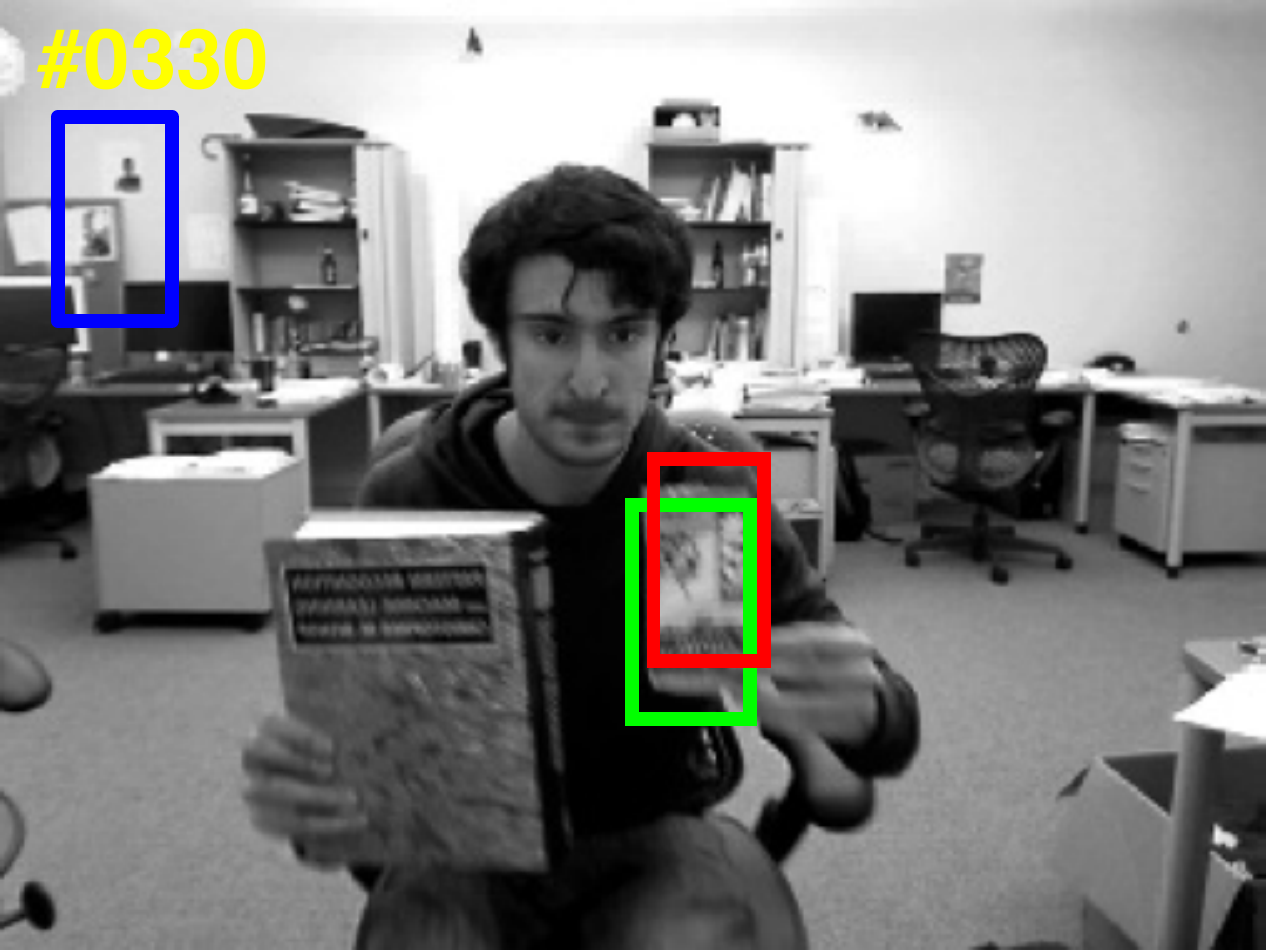}
\end{tabular}
}
\subfigure[Illumination variation on \textit{Skating1}.] {
\begin{tabular}{ccc}
\includegraphics[width=2.5cm,height=2cm]{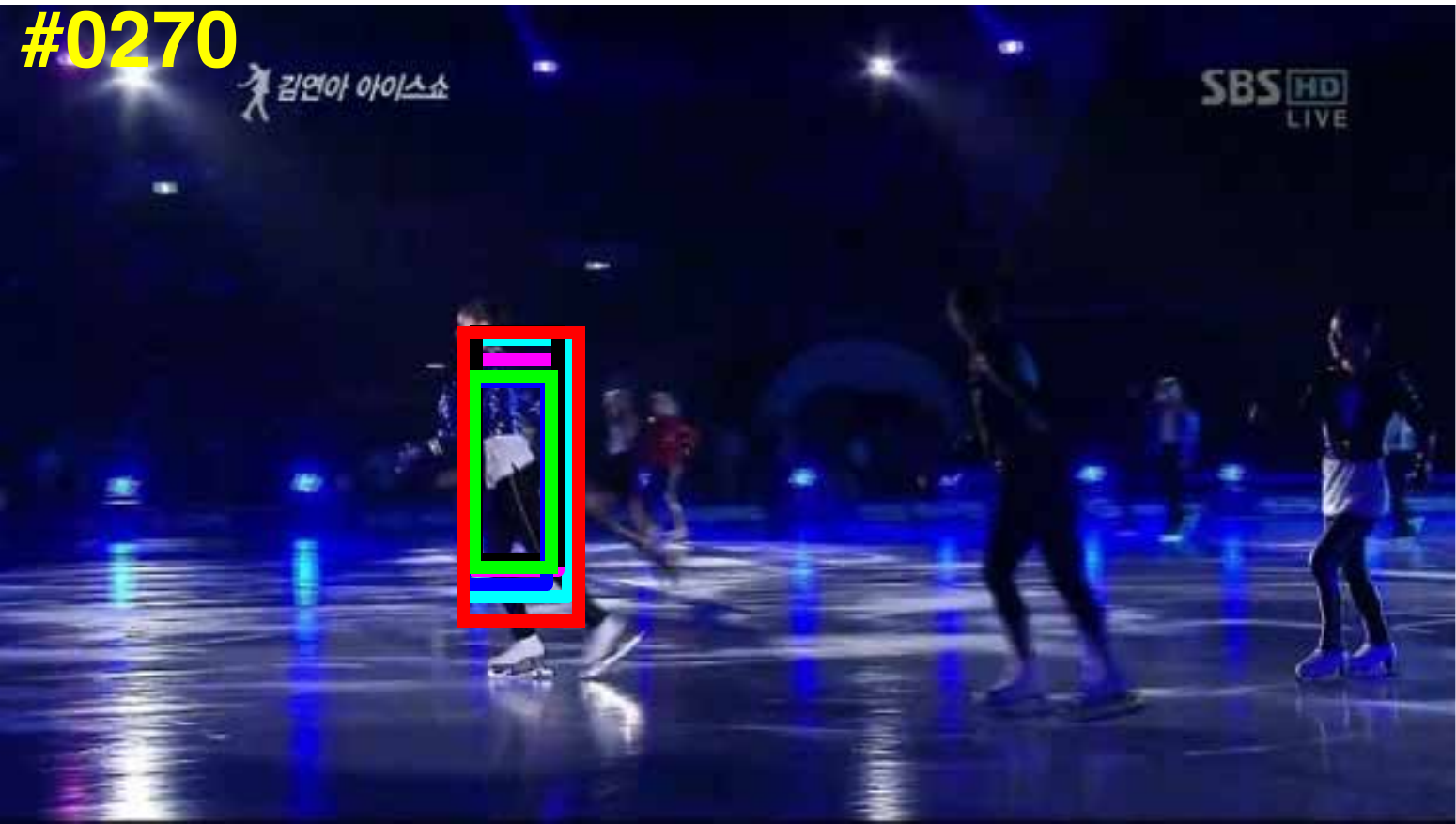}
\includegraphics[width=2.5cm,height=2cm]{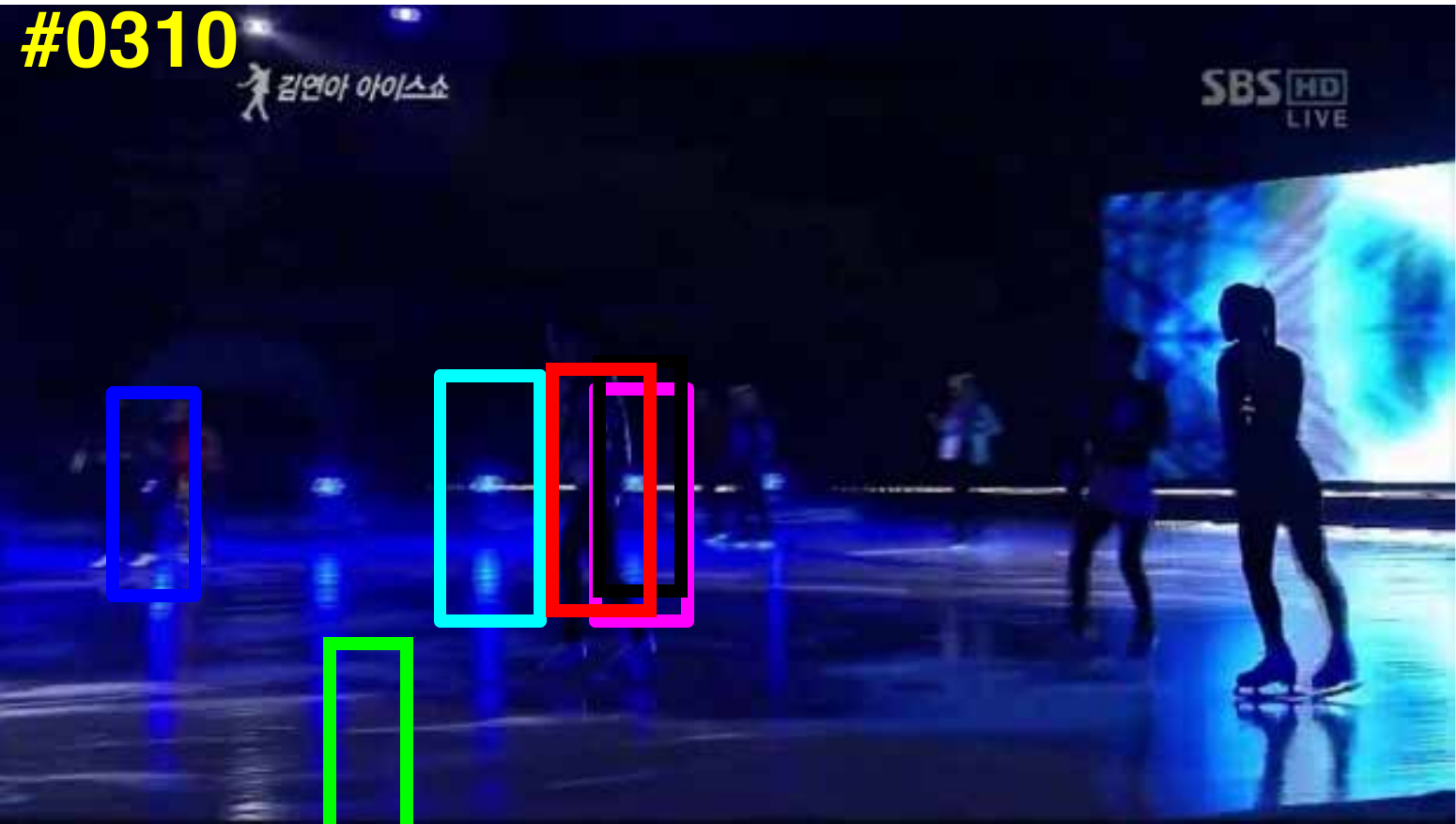}
\includegraphics[width=2.5cm,height=2cm]{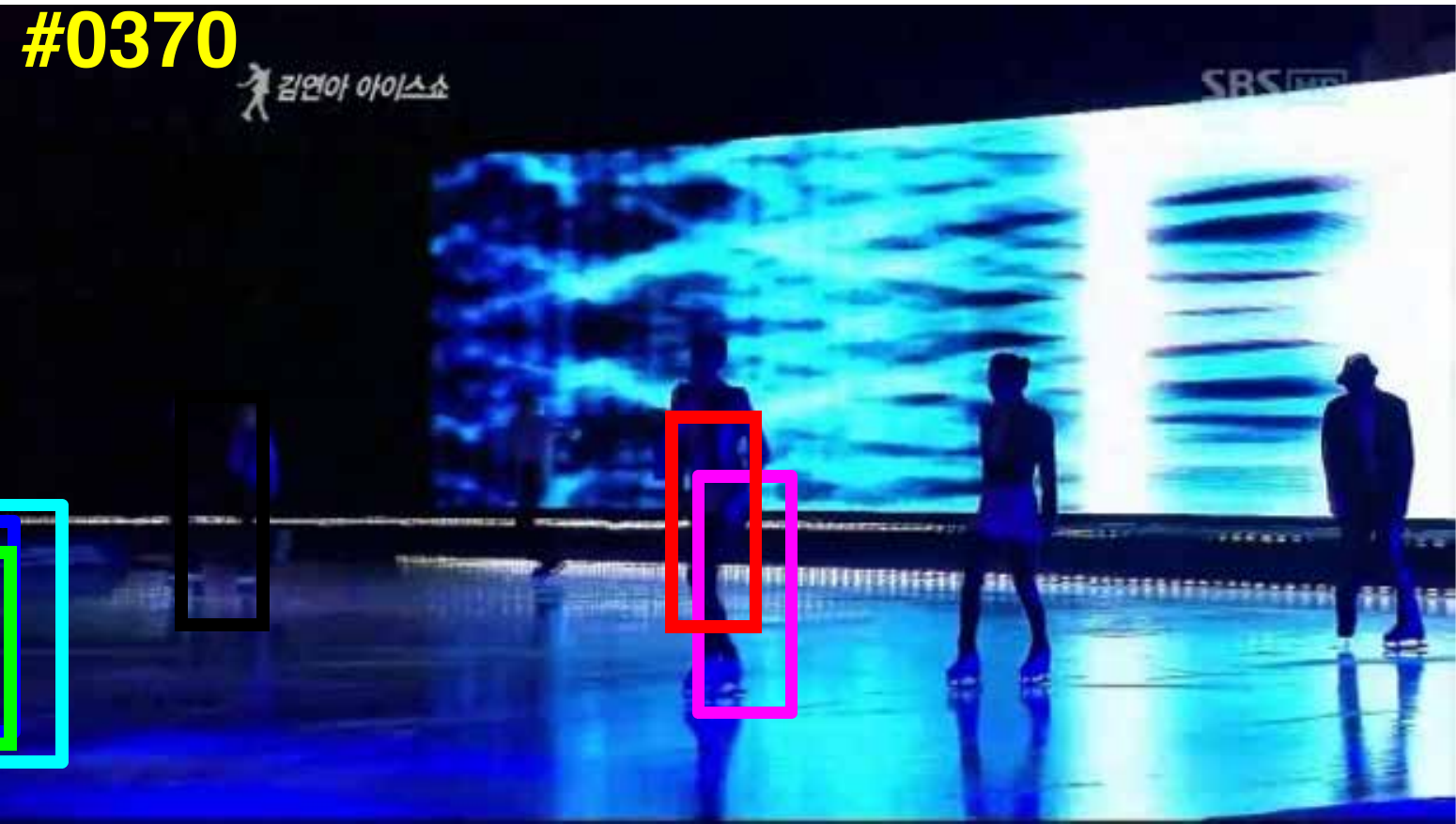}
\end{tabular}
}\\
\subfigure[Scale variation on \textit{Human5}.] {
\begin{tabular}{ccc}
\includegraphics[width=1.5cm,height=2cm]{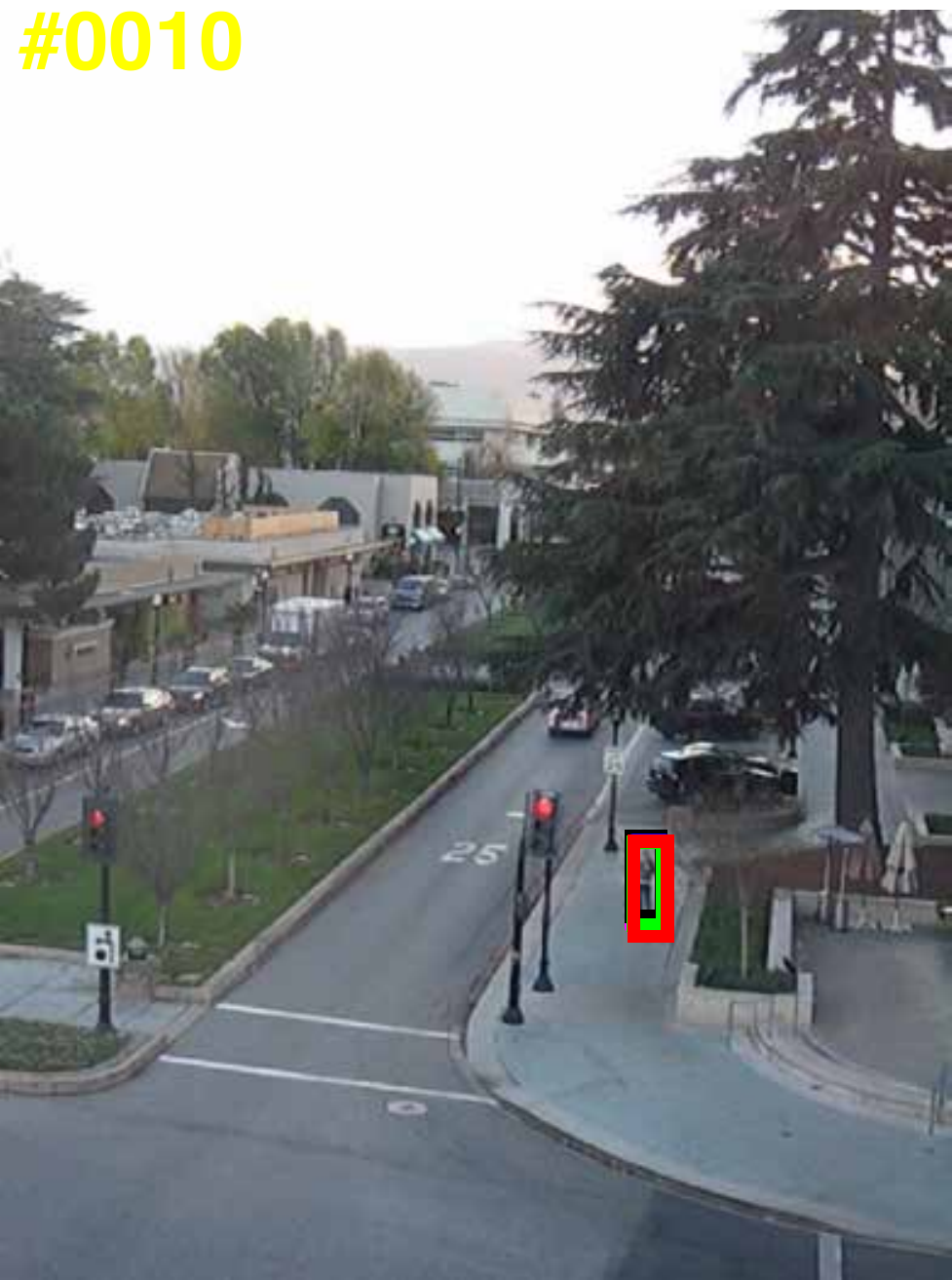}
\includegraphics[width=1.5cm,height=2cm]{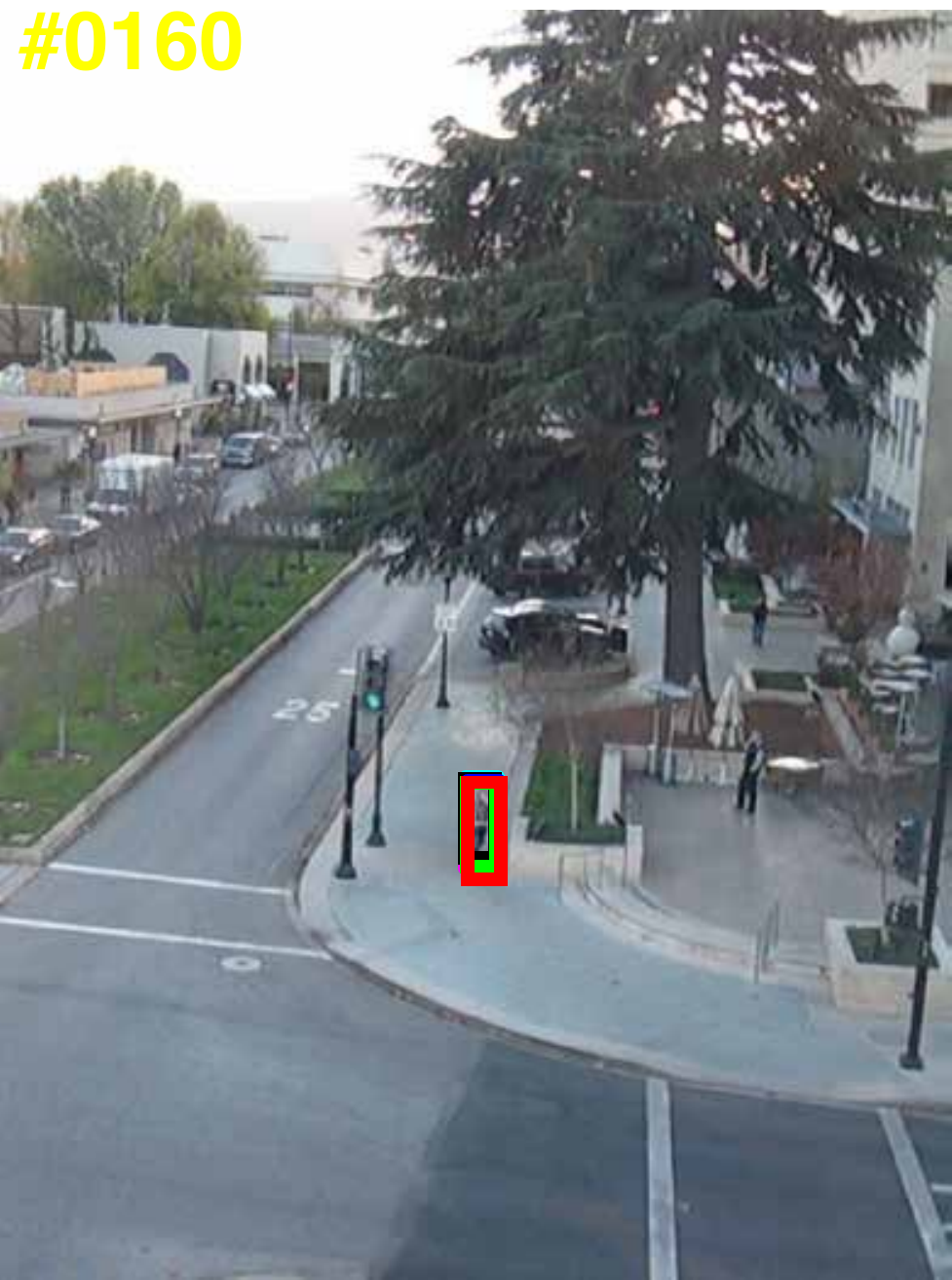}
\includegraphics[width=1.5cm,height=2cm]{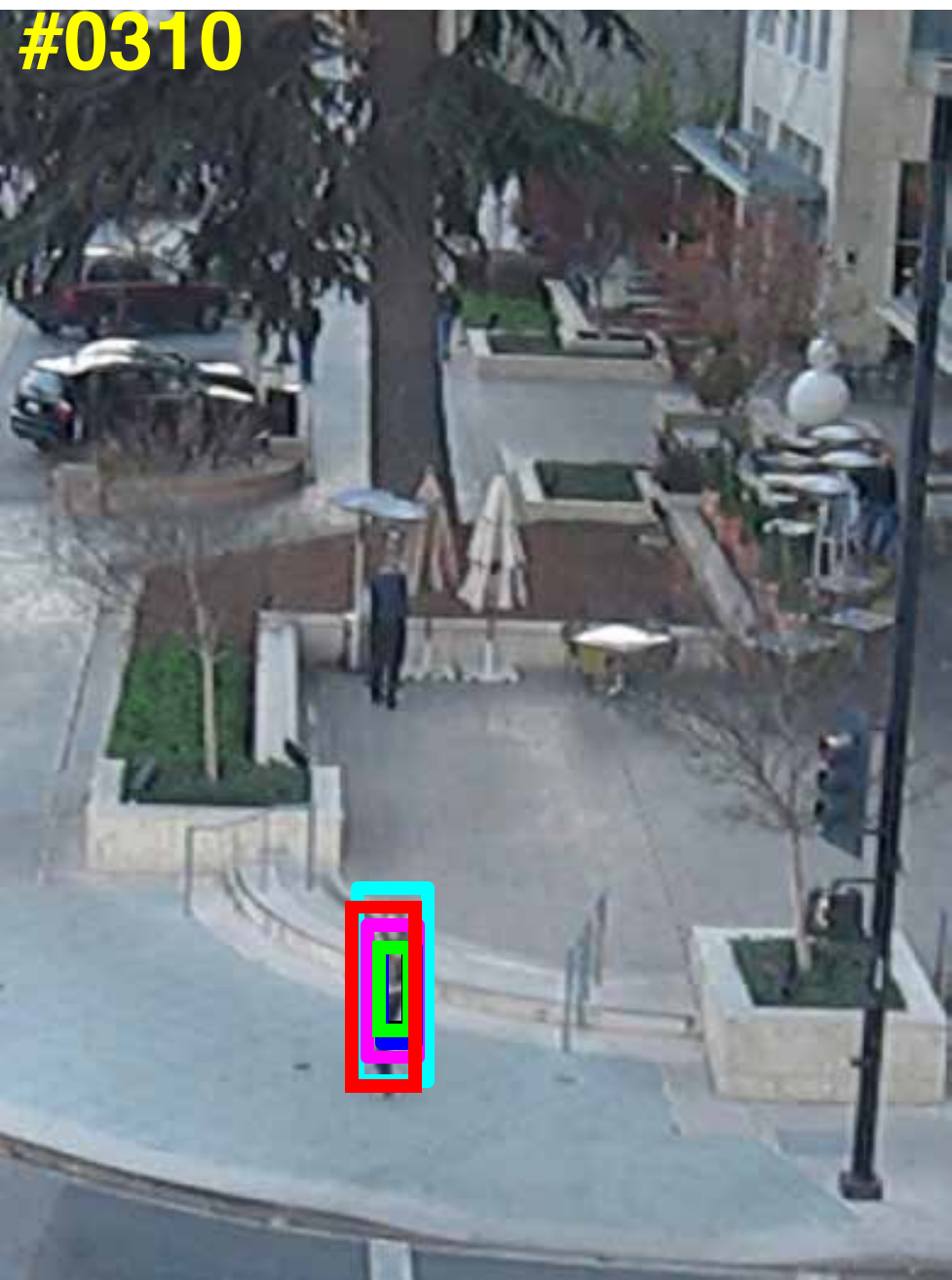}
\includegraphics[width=1.5cm,height=2cm]{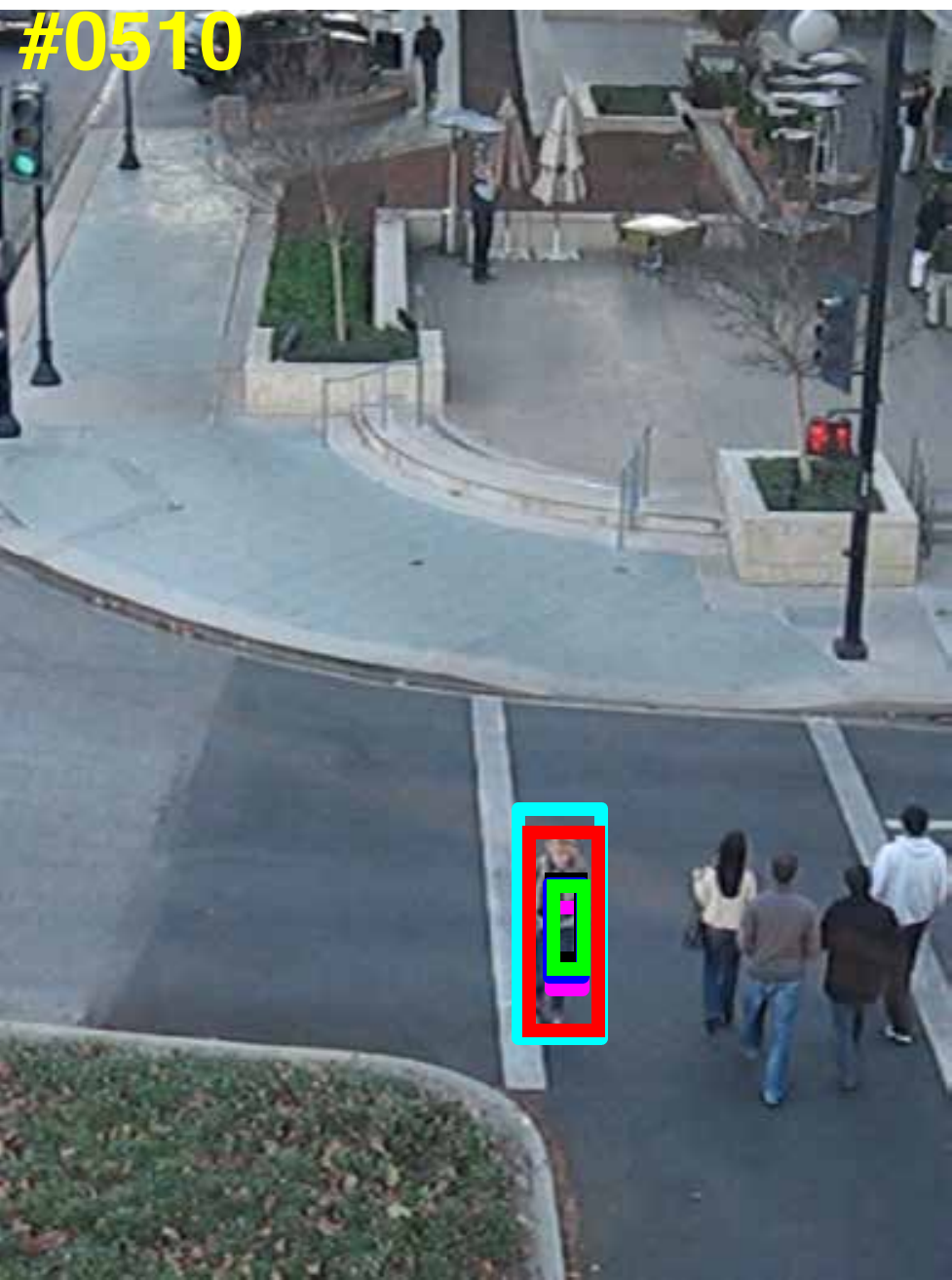}
\includegraphics[width=1.5cm,height=2cm]{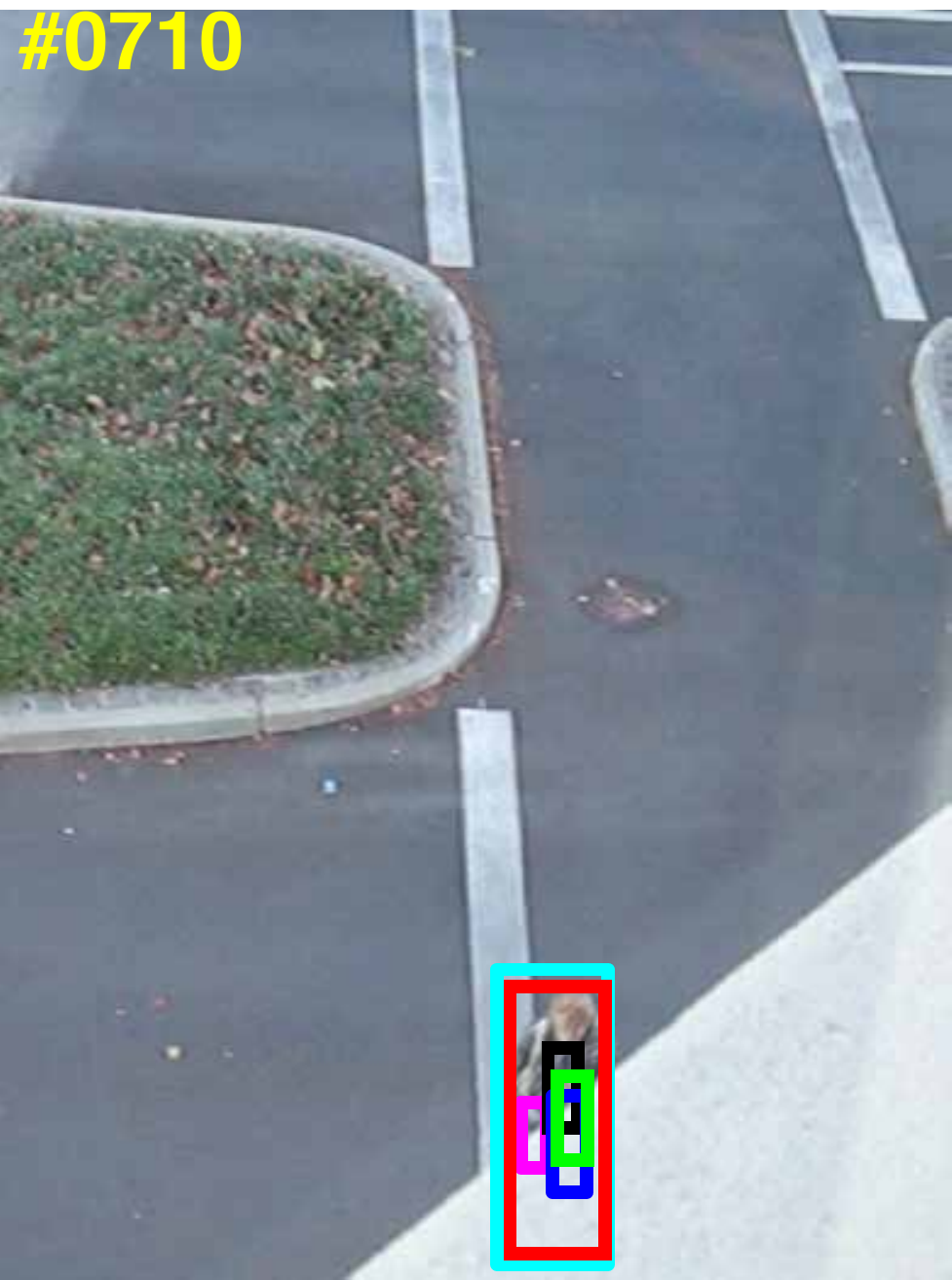}
\end{tabular}
}
\subfigure[Out-of-plane rotation on \textit{Tiger2}.] {
\begin{tabular}{ccc}
\includegraphics[width=2.5cm,height=2cm]{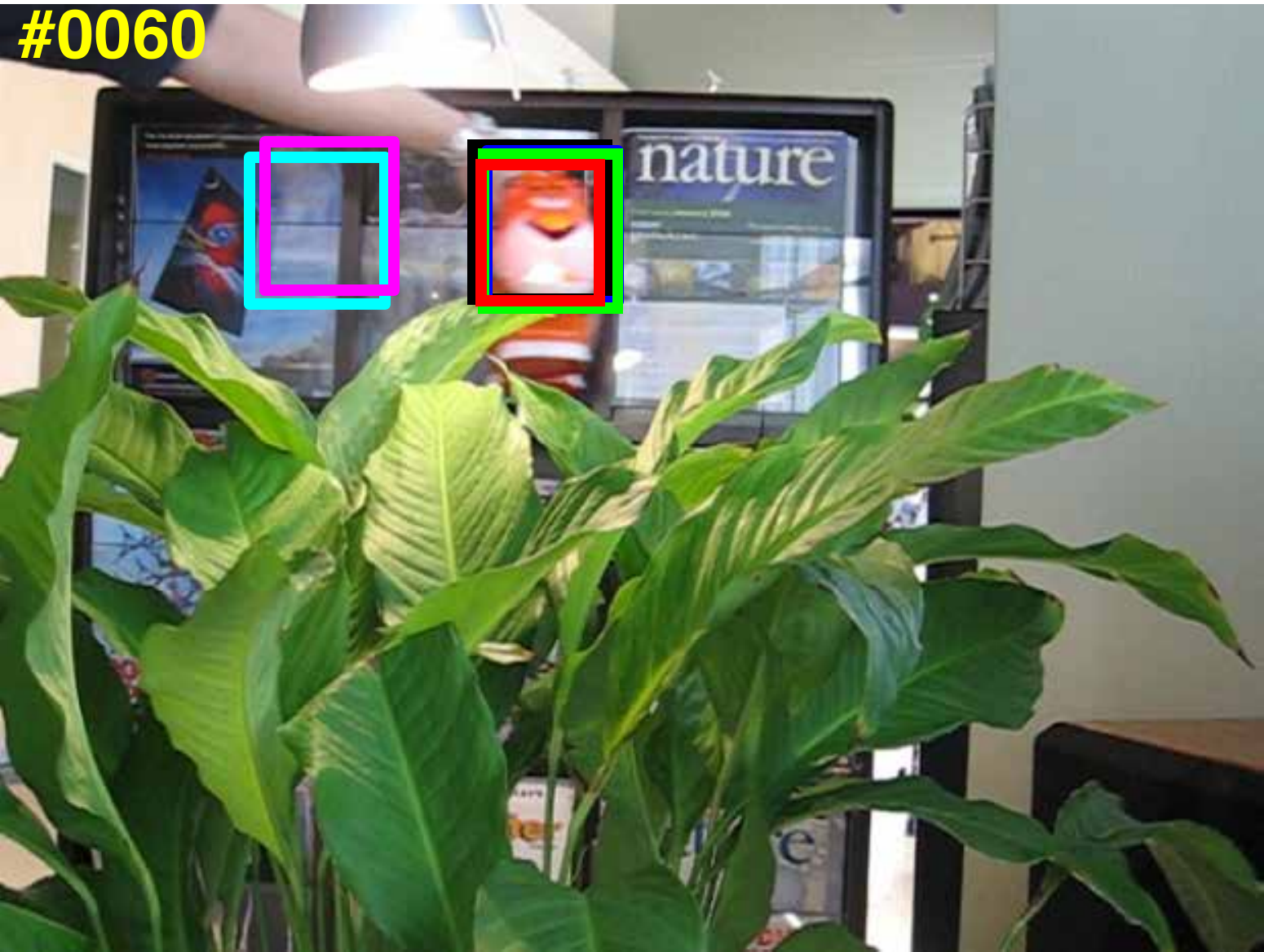}
\includegraphics[width=2.5cm,height=2cm]{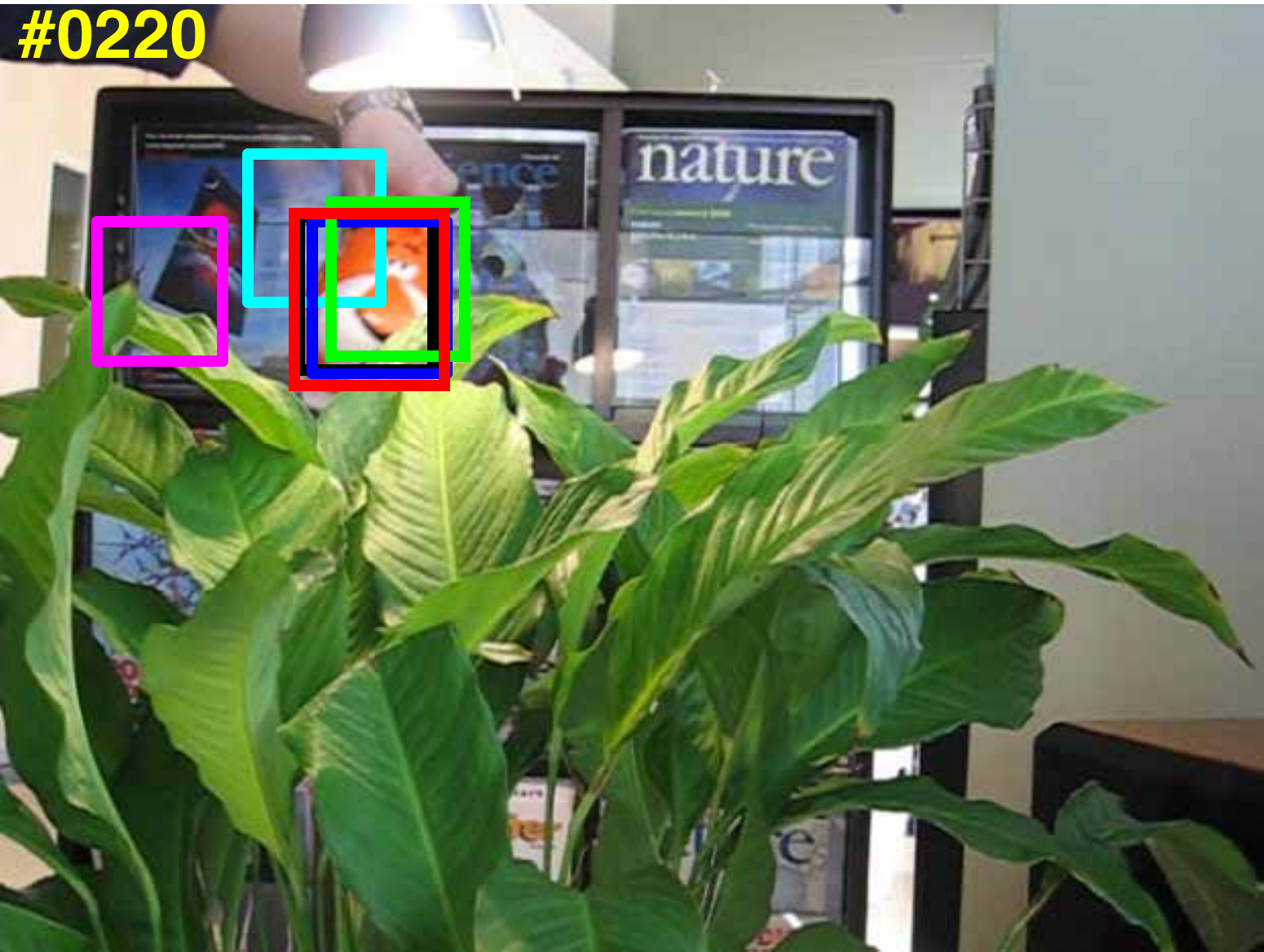}
\includegraphics[width=2.5cm,height=2cm]{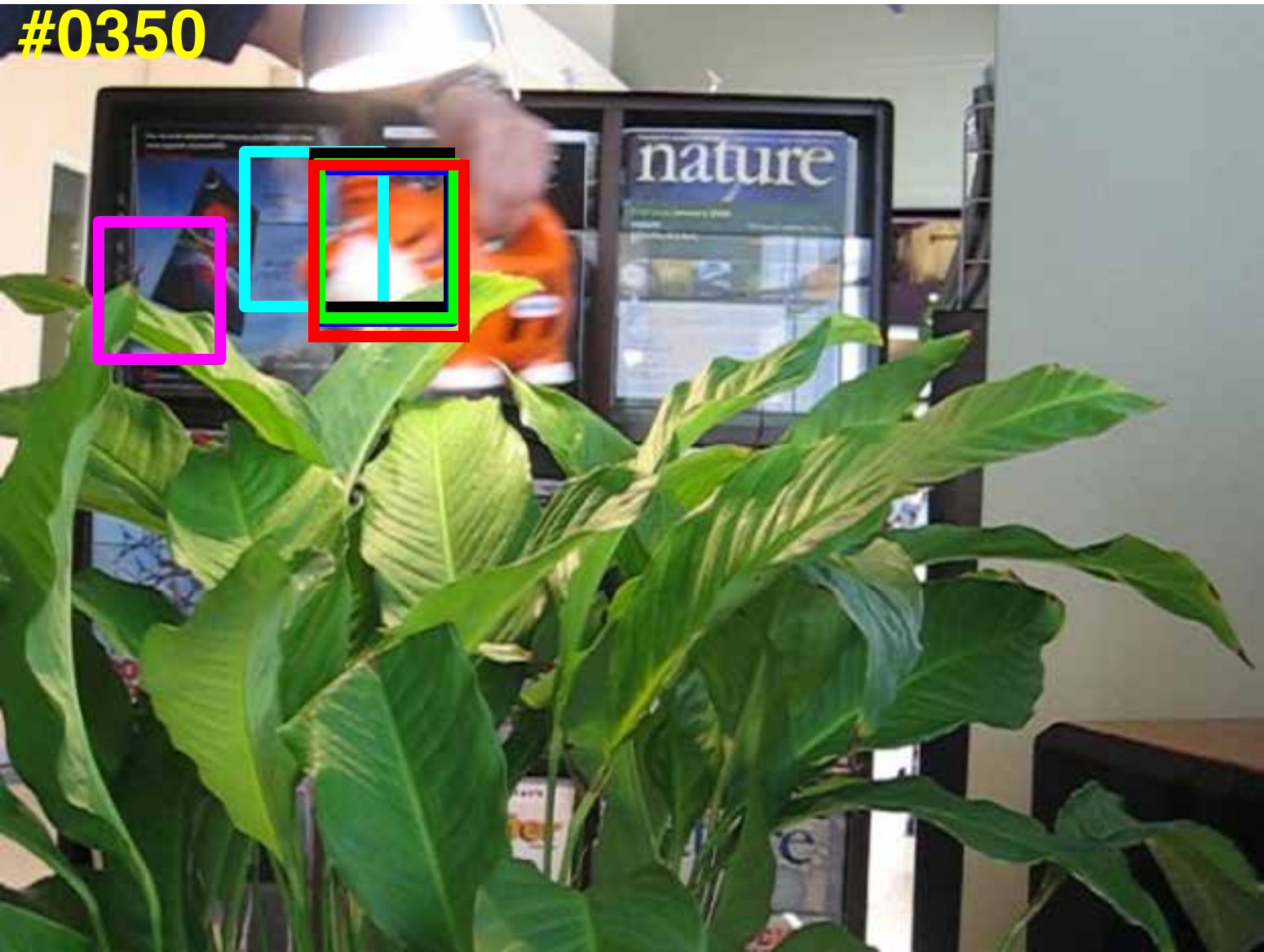}
\end{tabular}
}\\
\subfigure[Tracker legend] {\includegraphics[width=0.48\textwidth]{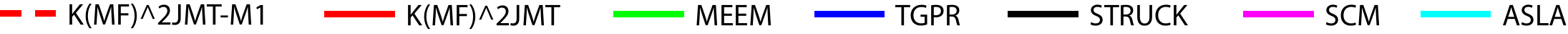}}
\caption{A qualitative comparison of our method with five state-of-the-art trackers. Tracking results are shown on four example videos from the OTB 2015 dataset. The videos show challenging situations, such as (a) occlusion, (b) illumination variation, (c) scale variations, and (d) out-of-plane rotation. Our approach offers superior performance compared to the existing trackers in these challenging situations. (e) shows tracker legend.}
\label{visual_comparison}
\end{figure*}

We also report the tracking results using Temporal Robustness Evaluation (TRE) and Spatial Robustness Evaluation (SRE). For TRE, it runs trackers on $20$ sub-sequences segmented from the original sequence with different lengths, and SRE evaluates trackers by initializing them with slightly shifted or scaled ground truth bounding boxes. With TRE and SRE, the robustness of each evaluated trackers can be comprehensively interpreted. The SRE and TRE evaluations are shown in Fig. \ref{OTBB_comparison plot}(b) and Fig. \ref{OTBB_comparison plot}(c) respectively. In both evaluations, our approach provides a constant performance gain over the majority of existing methods. The MEEM achieves better robustness than our approach. A possible reason is that MEEM combines the estimates of an ensemble of experts to mitigate inaccurate predictions or sudden drifts, so that the weaknesses of the trackers are reciprocally compensated. We also perform an attribute based analysis on our tracker. In OTB, each sequence is annotated with $11$ different attributes, namely: fast motion, background clutter, motion blur, deformation, illumination variation, in-plane rotation, low resolution, occlusion, out-of-plane rotation, out-of-view and scale variation. It is interesting to find that $\text{K}(\text{MF})^2\text{JMT}$ ranks the first on $7$ out of $11$ attributes (see Fig. \ref{OTBB_attributes_success_plot}), especially on illumination variation, occlusion, out-of-plane rotation and scale variation. This verifies the superiority of $\text{K}(\text{MF})^2\text{JMT}$ on target appearance representation and its capability on discovering reliable coherence from short-term memory. However, the overwhelming advantage no longer exists for fast motion and out-of-view. This is because the temporal consistency among consecutive frames becomes weaker in these two scenarios. The qualitative comparison shown in Fig.~\ref{visual_comparison} corroborates quantitative evaluation results. It is worth noting that, we strictly follow the protocol provided in~\cite{wu2015object} and use the same parameters for all sequences.

\begin{table}
\centering
\caption{A detailed comparison of our $\text{K}(\text{MF})^2\text{JMT}$ with MEEM \cite{zhang2014meem}, TGPR \cite{gao2014transfer} and MTMVTLAD \cite{mei2015robust}. The mean overlap precision (OP) score ($\%$) at threshold $0.5$ over all the $77$ color videos in the OTB dataset are presented.}\label{lab:detail_difference}
\begin{tabular}{ccccc}\hline
 & mean success rate & FPS \\\hline
$\text{MEEM}$ & $60.9$ & $13.53$ \\
$\text{TGPR}$ & $52.0$ & $0.70$  \\
$\text{MTMVTLAD}$ & $47.2$ & $0.30$ \\
$\text{K}(\text{MF})^2\text{JMT}$ & $64.3$ & $30.46$ \\\hline
\end{tabular}
\end{table}

\subsection{VOT 2015 Challenge}
In this section, we present results on the VOT $2015$ challenge. We compare our proposed $\text{K}(\text{MF})^2\text{JMT}$ with $62$ participating trackers in this challenge. For a fair comparison, the DSST \cite{danelljan2014accurate} is substituted with its fast version (i.e., fDSST \cite{danelljan2016discriminative}) raised by the same authors, as fDSST demonstrates superior performance than DSST as shown in~\cite{danelljan2016discriminative}. %In addition to the baseline experiment, we evaluate the robustness by performing the region noise experiment in the VOT protocol. This is performed by introducing noise in the initial bounding boxes and evaluating the tracker multiple times for each video. For more details about this experiment, we refer to \cite{kristan2014visual}.

\begin{figure}[!t]
\centering
\subfigure[AR rank (sequence pooling)] {\includegraphics[width=.24\textwidth]{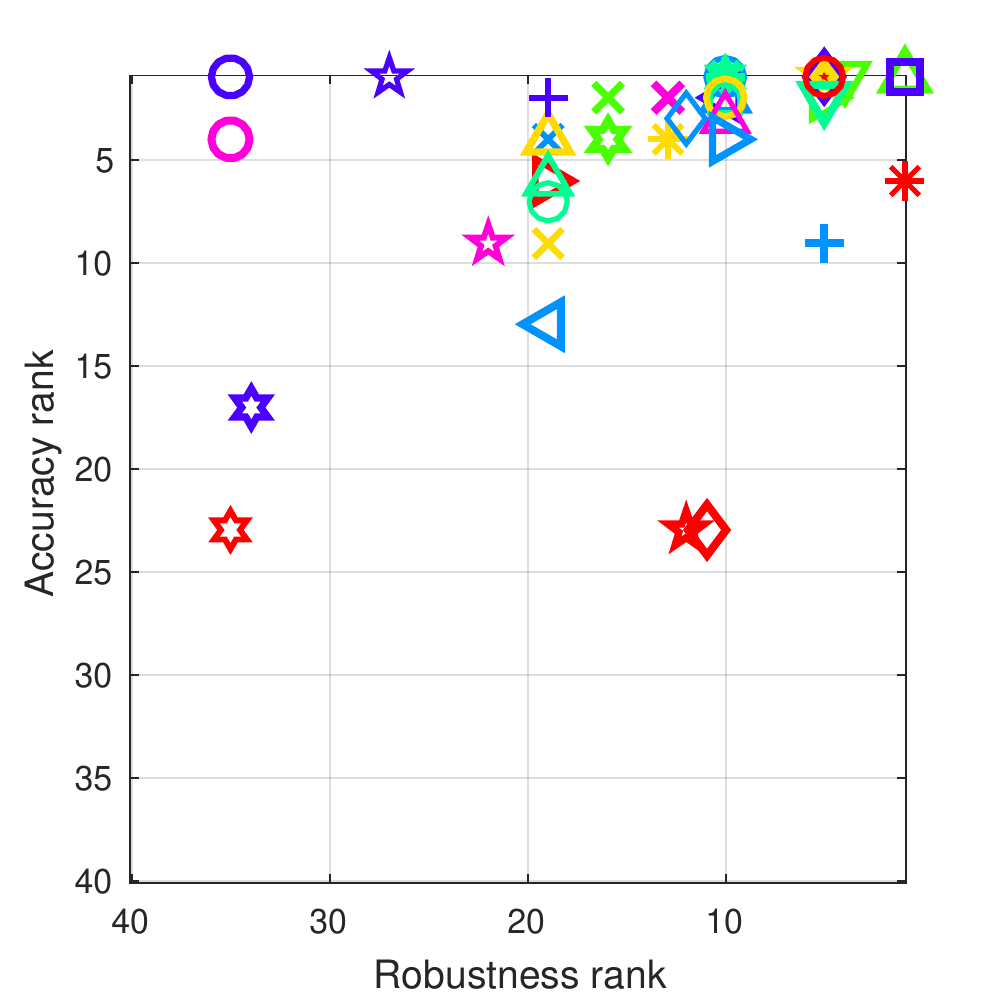}}
\subfigure[AR rank (attribute normalization)] {\includegraphics[width=.24\textwidth]{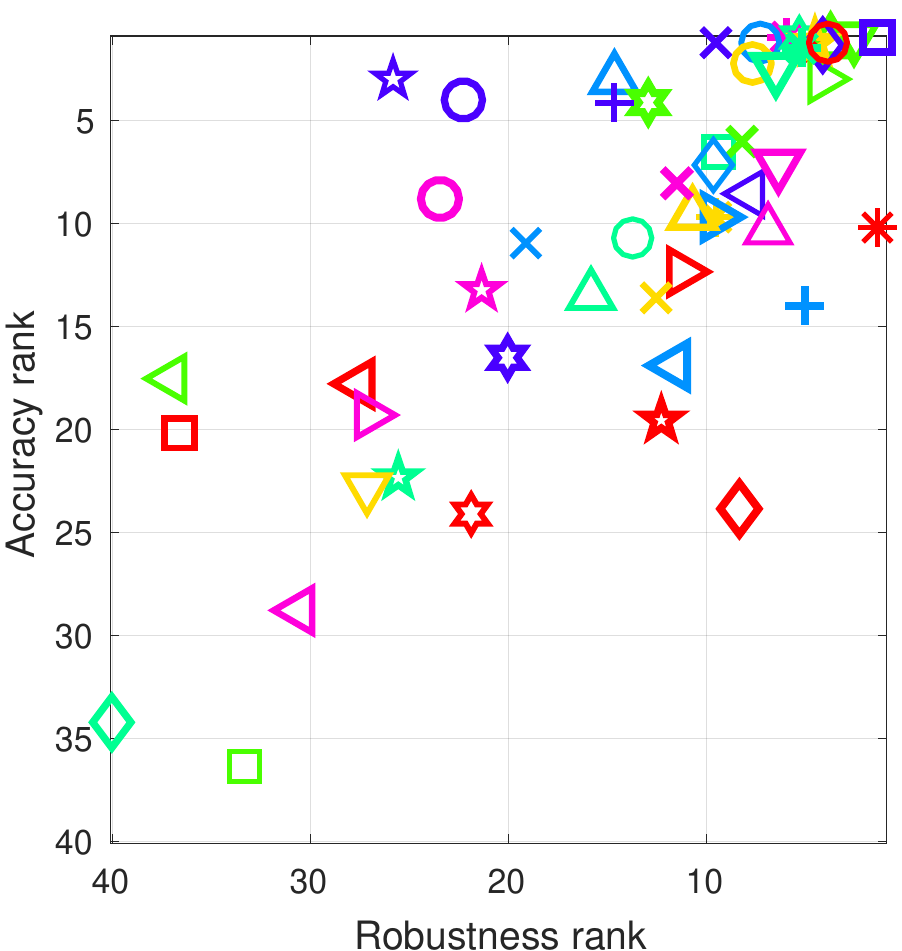}}\\
\subfigure[Zoomed-in AR rank (sequence pooling)] {\includegraphics[width=.24\textwidth]{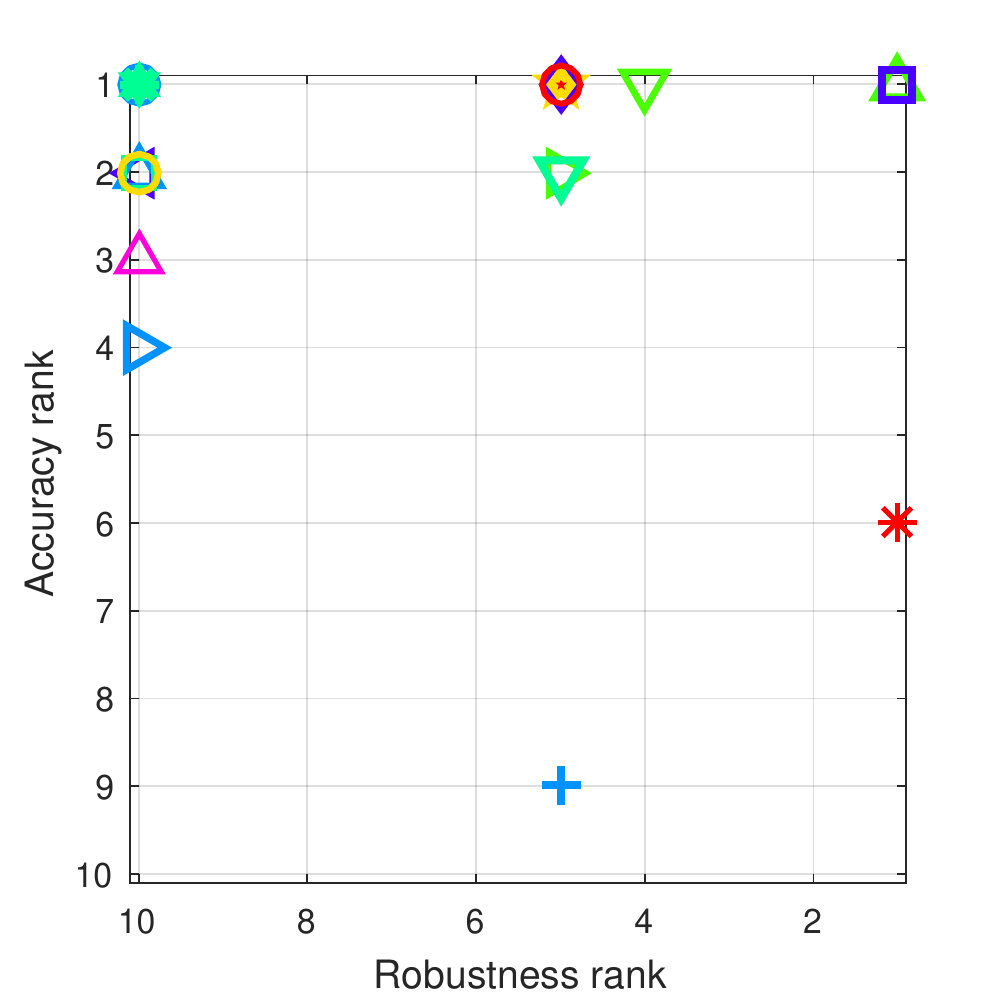}}
\subfigure[Zoomed-in AR rank (attribute normalization)] {\includegraphics[width=.24\textwidth]{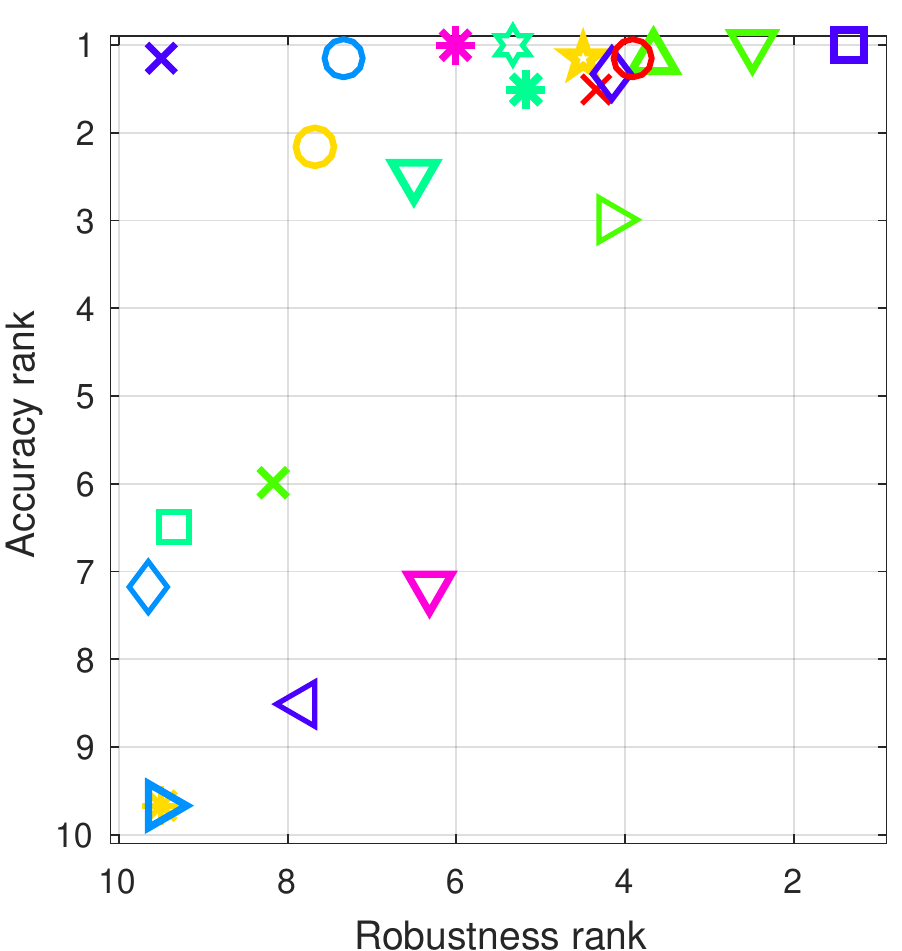}}\\
\subfigure[Tracker legend] {\includegraphics[width=0.48\textwidth]{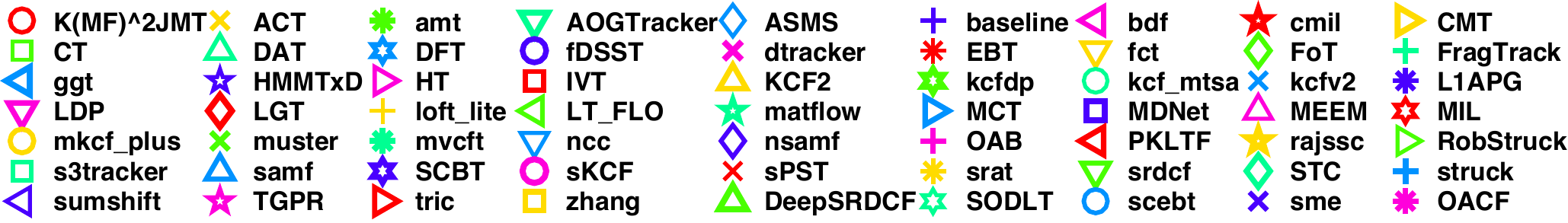}}
\caption{The accuracy and robustness (AR) rank plots generated by (a) sequence pooling and by (b) attribute normalization in the VOT 2015 dataset. The accuracy and robustness rank are plotted along the vertical and horizontal axis respectively. (c) and (d) demonstrate the zoomed-in figure of (a) and (b) respectively, in which only top-10 accuracy and robustness ranks are plotted. (e) shows the tracker legend. Our proposed $\text{K}(\text{MF})^2\text{JMT}$ (denoted by the red circle) achieves top-6 performance in terms of both accuracy and robustness among $63$ competitors in both experiments.}
\label{ranking_plots_VOT2015}
\end{figure}

%Table \ref{lab:VOT2014} presents the final ranking score over all the $25$ videos in the VOT $2014$ dataset. Apart from the final average rank, we also demonstrate partial results for the baseline and region noise experiments. Obviously, $\text{K}(\text{MF})^2\text{JMT}$ achieves the top average rank among all the $39$ trackers.

\begin{figure}[!t]
\centering
\includegraphics[width=.48\textwidth]{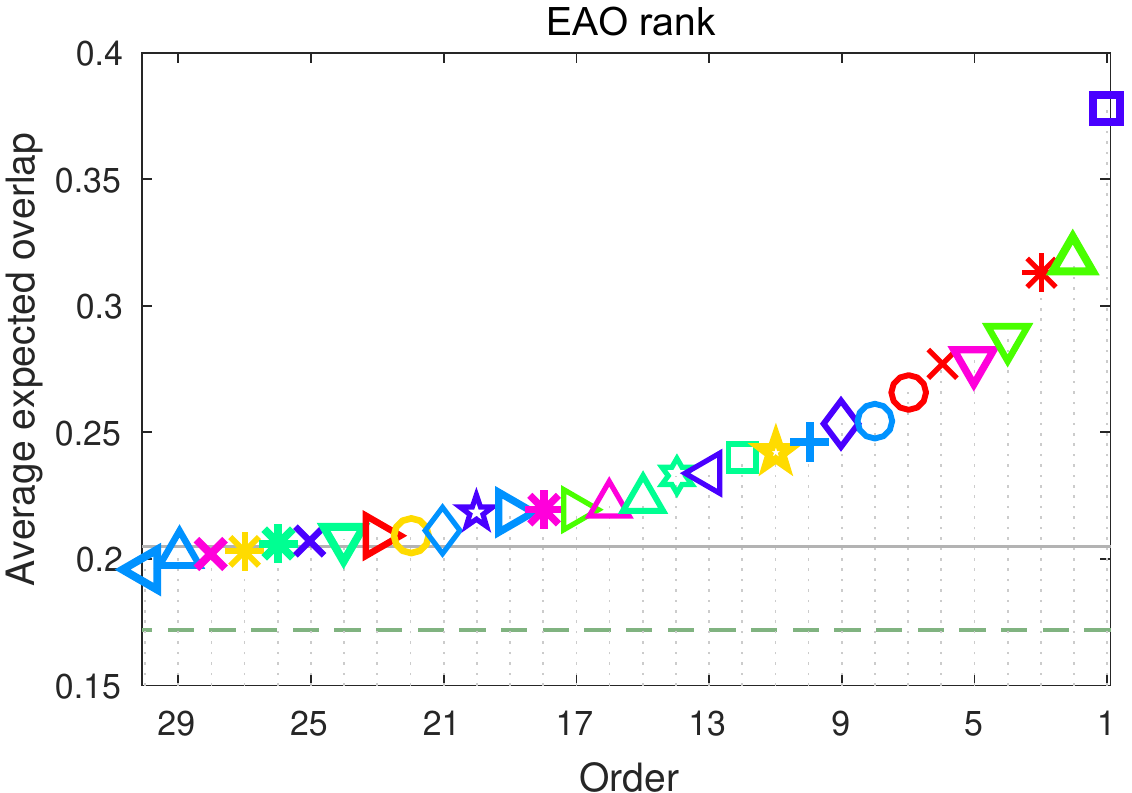}
\caption{The expected average overlap (EAO) graphs with trackers ranked from right to left. The right-most tracker is the top-performing in terms of EAO values. The grey solid line denotes the average performance of trackers published at ICCV, ECCV, CVPR, ICML or BMVC in $2014$/$2015$ (nine papers from 2015 and six from 2014), beyond which a tracker can be thought of as state of the art~\cite{kristan2015visual}. The green dashed line denotes the performance of VOT $2014$ winner (i.e., fDSST~\cite{danelljan2016discriminative}). See Fig.~\ref{ranking_plots_VOT2015} for legend.}
\label{EAO_plots_VOT2015}
\end{figure}

In the VOT $2015$ benchmark, each video is annotated by five different attributes: camera motion, illumination change, occlusion, size change and motion change. Different from OTB 2015 dataset, the attributes in VOT 2015 are annotated per-frame in a video.
Fig. \ref{ranking_plots_VOT2015} shows the accuracy and robustness (AR) rank plots generated by sequence pooling and attribute normalization. The pooled AR plots are generated by concatenating the experimental results on all sequences to directly obtain a rank list, while the attribute normalized AR rank plots are obtained based on the average of ranks achieved on these individual attributes. Fig.~\ref{EAO_plots_VOT2015} shows the EAO ranks. Only results for the top-$20$ trackers in the VOT challenge are reported for clarity.

It is easy to summarize some key observations from these figures:\\
1) The CFTs (including our tracker $\text{K}(\text{MF})^2\text{JMT}$, DeepSRDCF~\cite{danelljan2015convolutional}, SRDCF~\cite{danelljan2015learning}, RAJSSC~\cite{zhang2015joint}, NSAMF~\cite{kristan2015visual}, SAMF~\cite{li2014scale}) account for majority of the top-performing trackers. By fully exploiting the representation power of CNNs or rotating candidate samples to augment candidate set, MDNet~\cite{nam2016learning} and sPST~\cite{hua2015online} also demonstrate superior (or even the best) performance. This result indicates that a well-designed discriminative tracker with conventional tracking-by-detection scheme and random sampling in the detection phase can achieve almost the same tracking accuracy with state-of-the-art CFTs coupled with dense sampling, at the cost of high computational burden (as officially reported in~\cite{kristan2015visual}).\\
2) Among top-performing CFTs, $\text{K}(\text{MF})^2\text{JMT}$ tied for the first place with DeepSRDCF, SRDCF, RAJSSC and NSAMF in terms of tracking accuracy. However, the robustness of $\text{K}(\text{MF})^2\text{JMT}$ is inferior to DeepSRDCF and SRDCF. Both DeepSRDCF and SRDCF introduce a spatial regularization penalty term to circumvent boundary effects caused by conventional circular correlation, thus significantly mitigating inaccurate training samples and restricted searching regions. However, one should note that these two trackers can hardly run in real time. A thorough investigation between $\text{K}(\text{MF})^2\text{JMT}$ and DeepSRDCF is demonstrated in Section~\ref{experiments_section3}.\\
3) Our tracker outperforms RAJSSC, NSAMF and SAMF in terms of robustness and EAO values. All these trackers use HOGs and color information to describe target appearance under a scale-adaptive framework. The performance difference indicates that our $\text{K}(\text{MF})^2\text{JMT}$ enjoys an advanced feature fusion scheme and the integration of multiple frames is beneficial to performance gain.

Apart from these three observations, there are other interesting points. For example, the NSAMF achieves a large performance gain compared with SAMF. The main difference is that NSAMF substitutes the color name with color probability. On the other hand, the EBT~\cite{zhu2015tracking} reaches a fairly high robustness and EAO value. However, its tracking accuracy is desperately poor. One possible reason is that the adopted contour information does not have desirable adaptability to target scale and aspect ratio changes.

%Our approach performs favorably both in terms of accuracy and robustness compared to the other trackers. Contrary to \cite{danelljan2016discriminative}, fDSST provides inferior performance to SAMF herein. The second best method, SAMF \cite{li2014scale}, is a baseline CFT that uses multiple feature cues and applies a multi-resolution strategy for estimating the target scale. Although both $\text{K}(\text{MF})^2\text{JMT}$ and SAMF select HOGs and Color names as feature cues and explicitly consider the impact of scale estimation, the modeling methods employed are totally different. In particular, SAMF employs a naive feature concatenating scheme combined with a multi-resolution searching strategy to estimate the target scale, whereas $\text{K}(\text{MF})^2\text{JMT}$ jointly models different features in a MVL module and a simpler yet more effective scale estimation method. Moreover, SAMF discards all the temporal information which has been proven effective in visual tracking.

%We also report the average overlap and failure rate over all videos in the baseline experiment. Table \ref{lab:VOT2014_attribute} and Table \ref{lab:VOT2014_attribute_noise} show the average overlap and failures for all attributes on VOT $2015$.

%Compared to SAMF, our approach enjoys a reduced failure rate for four attributes. We leave extension works on promoting robustness of $\text{K}(\text{MF})^2\text{JMT}$ as future works. Possible solutions include using robust loss functions (e.g., \cite{sui2016real}) and integrating online feature selection strategies (e.g., \cite{roffo2016online}).

%On the other hand, the PLT-$13$ and PLT-$14$ achieves better robustness on all attributes compared to the CFTs. This is because the PLT approaches employ a feature selection strategy to enhance the robustness, while CFTs compute a dense set of classification scores in a limited search region without any feature selection. We leave extension works on promoting robustness of $\text{K}(\text{MF})^2\text{JMT}$ as future works. Possible solutions include using robust loss functions (e.g., \cite{sui2016real}) and integrating online feature selection strategies (e.g., \cite{roffo2016online}).

\subsection{Comparison among correlation filter-based trackers (CFTs)} \label{experiments_section3}
%Correlation-filter-based Trackers (CFTs) have aroused increasing interests in visual tracking over these years and achieved extremely compelling results in accuracy, robustness and speed \cite{chen2015experimental}.

In the last section, we investigate the performance of $15$ representative CFTs. Our goal is to demonstrate the effectiveness of $\text{K}(\text{MF})^2\text{JMT}$ and reveal its properties when compared with other CFT counterparts. We also attempt to illustrate the future research directions of CFTs through a comprehensive evaluation. Table \ref{lab:CFTs_difference} summarized the basic information of the selected CFTs as well as their corresponding FPS values (some descriptions are adapted from \cite{chen2015experimental}). %The success plot in Fig. \ref{Comparison_correlation_family}(a) shows OPE result. Our approach significantly outperforms traditional CFTs, by achieving an AUC score of $52.9\%$. It is worth noting that our approach provides a gain of $7.7\%$ and $0.2\%$ in AUC compared to the KCF and SAMF trackers respectively. Figs. \ref{Comparison_correlation_family}(b) and \ref{Comparison_correlation_family}(c) show the success plots for the TRE and SRE analysis. In both evaluations, our approach performs favorably compared to existing methods.
%\textcolor{red}{The success plots for OPE, TRE and SRE are shown in Fig. \ref{Comparison_correlation_family}(a)-(c) respectively. In all evaluations, our approach performs favorably compared to existing methods.}

\begin{figure*}[!t]
\centering
\subfigure[One pass evaluation (OPE)] {\includegraphics[width=.30\textwidth]{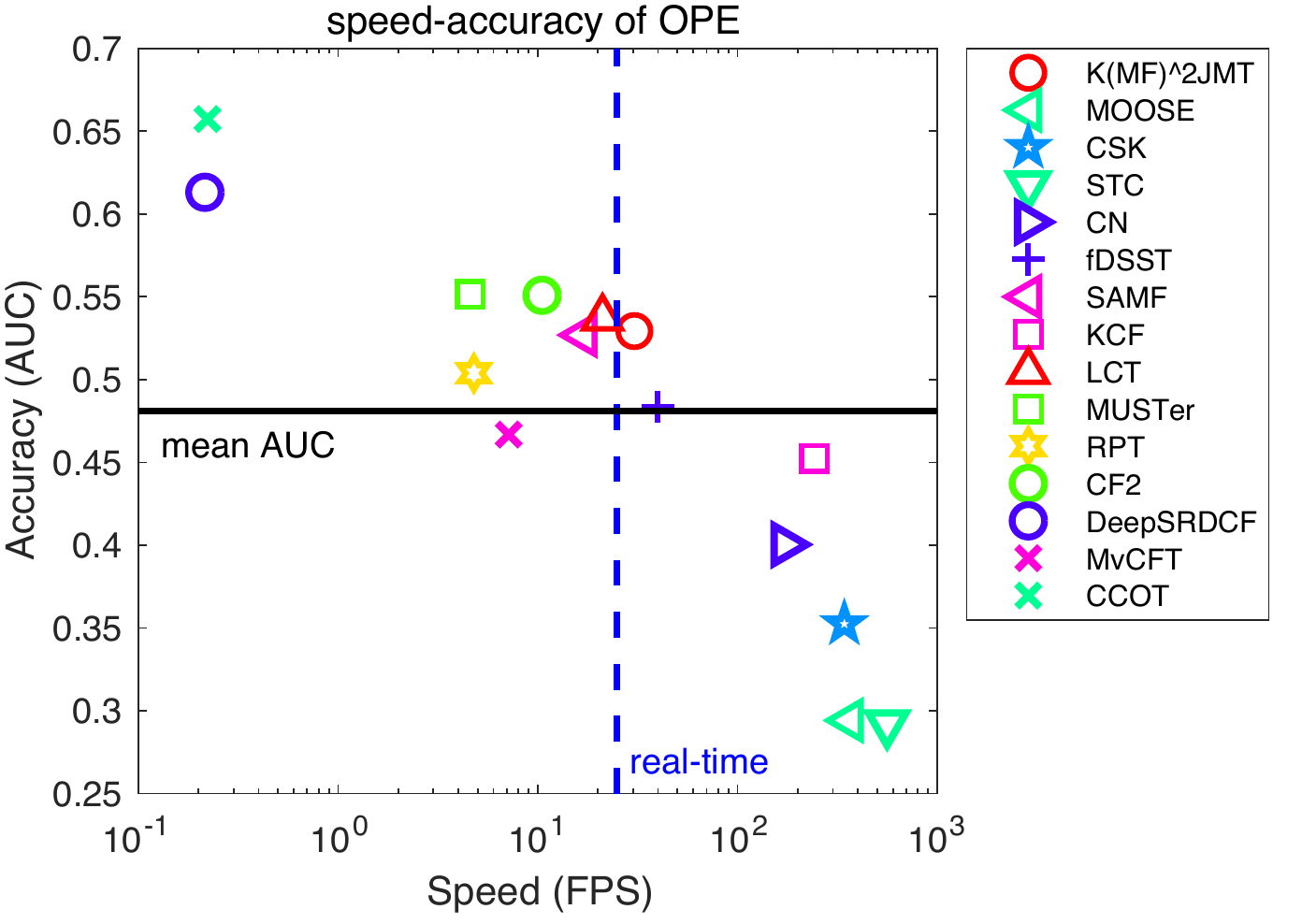}}
\subfigure[Temporal robustness evaluation (TRE)] {\includegraphics[width=.30\textwidth]{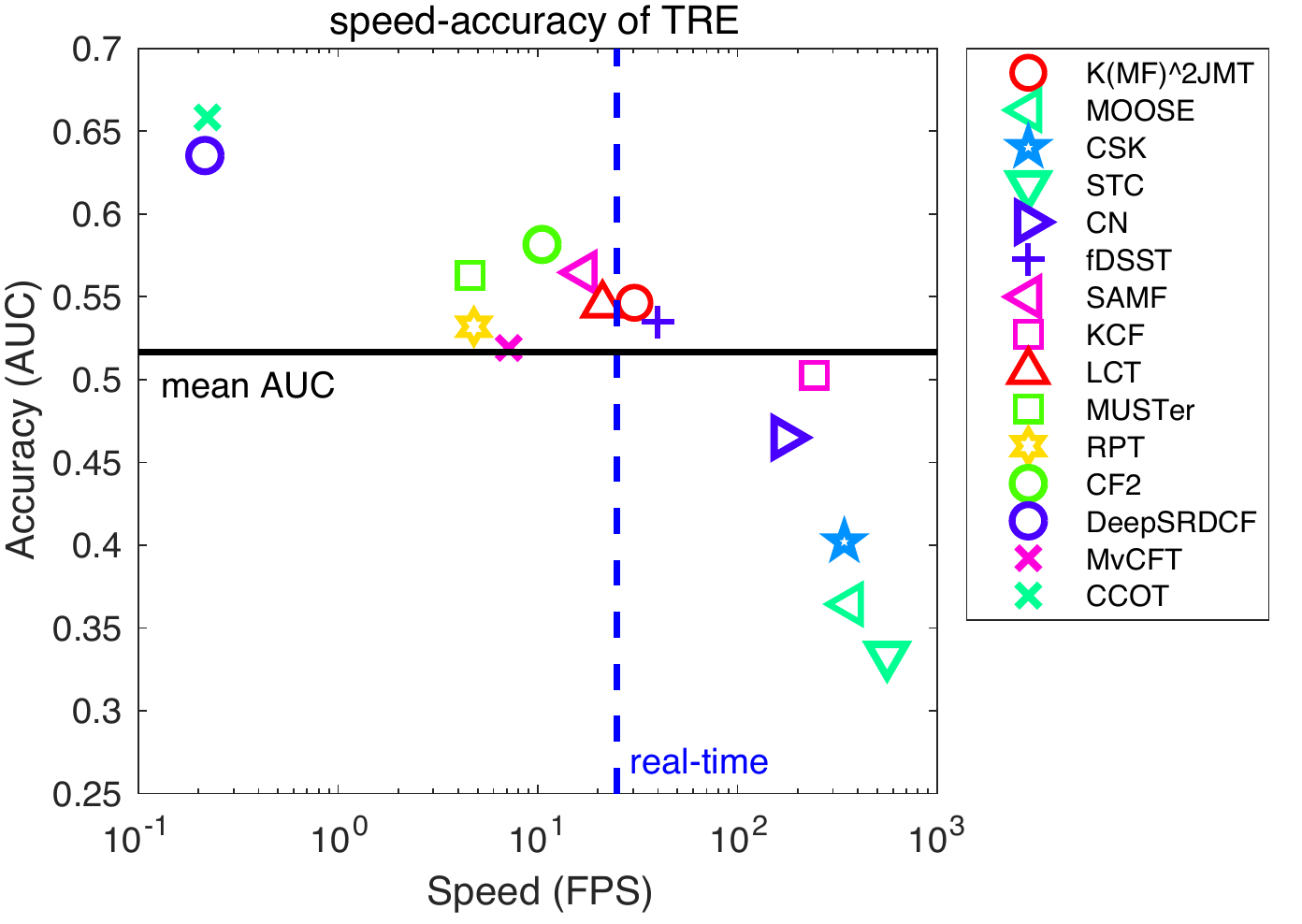}}
\subfigure[Spatial robustness evaluation (SRE)] {\includegraphics[width=.30\textwidth]{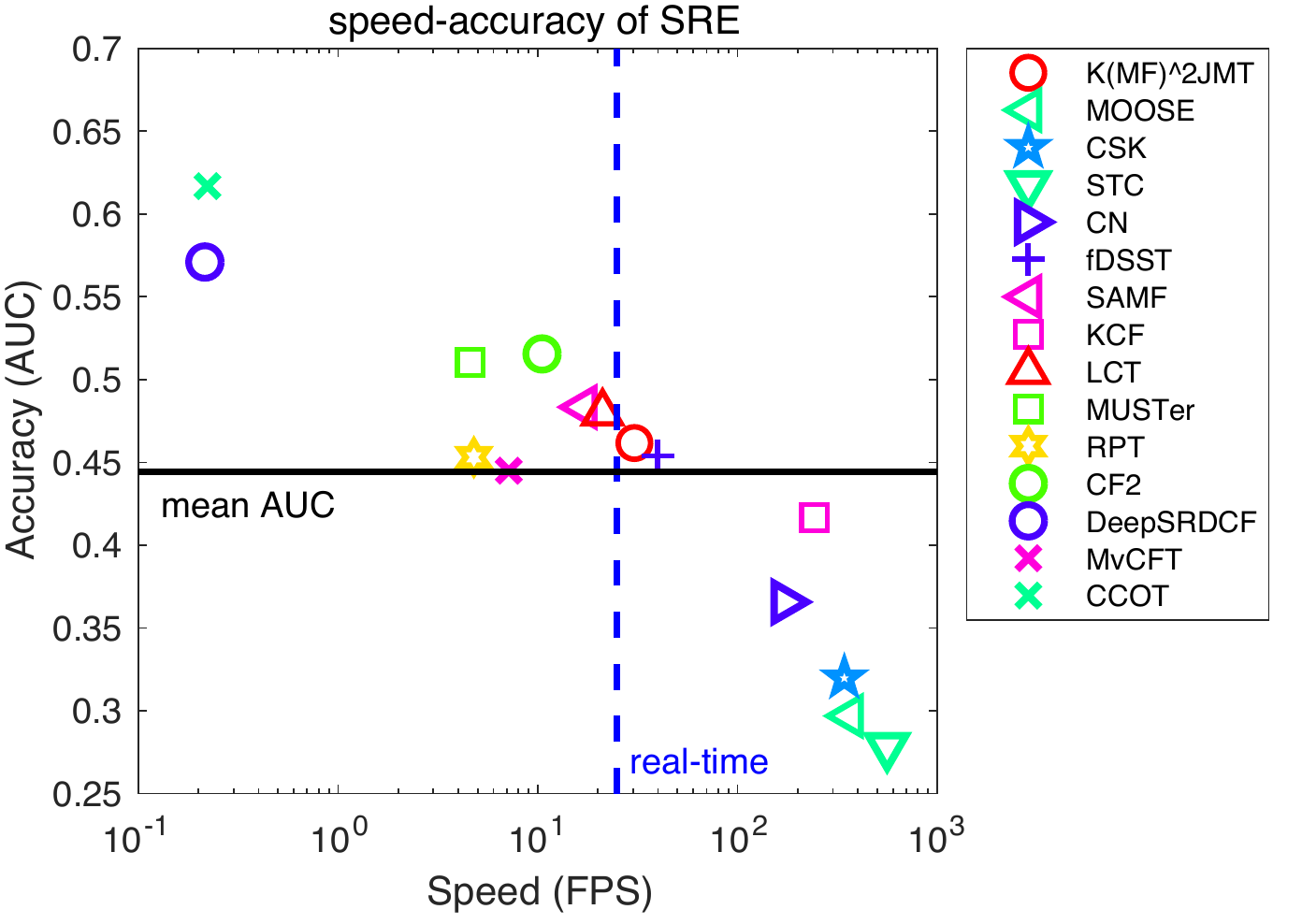}}
\caption{Speed and accuracy plot of state-of-the-art CFTs on OTB 2015 dataset. We use the AUC score of success plots to measure tracker accuracy or robustness. The dashed vertical line (with a FPS value of $25$~\cite{fan2017parallel}) separates the trackers into real-time trackers and those cannot run in real time. Meanwhile, the solid horizontal line (mean of AUC values for all competing CFTs) separates trackers into well-performed trackers and those perform poorly in terms of tracking accuracy. The proposed $\text{K}(\text{MF})^2\text{JMT}$ achieves the best accuracy among all real-time trackers in terms of (a) OPE, (b) TRE and (c) SRE.}
\label{Comparison_correlation_family}
\end{figure*}

\begin{table*}
\scriptsize
\centering
\caption{Selected competing CFTs (including summarized major contributions) and their corresponding FPS values (adapted from \cite{chen2015experimental}).}\label{lab:CFTs_difference}
\begin{tabular}{cccc}\hline
 & Published year & Major contribution & FPS \\\hline
$\text{MOSSE}$ \cite{bolme2010visual} & $2010$ & \text{Pioneering work of introducing correlation filters for visual tracking.} & $367.31$ \\
$\text{CSK}$ \cite{henriques2012exploiting} & $2012$ & \text{Introduced Ridge Regression problem with circulant matrix to apply kernel methods.} & $336.00$ \\
$\text{STC}$ \cite{zhang2014fast} & $2014$ & \text{Introduced spatio-temporal context information.} & $572.03$ \\
$\text{CN}$ \cite{danelljan2014adaptive} & $2014$ & \text{Introduced color attributes as effective features.} & $171.69$ \\
$\text{DSST}$ \cite{danelljan2014accurate} & $2014$ & \text{Relieved the scaling issue using feature pyramid and 3-dimensional correlation filter.} & $39.54$ \\
$\text{SAMF}$ \cite{li2014scale} & $2014$ & \text{Integrated both color feature and HOG feature; Applied a scaling pool to handle scale
variations.} & $17.23$ \\
$\text{KCF}$ \cite{henriques2015high} & $2015$ & \text{Formulated the work of CSK and introduced multi-channel HOG feature.} & $241.70$ \\
$\text{LCT}$ \cite{ma2015long} & $2015$ & \text{Introduced online random fern classifier as re-detection component for long-term
tracking.} & $21.63$ \\
$\text{MUSTer}$ \cite{hong2015multi} & $2015$ & \text{Proposed a biology-inspired framework to integrate short-term processing and long-term processing.} & $4.54$ \\
$\text{RPT}$ \cite{li2015reliable} & $2015$ & \text{Introduced reliable local patches to facilitate tracking.} & $4.76$ \\
$\text{CF2}$ \cite{ma2015hierarchical} & $2015$ & \text{Introduced features extracted from convolutional neural networks (CNN) for visual tracking.} & $10.76$ \\
$\text{DeepSRDCF}$ \cite{danelljan2015convolutional} & $2015$ & \text{Introduced CNN features for visual tracking and a spatial regularization term to handle bound effect.} & $0.21$ \\
$\text{MvCFT}$ \cite{li2016multi} & $2016$ & \text{Introduced Kullback-Leibler (KL) divergence to fuse multiple feature cues.} & $7.07$ \\
$\text{C-COT}$ \cite{danelljan2016beyond} & $2016$ & \text{Employed an implicit interpolation model to train convolutional filters in continuous spatial domain.} & $0.22$ \\
$\text{K}(\text{MF})^2\text{JMT}$ & $2018$ & \text{Integrated multiple feature cues and temporal consistency in a unified model.} & $30.46$ \\\hline
\end{tabular}
\end{table*}

Fig.~\ref{Comparison_correlation_family} shows the AUC scores of success plots vs. FPS for all competing CFTs in OPE, TRE and SRE, respectively. The dashed vertical line (with a FPS value of $25$~\cite{fan2017parallel}) separates the trackers into real-time trackers and those cannot run in real time. Meanwhile, the solid horizontal line (mean of AUC values for all competing CFTs) separates trackers into well-performed trackers and those perform poorly in terms of tracking accuracy. As can be seen, our method performs the best in terms of accuracy among all real-time trackers, thus achieving a good trade-off in speed and accuracy. This suggests that our modifications on the basic KCF tracker are effective and efficient. By contrast, although introducing long-term tracking strategy (e.g., MUSTer, LCT) or imposing spatial regularization penalty term (e.g., DeepSRDCF) can augment tracking performance as well, most of these modifications cannot be directly applied in real applications where the real-time condition is a prerequisite. This unfortunate fact also applies to C-COT, in which the hand-crafted features are substituted with powerful CNN features. Therefore, a promising research direction is to investigate computational-efficient long term tracking strategy or spatial regularization with little or no sacrifice in speed.

Finally, it is worth mentioning that our approach provides a consistent gain in performance compared to MvCFT, although both methods employ the same feature under a MVL framework. The MvCFT only fuses tracking results from different view to provide a more precise prediction. By contrast, our approach enjoys a more reliable and computational-efficient training model. On the other hand, it is surprising to find that SAMF can achieve desirable performance on both TRE and SRE experiments. One possible reason is that the scale estimation method in SAMF, i.e., exhaustively searching a scaling pool, is more robust to scale variations (although time-consuming).

%Next, an emerging trend towards improving tracking accuracy is to use powerful CNN features. Such methods (e.g., C-COT, DeepSRDCF), however, suffer from high computational burden and hardly run in real-time. Finally, introducing long-term tracking strategy seems to be a promising direction, as it has been proven effective in augmenting the performance of traditional CFTs (see the results of MUSTer and LCT).

\begin{comment}
\begin{figure}[!htbp]
\centering
\includegraphics[width=.45\textwidth]{auc_fps_plot}
\caption{\textcolor{red}{Speed and accuracy plot of state-of-the-art CFTs on OTB 2015 dataset. The proposed $\text{K}(\text{MF})^2\text{JMT}$ achieves the best accuracy among all real-time trackers.}}
\label{auc_fps_plot}
\end{figure}
\end{comment}

%First, it can be found that the possibilities of drifting are significantly reduced with the assist of powerful CNN features (see the results of DeepSRFCF and CF2). Thus, it is worthy to try more types of features. Next, introducing long-term tracking strategy seems to a promising direction, as it has been proven effective in augmenting the performance of traditional CFTs (see the results of MUSTer and LCT). Finally, it is desirable to develop more computational efficient filter learning methods.

%Although C-COT, DeepSRDCF or MUSTer enjoy superior performance than our approach, their computational complexity are far from optimal and much worse than ours. We expect further improvement in the performance of our tracker by incorporating the aforementioned strategies. We leave it as future work.

\section{Conclusions and future work}

%In this paper, we proposed kernel multi-frame multi-feature joint modeling tracker ($\text{K}(\text{MF})^2\text{JMT}$) to promote the original correlation filter-based trackers (CFTs). By exploiting multiple feature cues and the temporal consistency in a unified framework, $\text{K}(\text{MF})^2\text{JMT}$ leads to a more discriminative appearance representation and a more reliable model training scheme. A fast $\emph{blockwise diagonal matrix}$ inversion algorithm has been developed to speed up learning and detection. To handle scale variations, an adaptive scale estimation mechanism was applied.

%Experiments validated the rationale and effectiveness of $\text{K}(\text{MF})^2\text{JMT}$.

In this paper, we proposed kernel multi-frame multi-feature joint modeling tracker ($\text{K}(\text{MF})^2\text{JMT}$) to promote the original correlation filter-based trackers (CFTs) by exploiting multiple feature cues and the temporal consistency in a unified framework. A fast $\emph{blockwise diagonal matrix}$ inversion algorithm has been developed to speed up learning and detection. An adaptive scale estimation mechanism was incorporated to handle scale variations. Experiments on OTB 2015 and VOT 2015 datasets show that $\text{K}(\text{MF})^2\text{JMT}$ improves tracking performance in contrast with most state-of-the-art trackers. Our tracker performs well in terms of overlap success in the context of large appearance variations caused by occlusion, illumination, scale variation, etc. Our tracker also demonstrates favorable tracking accuracy and robustness compared with prevalent trackers from different categories (not limited to CFTs). We finally show that $\text{K}(\text{MF})^2\text{JMT}$ can achieve the best tracking accuracy among state-of-the-art real-time correlation filter-based trackers (CFTs).

In future work, we will study how to effectively handle severe drifts and shot changes. For the problem of drifts, possible solutions include closed-loop system design~\cite{Hu2017Correlation} or tracking-and-verifying framework~\cite{fan2017parallel}. On the other hand, to circumvent the existence of shot (or scene) changes, possible modifications include assigning different weights to different frames in the overall objective Eq.~(\ref{MTMVmodel2}) or explicitly incorporating a short change detector (e.g.,~\cite{bouthemy1999unified}), such that the tracker can automatically detect the shot changes. Once a shot change is confirmed, the tracker needs to re-identify the location of the target (see Supplementary Material for initial results). At the same time, we are also interested in investigating computational efficient long term tracking strategy or spatial regularization to further augment tracking accuracy.

% or


%\appendices
%% you can choose not to have a title for an appendix
%% if you want by leaving the argument blank
%\section{}

\section*{Appendix A \\ Circular matrix and blockwise circular matrix}
Given a vector $\ve{x}=[x_0, x_1, ..., x_{n-1}]^T$ of length $n$ and its Discrete Fourier Transform (DFT) $\hat{\ve{x}}=\mathcal{F}(\ve{x})$, the \emph{circular matrix} $X=C(\ve{x})$ generated by $\ve{x}$ has the following form:
\begin{eqnarray}
X=C(\ve{x})=\left(\begin{array}{*{5}{c}}
x_0 & x_1 & x_2 & \cdots & x_{n-1}\\
x_{n-1} & x_0 & x_1 & \cdots & x_{n-2}\\
x_{n-2} & x_{n-1} & x_0 & \cdots & x_{n-3}\\
\vdots & \vdots & \vdots & \ddots & \vdots\\
x_1 & x_2 & x_3 & \cdots & x_0
\end{array}
\right).
\end{eqnarray}
\cite{gray2006toeplitz} proved that $X$ can be diagonalized as:
\begin{eqnarray}
X = F\mbox{diag}(\hat{\ve{x}})F^H,
\end{eqnarray}
where $F$ is known as the \emph{DFT matrix} ($\hat{\ve{x}}=\sqrt{n}F\ve{x}$), and $\mbox{diag}(\hat{\ve{x}})$ denotes a square diagonal matrix with elements of $\hat{\ve{x}}$ on the diagonal.

We term $\mat{X}$ \emph{blockwise circular matrix} if it consists of ${M}\times{M}$ blocks $\mat{X}_{ij}$ ($i,j=1,...,M$) and each block $\mat{X}_{ij}$ is a circular matrix generated by $\ve{x}_{ij}$ of length $n$. We then denote $\ve{F}$ a \emph{block diagonal matrix} with blocks $F$ on the main diagonal:
\begin{eqnarray}
\ve{F}=\left(\begin{array}{ccc}
F & \cdots & 0\\
\vdots & \ddots & \vdots\\
0 & \cdots & F
\end{array}
\right)
\label{EqF}
\end{eqnarray}
and $\ve{\Sigma}$ a \emph{blockwise diagonal matrix}:
\begin{eqnarray}
\ve{\Sigma}=\left(\begin{array}{ccc}
\mbox{diag}(\hat{\ve{x}}_{11}) & \cdots & \mbox{diag}(\hat{\ve{x}}_{1M})\\
\vdots & \ddots & \vdots\\
\mbox{diag}(\hat{\ve{x}}_{M1}) & \cdots & \mbox{diag}(\hat{\ve{x}}_{MM})
\end{array}
\right),
\end{eqnarray}
then $\ve{X}$ can be decomposed as:
\begin{eqnarray}
\ve{X}=\ve{F}\ve{\Sigma}\ve{F}^H.
\end{eqnarray}

%\section*{Appendix B \\ Attribute-based comparison using success plots}

\section*{Acknowledgment}
This work was supported in part by the Key Science and Technology of Shenzhen~(No. CXZZ20150814155434903), the Key Program for International S\&T Cooperation Projects of China~(No. 2016YFE0121200), the Key Science and Technology Innovation Program of Hubei Province~(No. 2017AAA017), the Special Projects for Technology Innovation of Hubei Province~(2018ACA135), the National Natural Science Foundation of China~(No. 61571205 and No. 61772220).

\ifCLASSOPTIONcaptionsoff
  \newpage
\fi

\bibliographystyle{IEEEtran}
\bibliography{mybibfile}


%\end{IEEEbiography}


%\newpage



%\vfill


%\end{multicols} %just before \end{document}

\clearpage
\onecolumn

\section*{Supplementary Material to Robust Visual Tracking using Multi-Frame Multi-Feature Joint Modeling}

\subsection{$\text{K}(\text{MF})^2\text{JMT}$ on video sequences with shot changes and possible modifications}

In this section, we provide tracking results of $\text{K}(\text{MF})^2\text{JMT}$ and five state-of-the-art trackers (i.e., MEEM \cite{zhang2014meem}, TGPR \cite{gao2014transfer}, Struck \cite{hare2011struck}, SCM \cite{zhong2012robust}, ASLA \cite{jia2012visual}) on five video sequences (three are from OTB $2015$, the remaining two are from VOT$2015$ benchmark) with shot changes or scene cuts. We also suggest two modifications to our current $\text{K}(\text{MF})^2\text{JMT}$ to alleviate the negative effects incurred by the these changes.

The first modification is to give different weights to different frames in the overall objective of $\text{K}(\text{MF})^2\text{JMT}$ (i.e., Eq.~($1$) in the main text). The motivation is intuitive: in the scenarios of shot changes or scene cuts, the temporal coherence (from previous frame) becomes weaker and the tracker needs to assign more weight to the most adjacent (or neighboring) frame to better capture the instantaneous information. The second modification is to incorporate a shot change detector (e.g.,~\cite{bouthemy1999unified,birinci2014perceptual}) into our $\text{K}(\text{MF})^2\text{JMT}$, such that the system can automatically detect the shot changes. Once a shot change is confirmed, the system needs to re-detect or re-identify the location of the target. However, one should note that, there is no guarantee that the selected shot detector can reconcile with the given tracker. Moreover, the integration of shot detector will introduce more hyper-parameters.

The selected videos are \textit{DragonBaby}, \textit{BlurOwl}, \textit{Soccer}, \textit{Singer1} and \textit{Singer3}. In the video \textit{DragonBaby}, the shot change is caused by varying camera-subject distances, i.e., there is shot change from full shot to medium shot\footnote{Please refer to~\cite{sethi1995statistical} for definitions of full shot, medium shot, etc.}. In the video \textit{BlurOwl}, the shot change is caused by the sudden changes of camera point-of-view or angle. In the video \textit{Soccer}, the shot change is caused by either the gradual changes of camera point-of-view or the varying camera-subject distances. In the videos \textit{Singer1} and \textit{Singer3}, the shot change is caused by (rapid) changes of both camera point-of-view and camera-subject distances.

\begin{figure*}[!t]
\centering
\subfigure[Shot changes in \textit{DragonBaby}: there are abrupt shot changes from full shot to medium shot (see frame $78$ to frame $84$) and from medium shot to full shot (see frame $88$ to frame $94$).] {
\begin{tabular}{ccc}
\includegraphics[width=2.5cm,height=2cm]{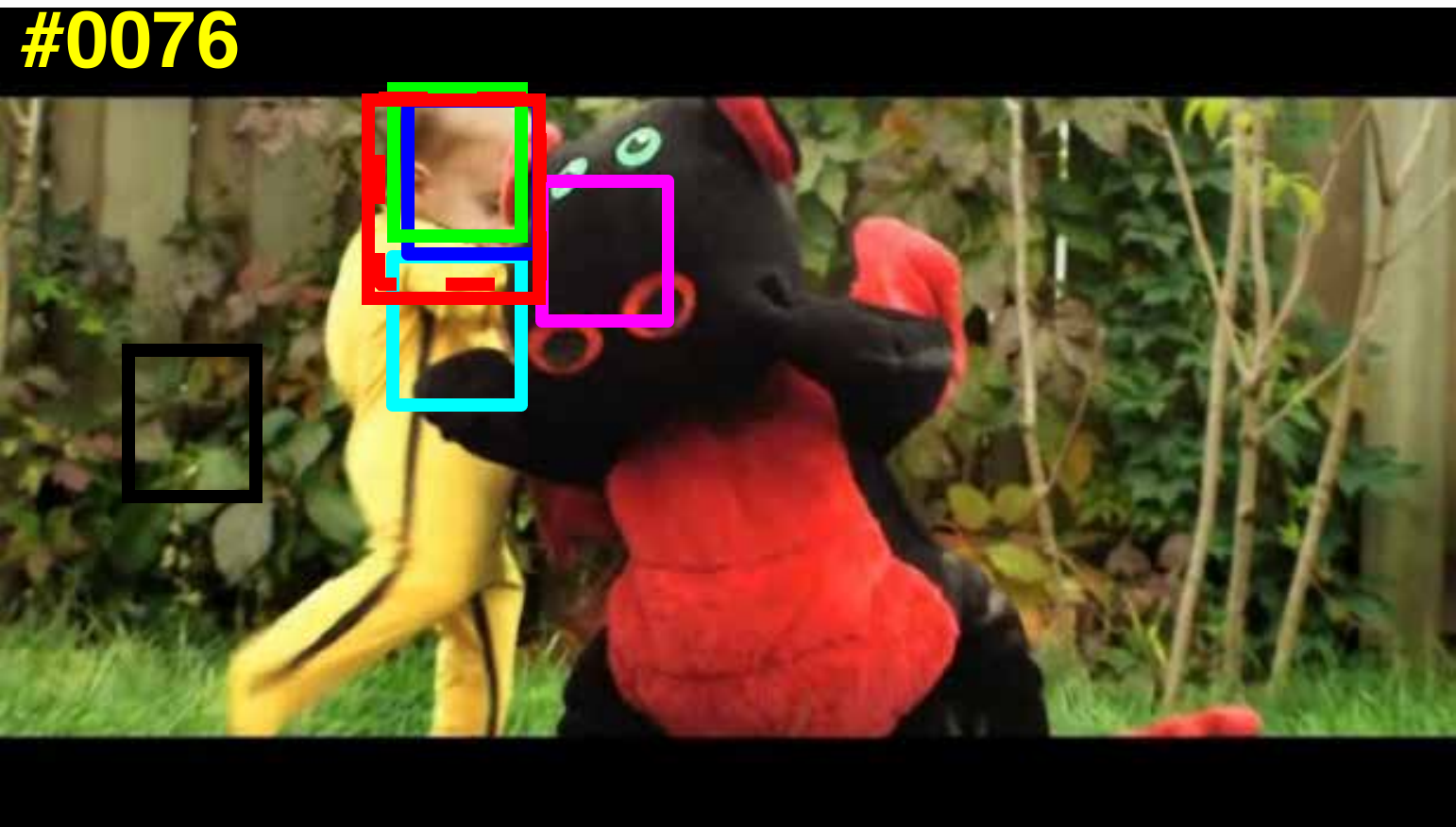}
\includegraphics[width=2.5cm,height=2cm]{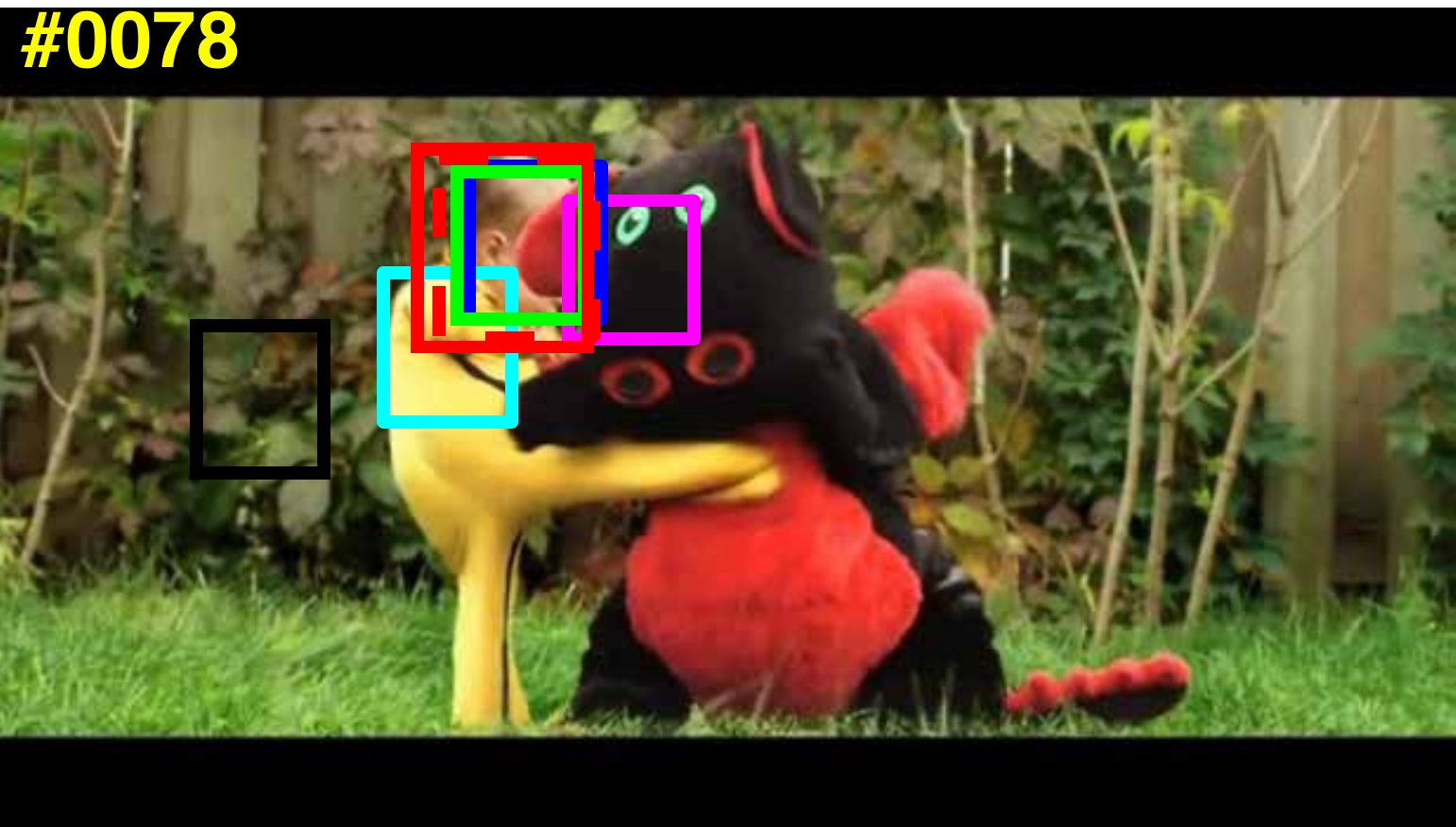}
\includegraphics[width=2.5cm,height=2cm]{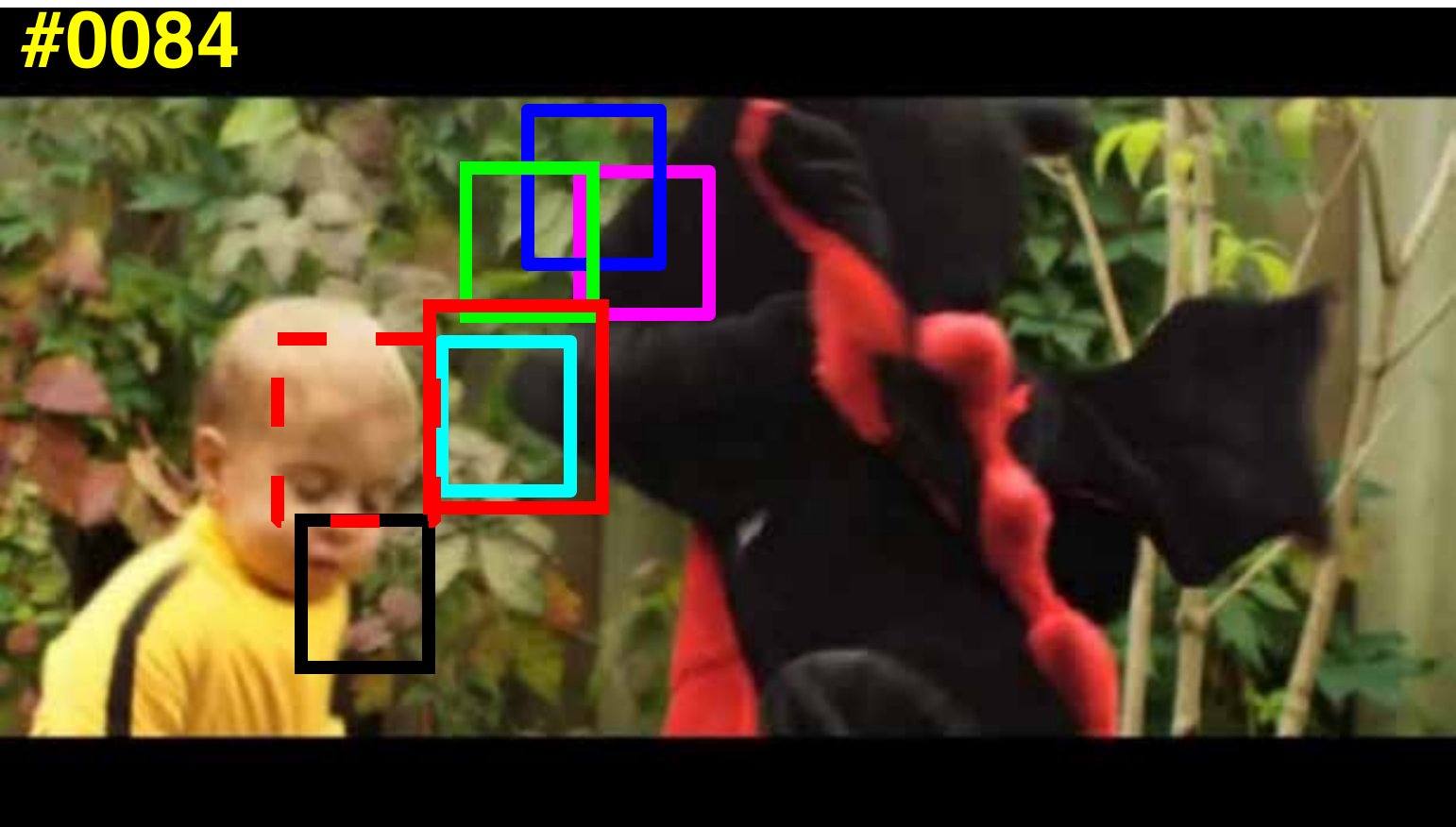}
\includegraphics[width=2.5cm,height=2cm]{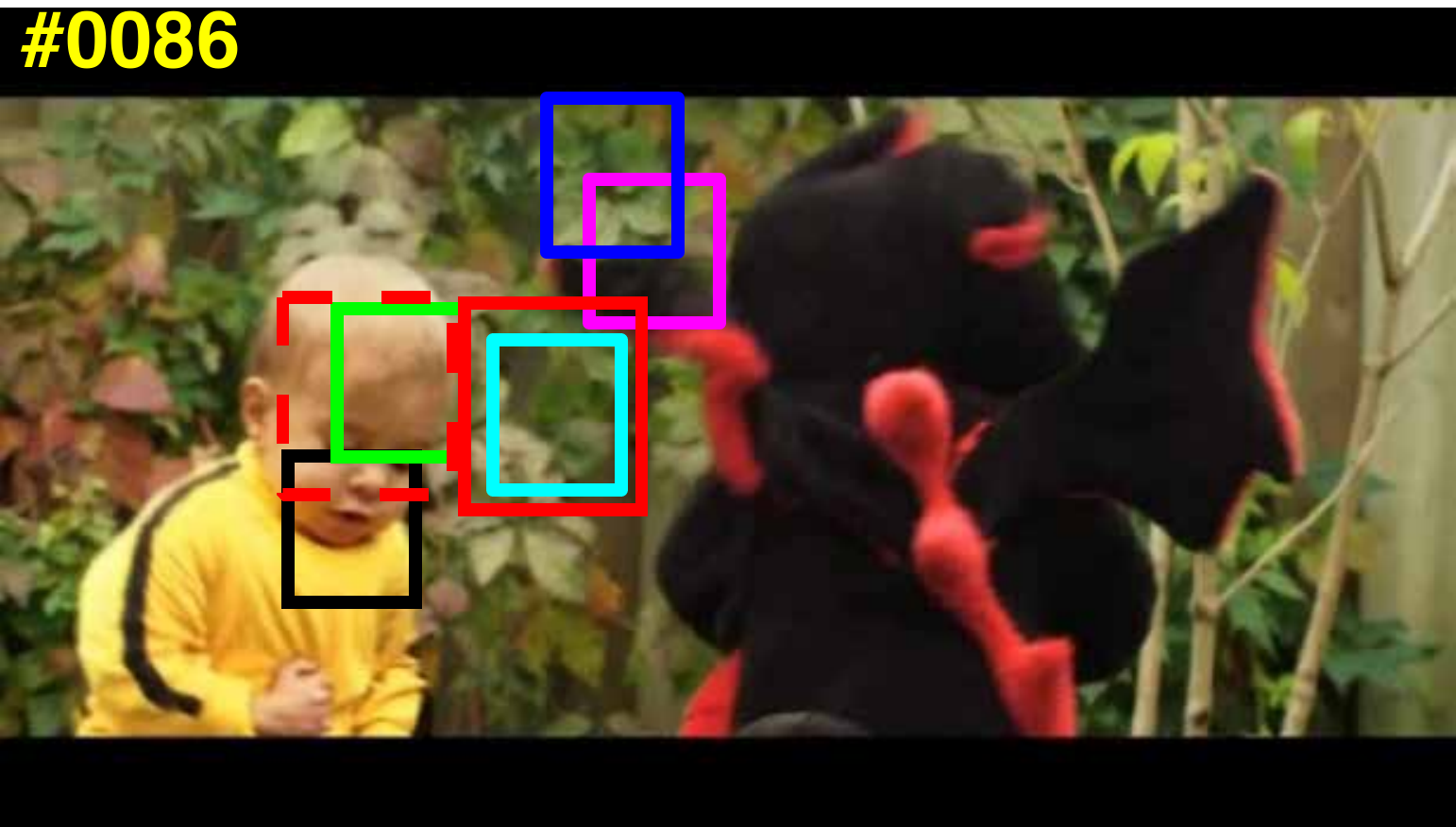}
\includegraphics[width=2.5cm,height=2cm]{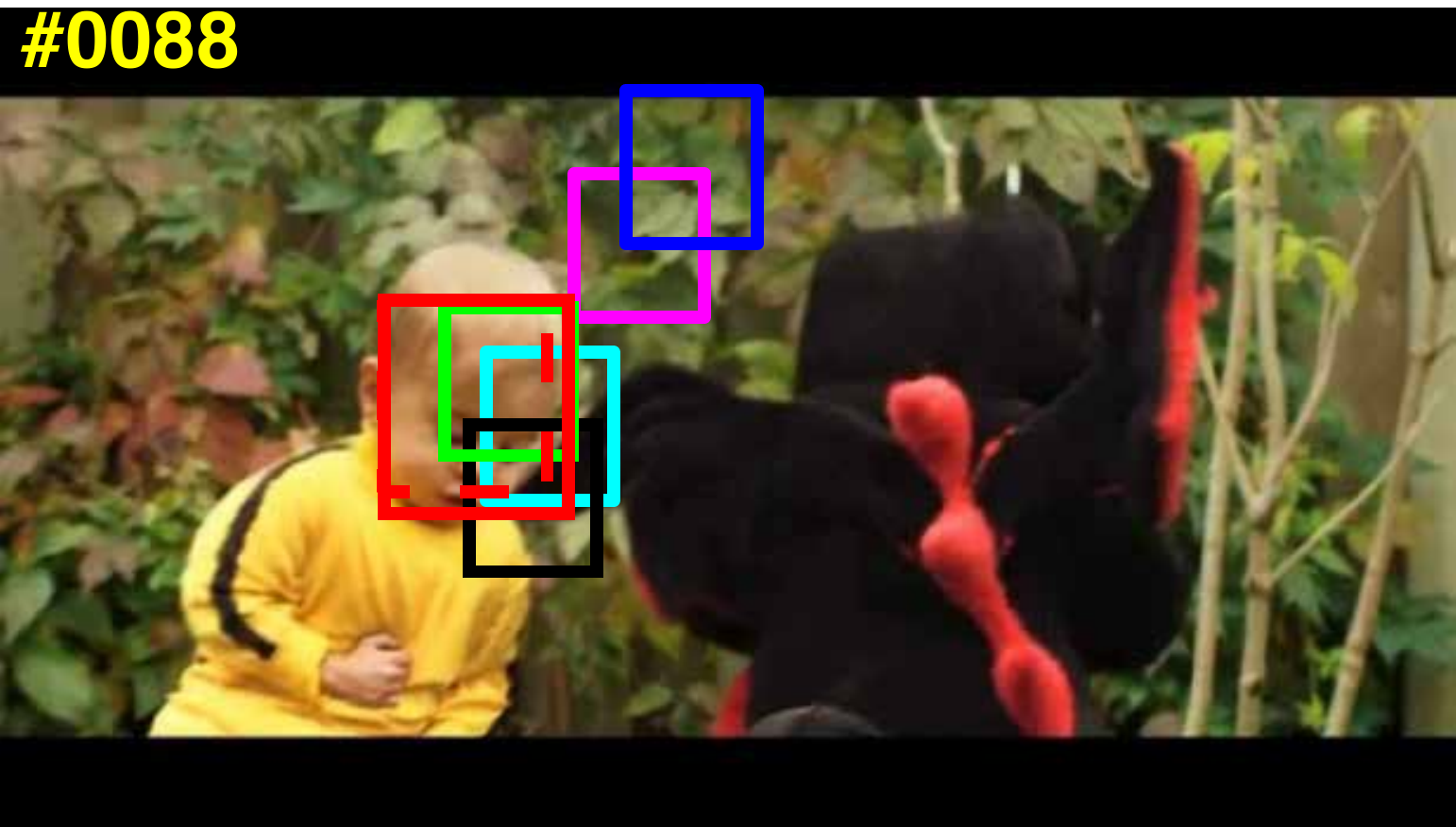}
\includegraphics[width=2.5cm,height=2cm]{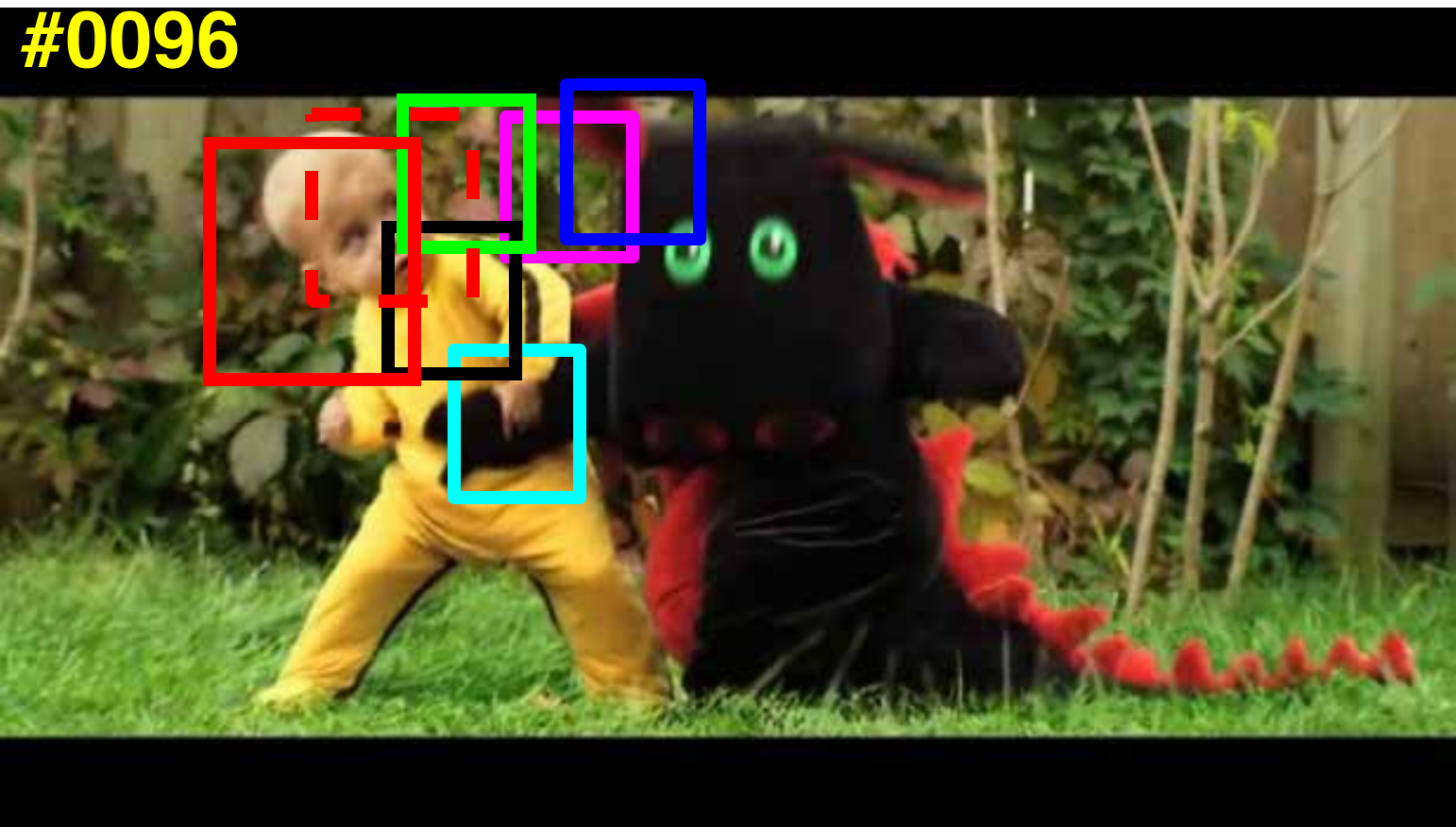}
\includegraphics[width=2.5cm,height=2cm]{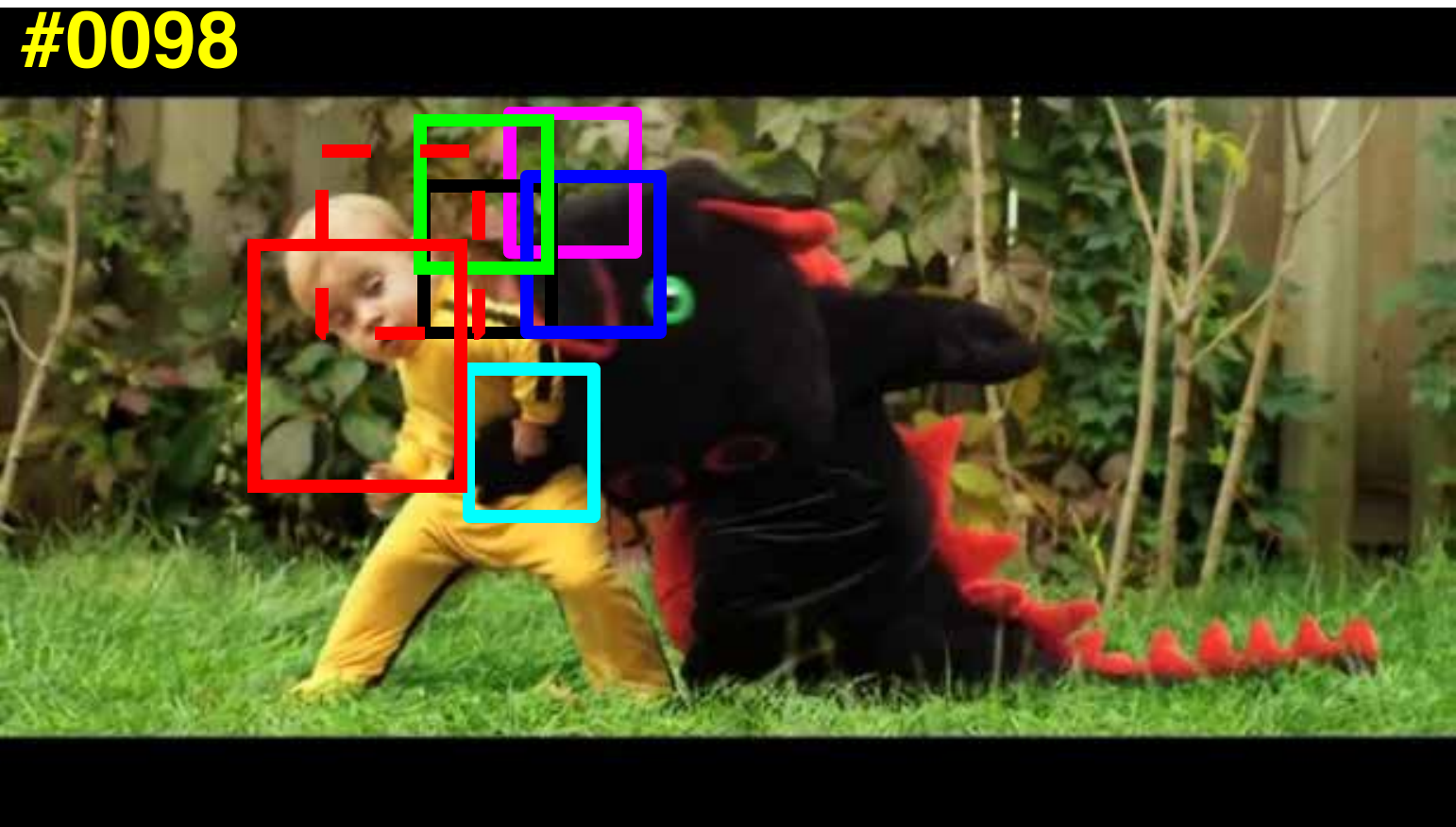}
\end{tabular}
}\\
\subfigure[Shot changes in \textit{BlurOwl}: there are abrupt shot changes due to the sudden change of camera point-of-view (see the transition between frame $375$ and frame $380$ or the transition between frame $390$ and frame $395$).] {
\begin{tabular}{ccc}
\includegraphics[width=2.5cm,height=2cm]{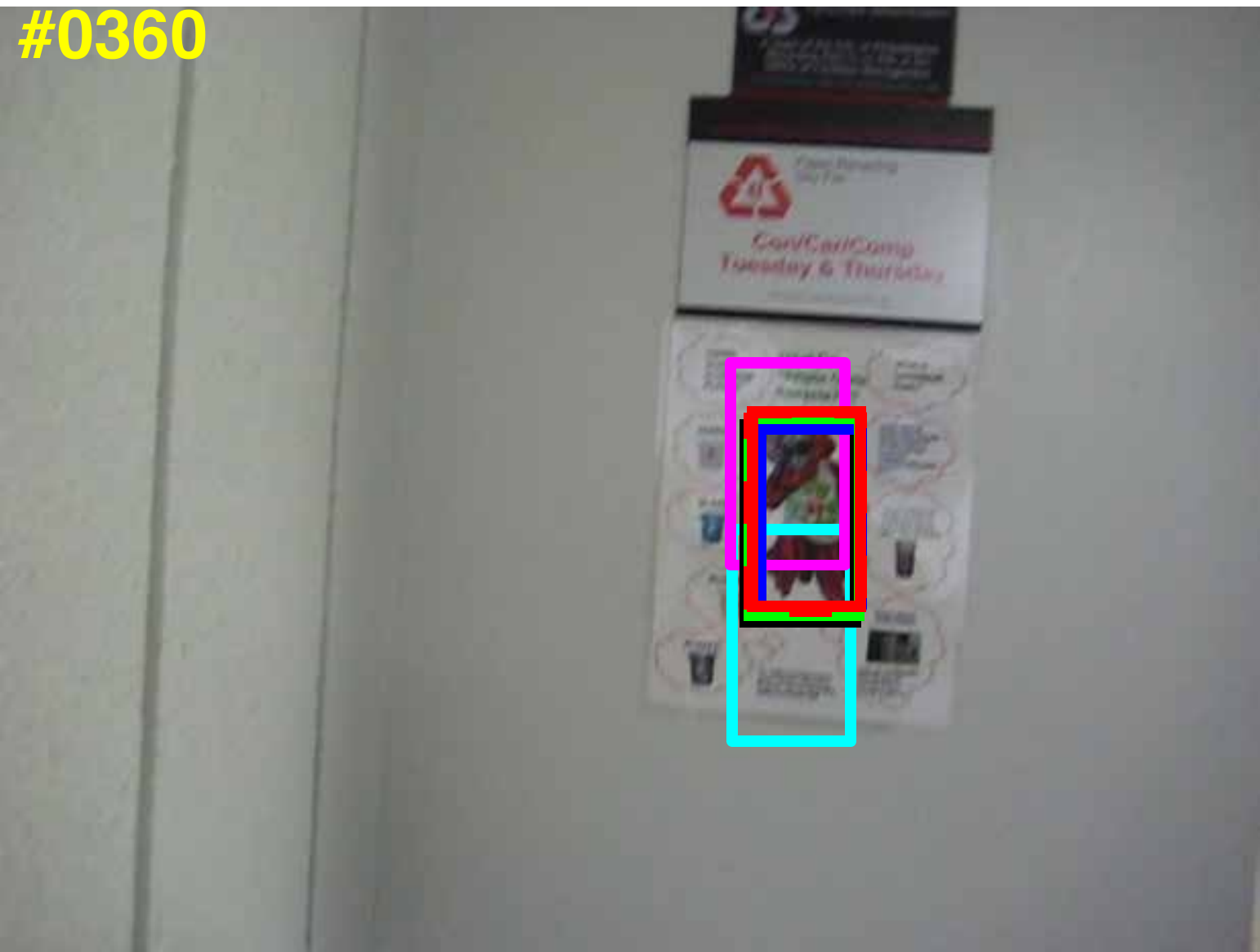}
\includegraphics[width=2.5cm,height=2cm]{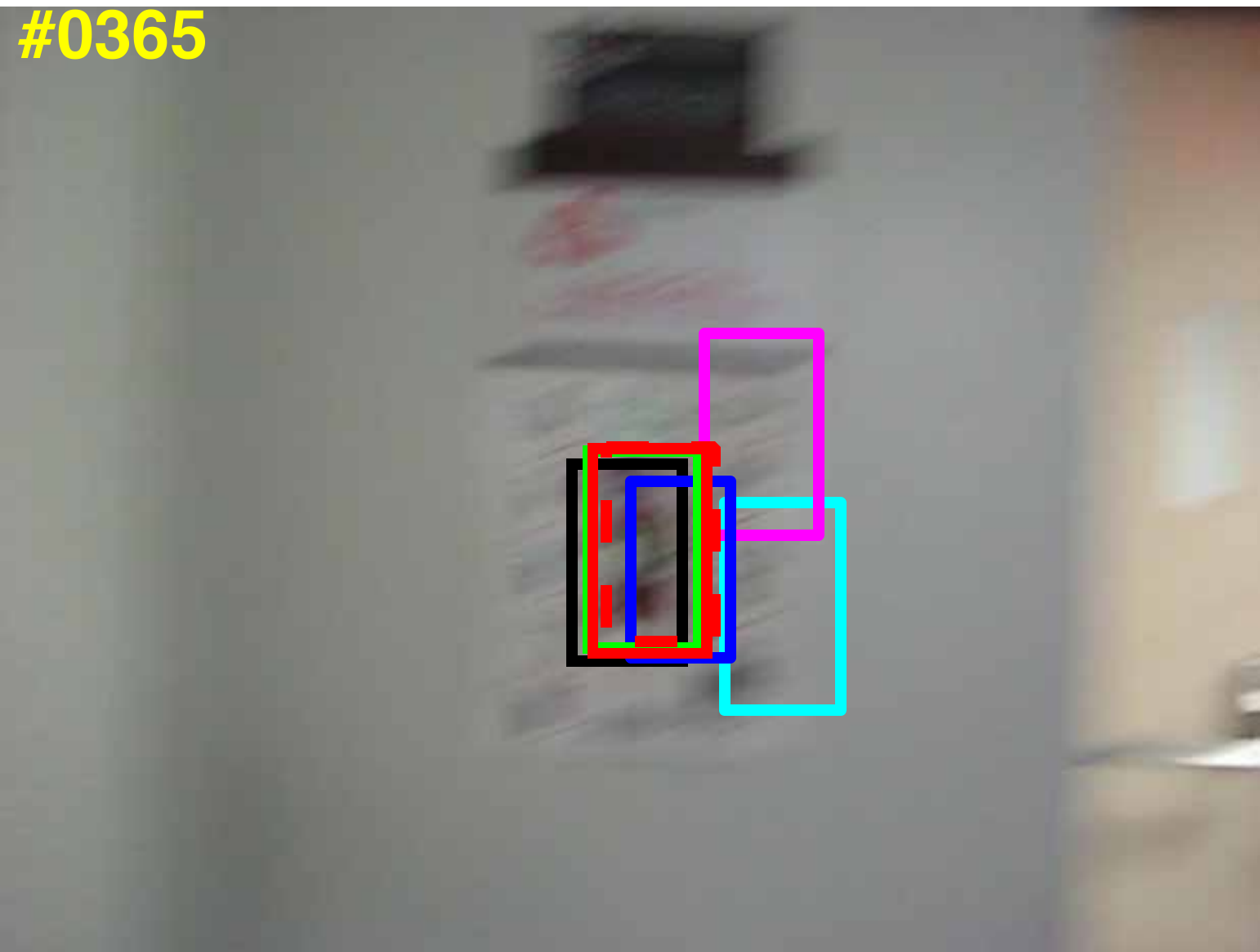}
\includegraphics[width=2.5cm,height=2cm]{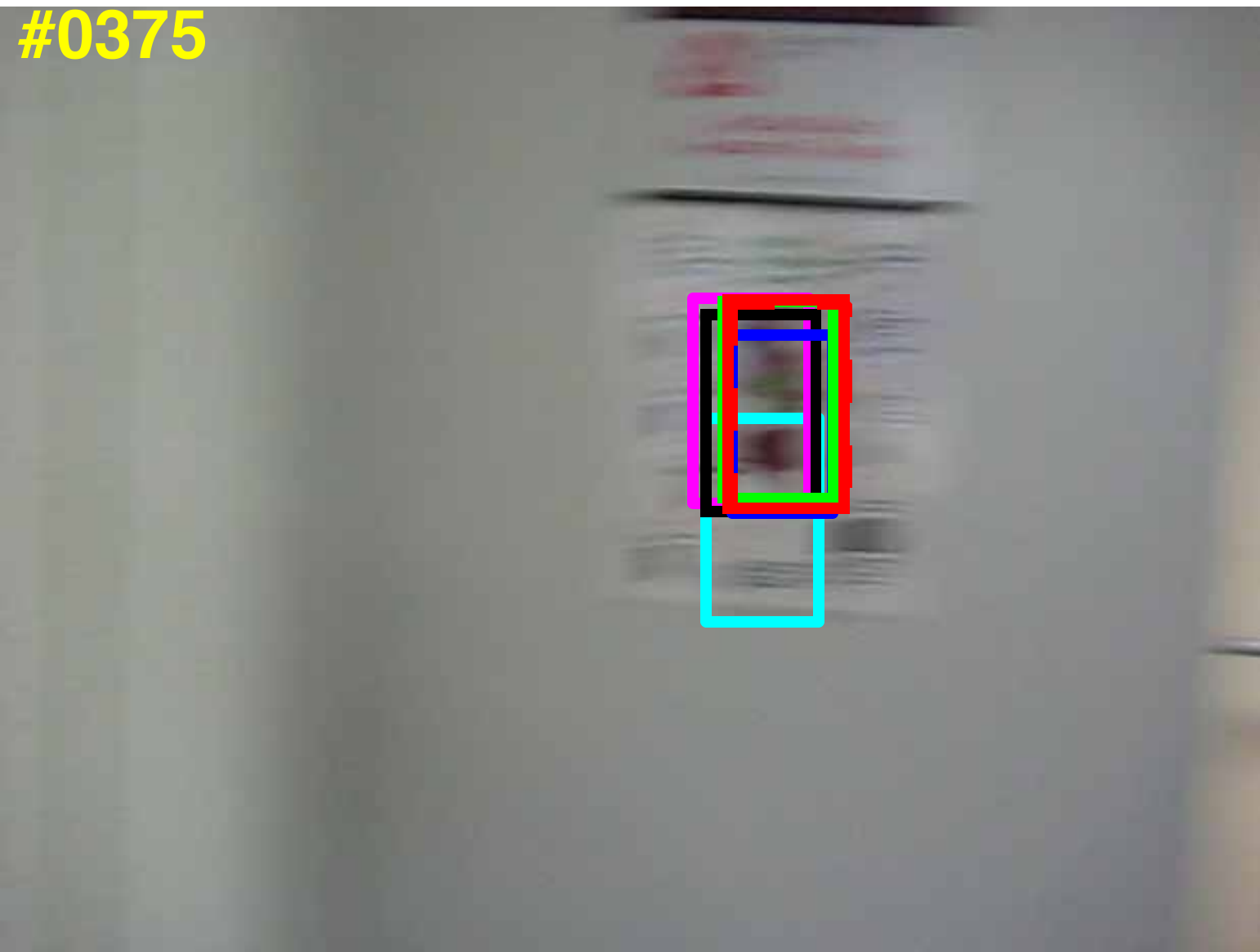}
\includegraphics[width=2.5cm,height=2cm]{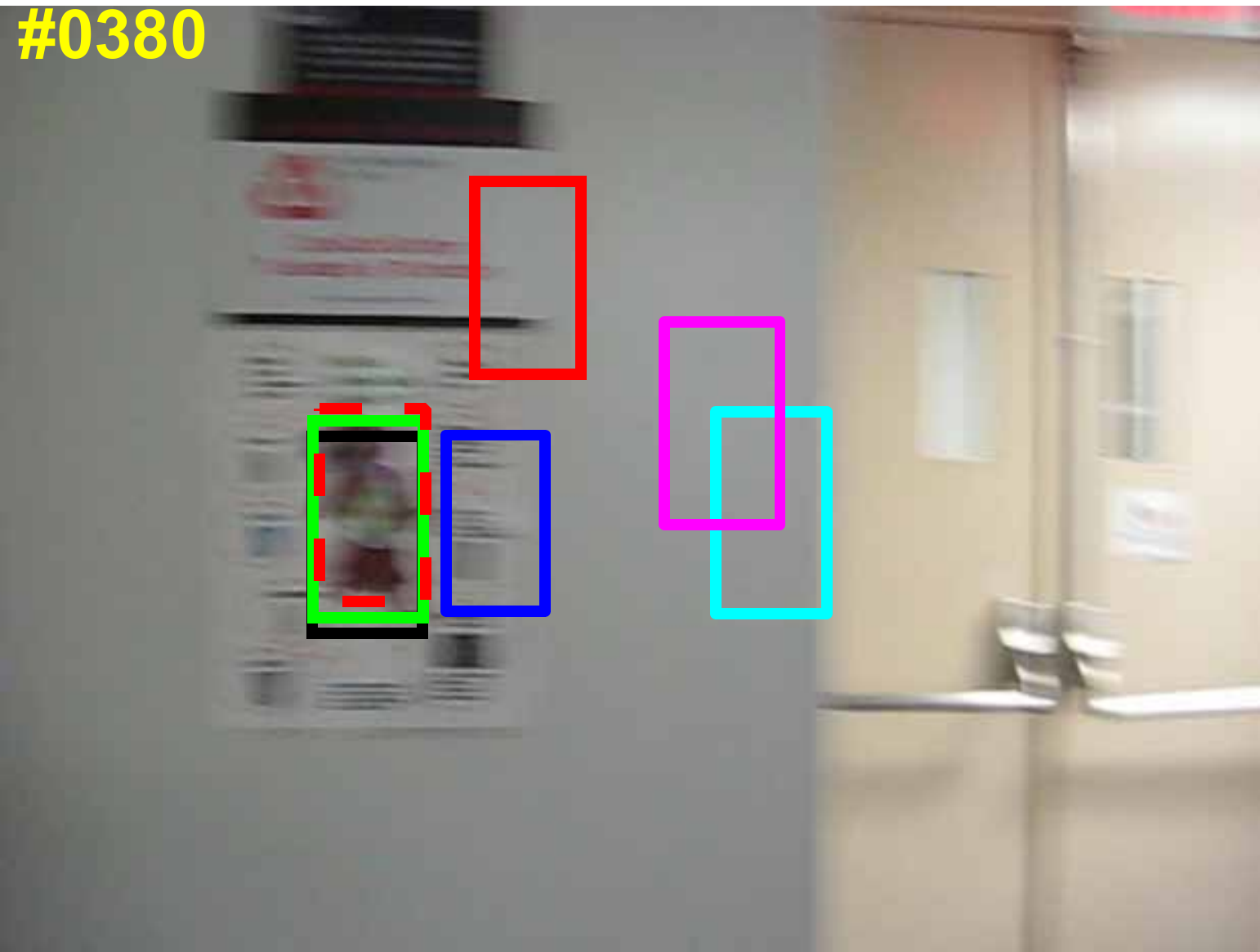}
\includegraphics[width=2.5cm,height=2cm]{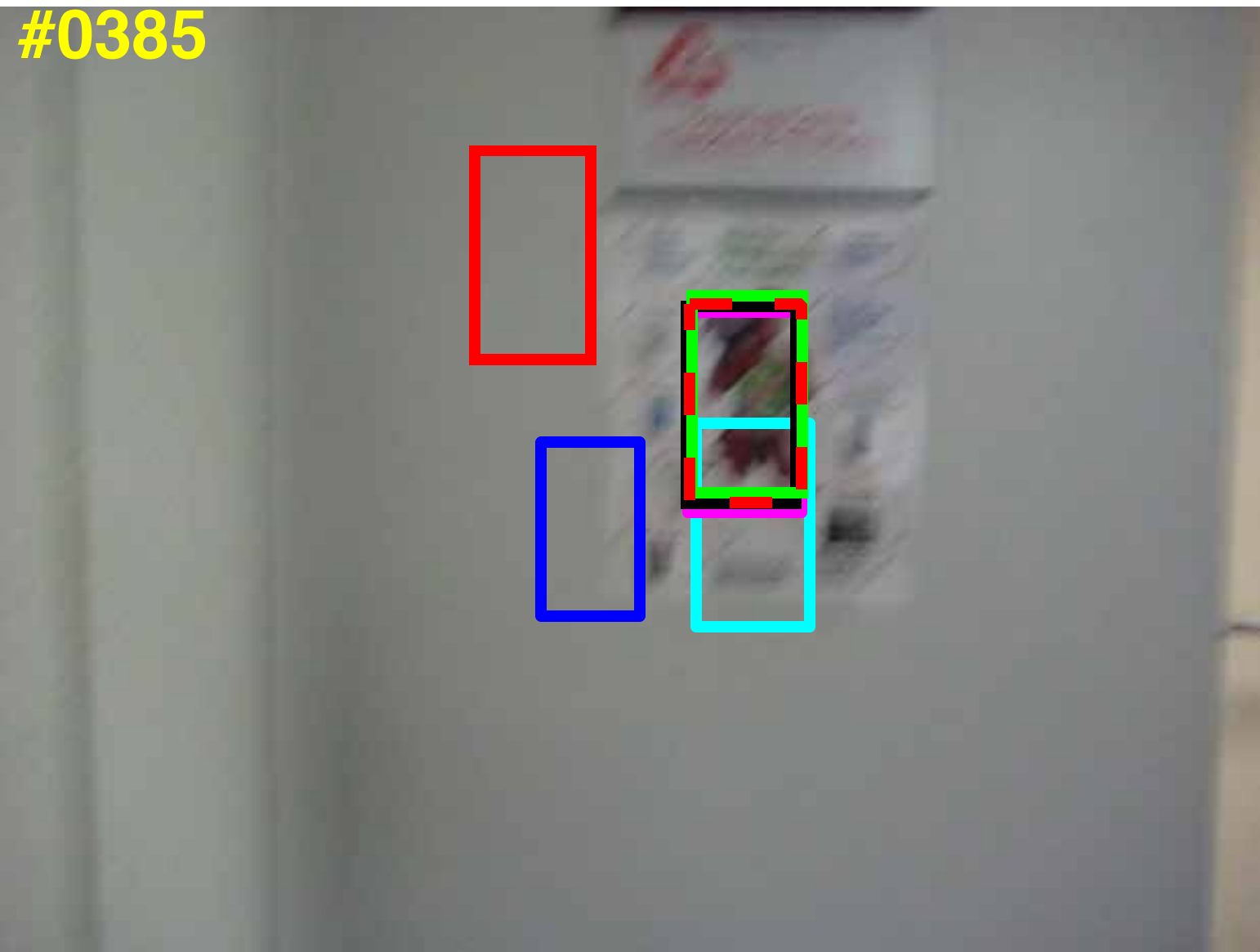}
\includegraphics[width=2.5cm,height=2cm]{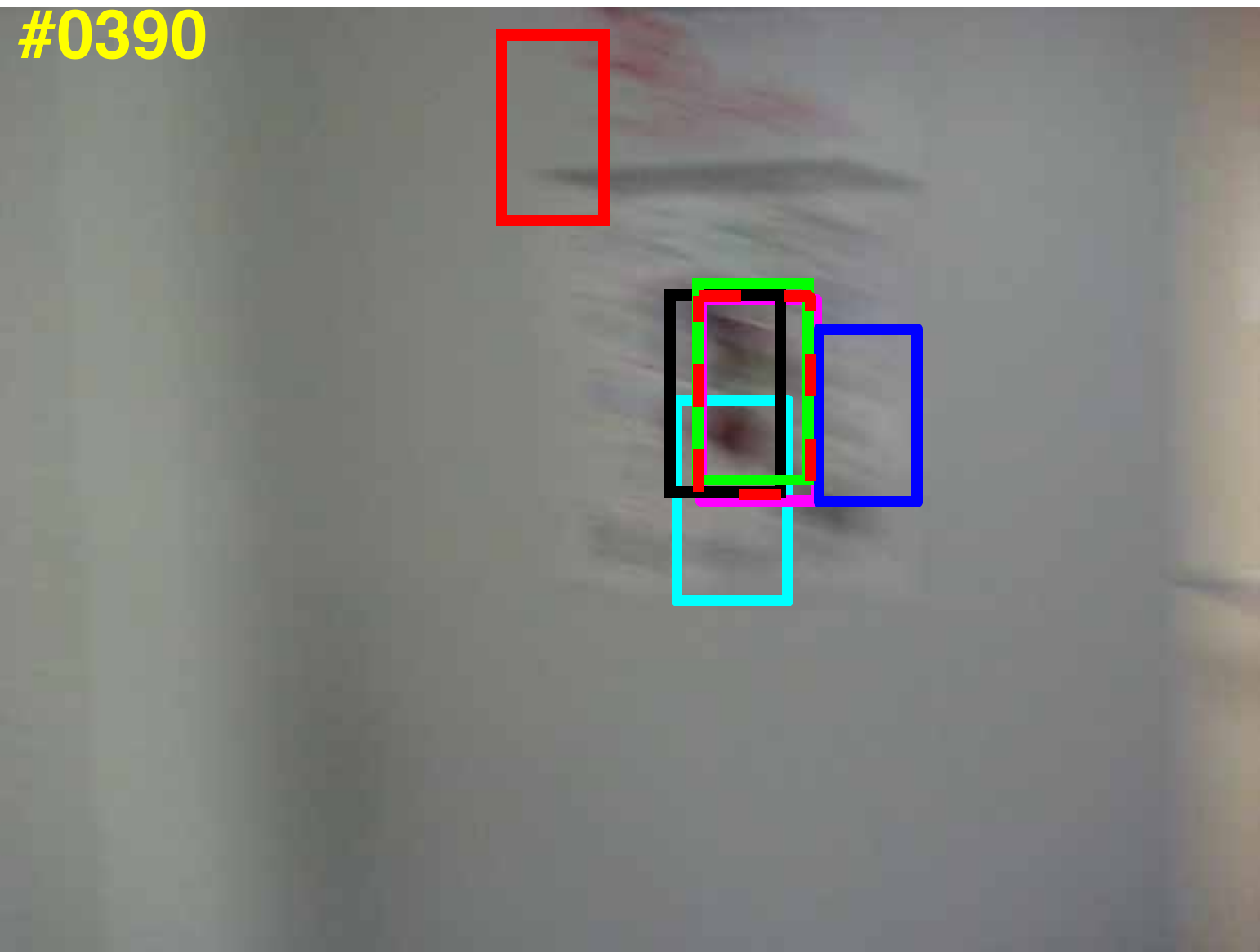}
\includegraphics[width=2.5cm,height=2cm]{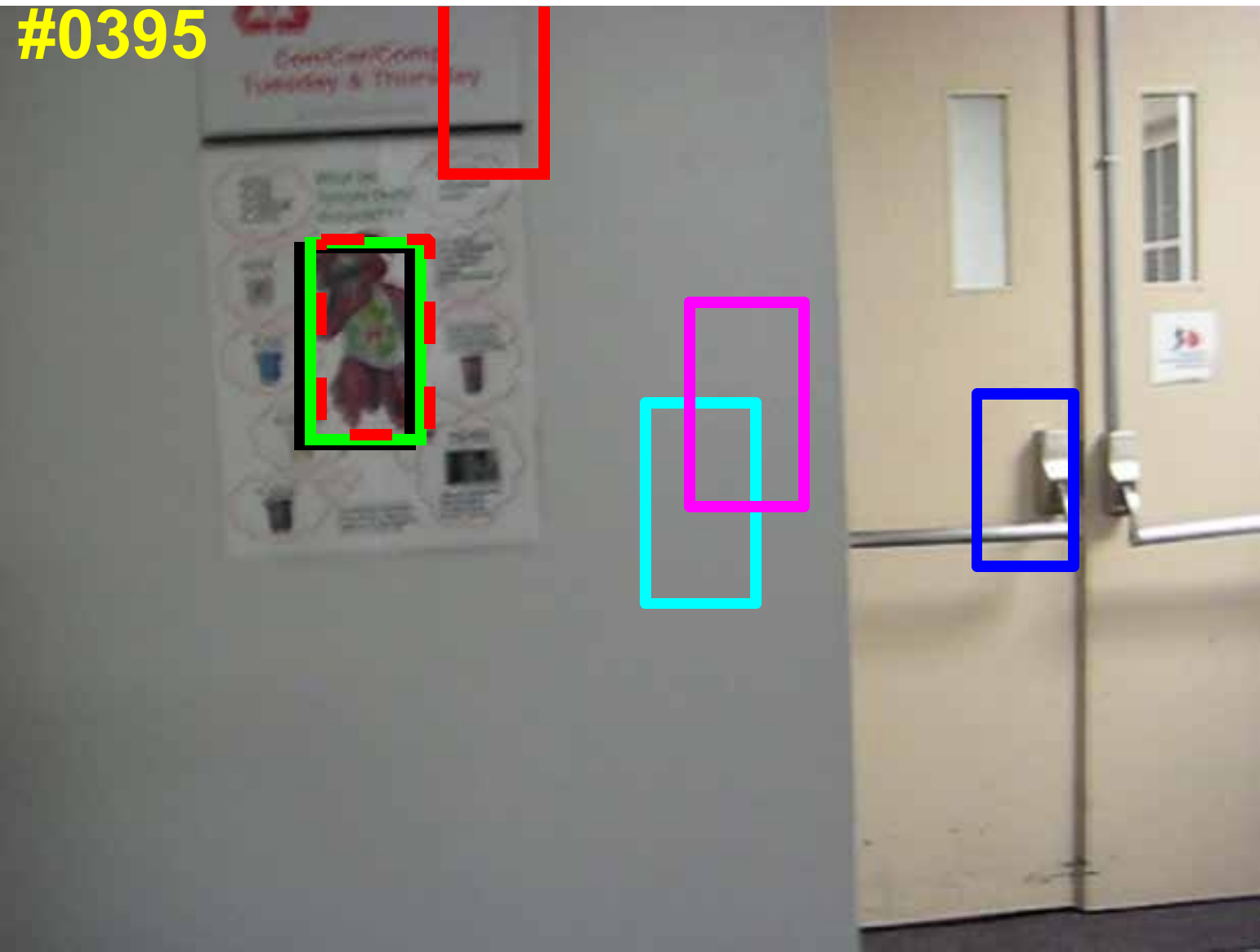}
\end{tabular}
}\\
\subfigure[Shot changes in \textit{Soccer}: there are gradual shot changes due to the changes of camera point-of-view  (see frame $45$ to frame $90$ and frame $335$ to frame $375$) or incurred by varying camera-subject distances (see frame $200$ to frame $230$ and finally to frame $305$).] {
\begin{tabular}{ccc}
\includegraphics[width=2.5cm,height=2cm]{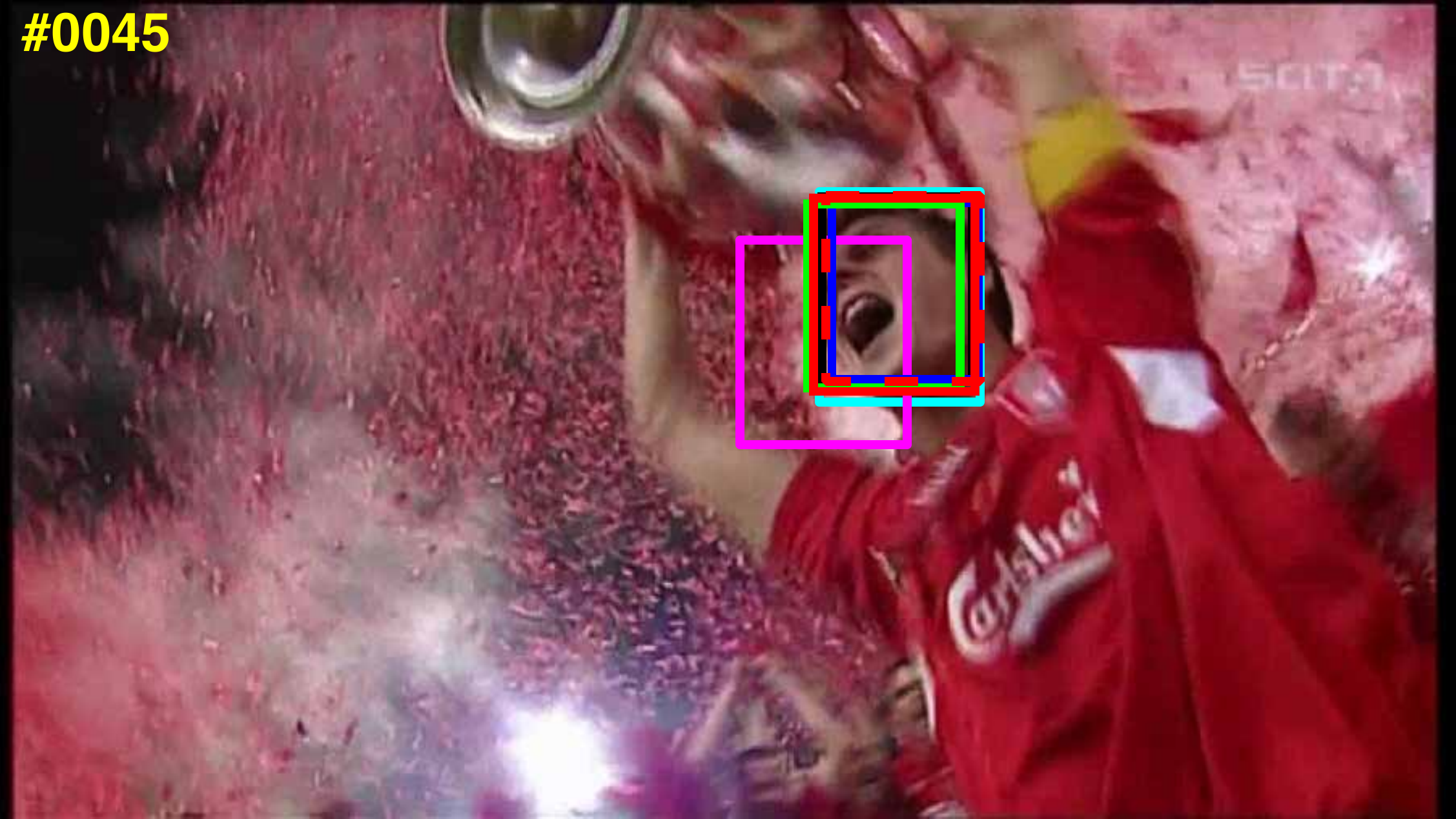}
\includegraphics[width=2.5cm,height=2cm]{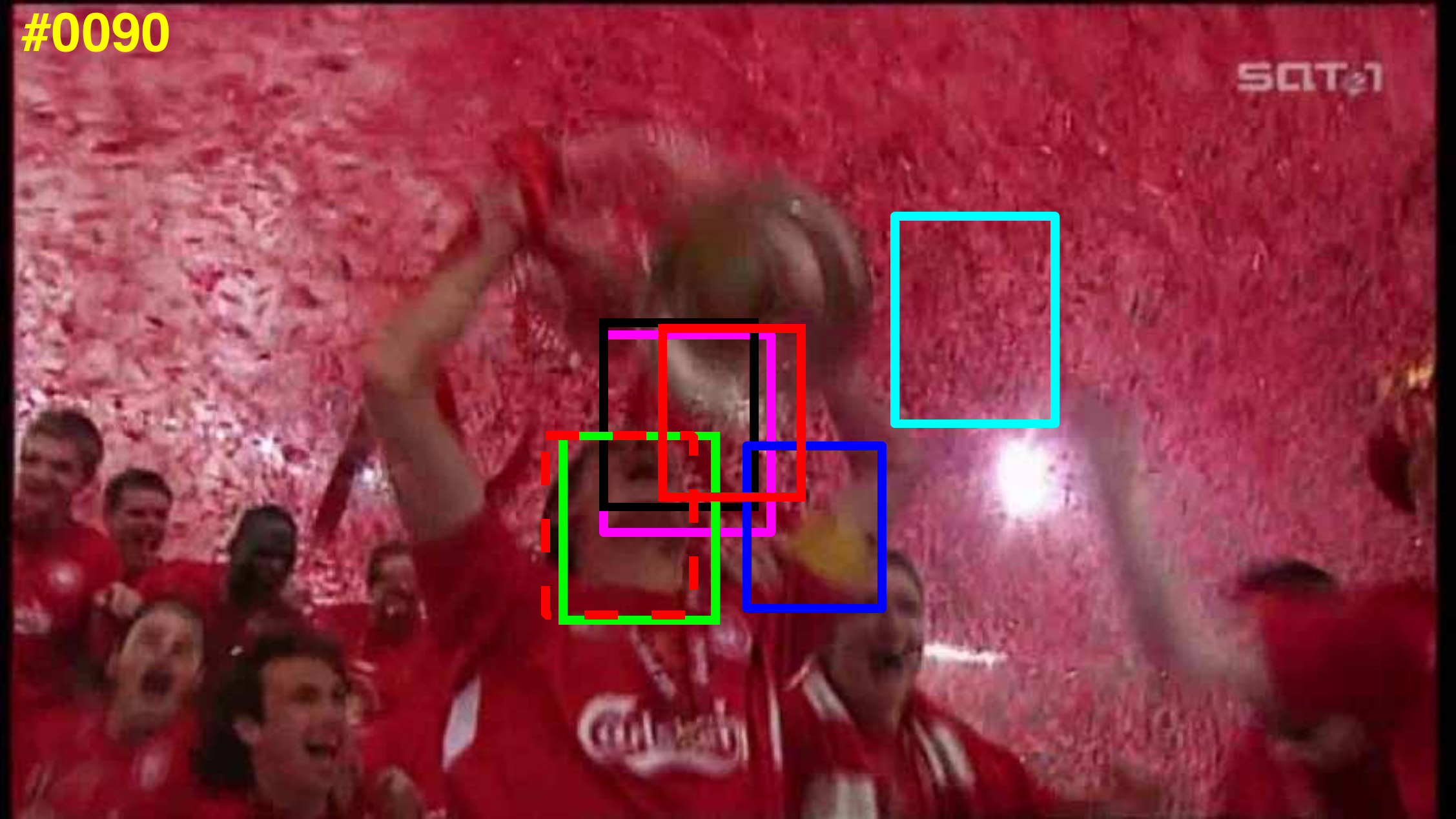}
\includegraphics[width=2.5cm,height=2cm]{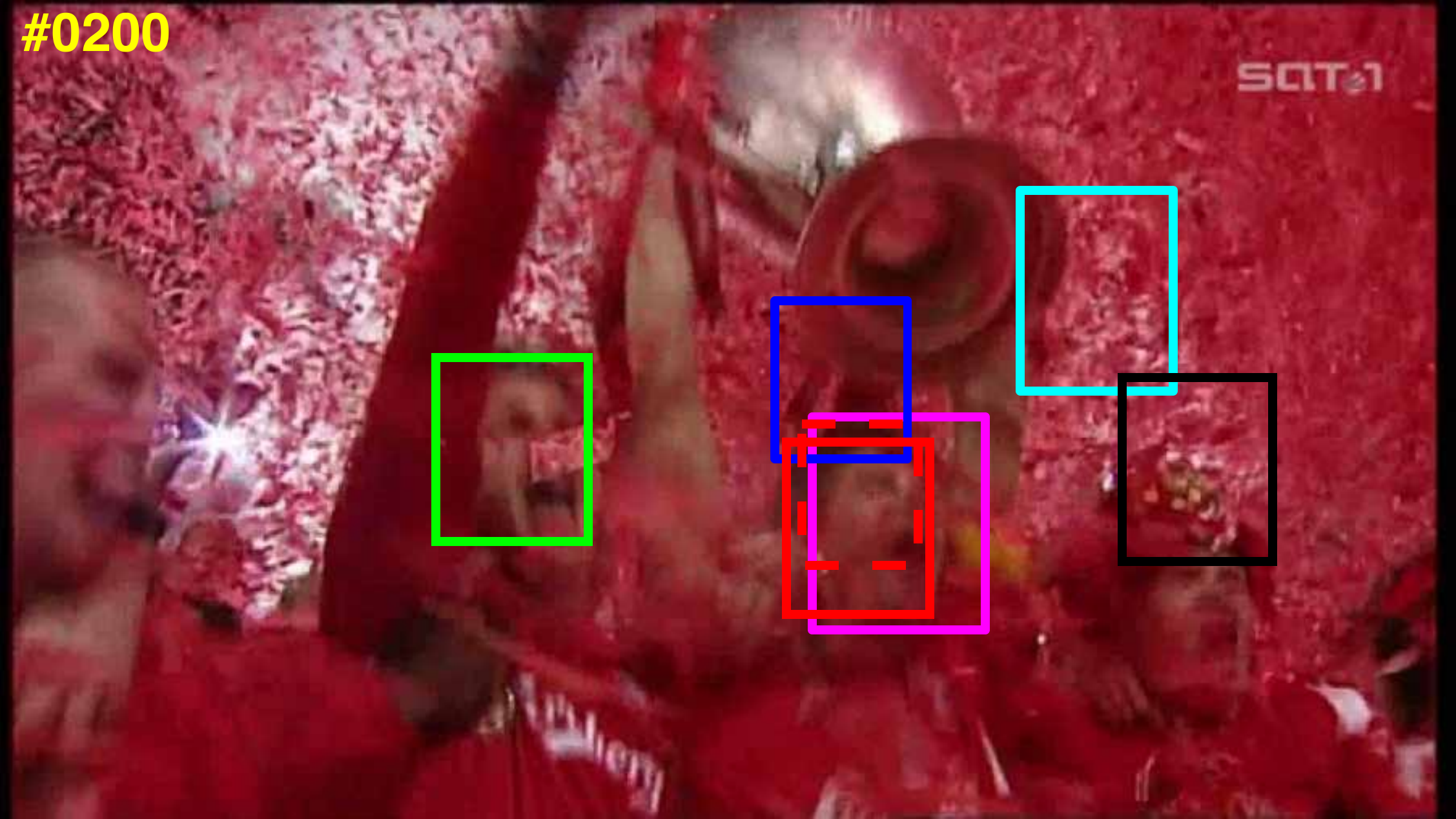}
\includegraphics[width=2.5cm,height=2cm]{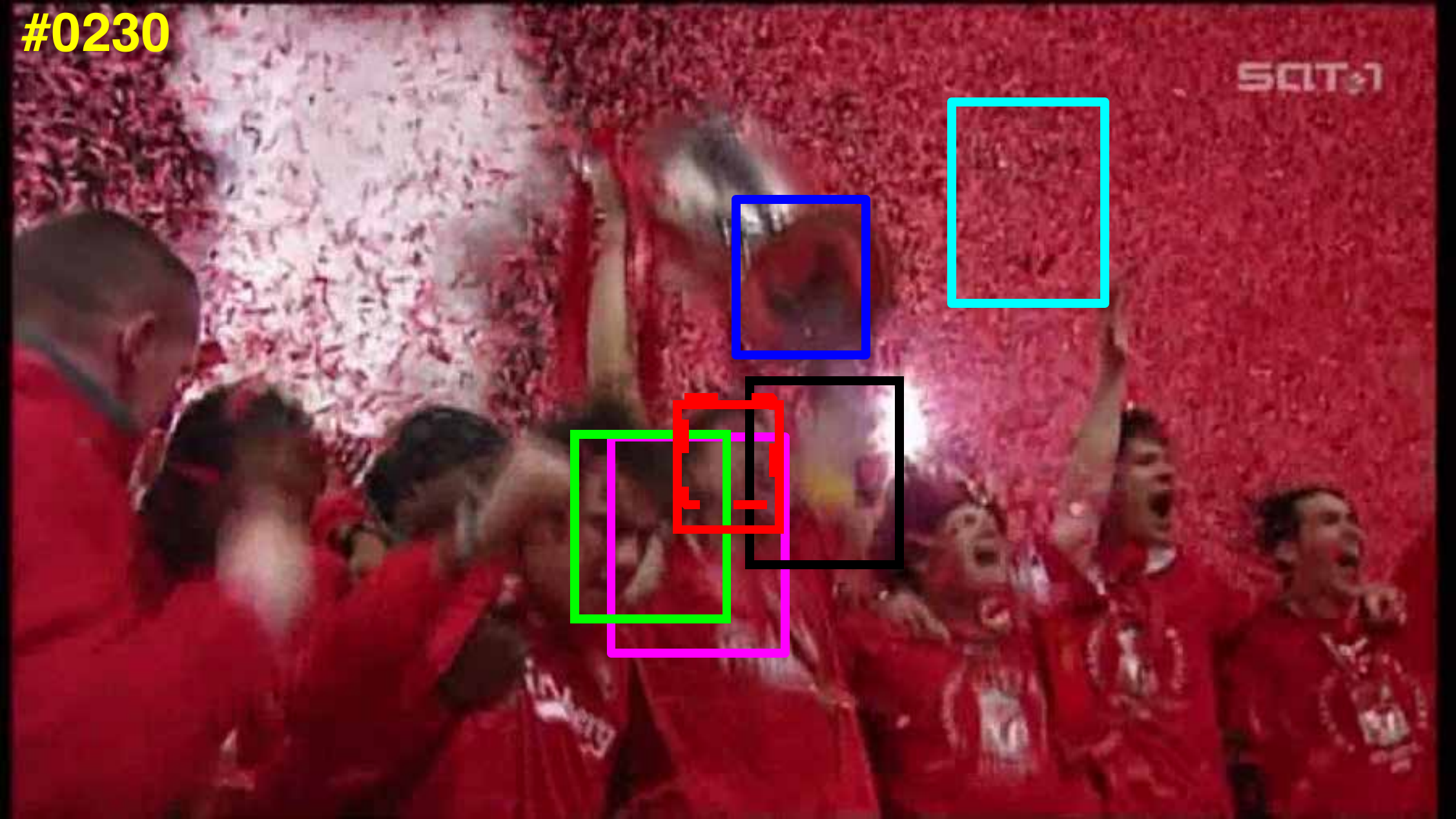}
\includegraphics[width=2.5cm,height=2cm]{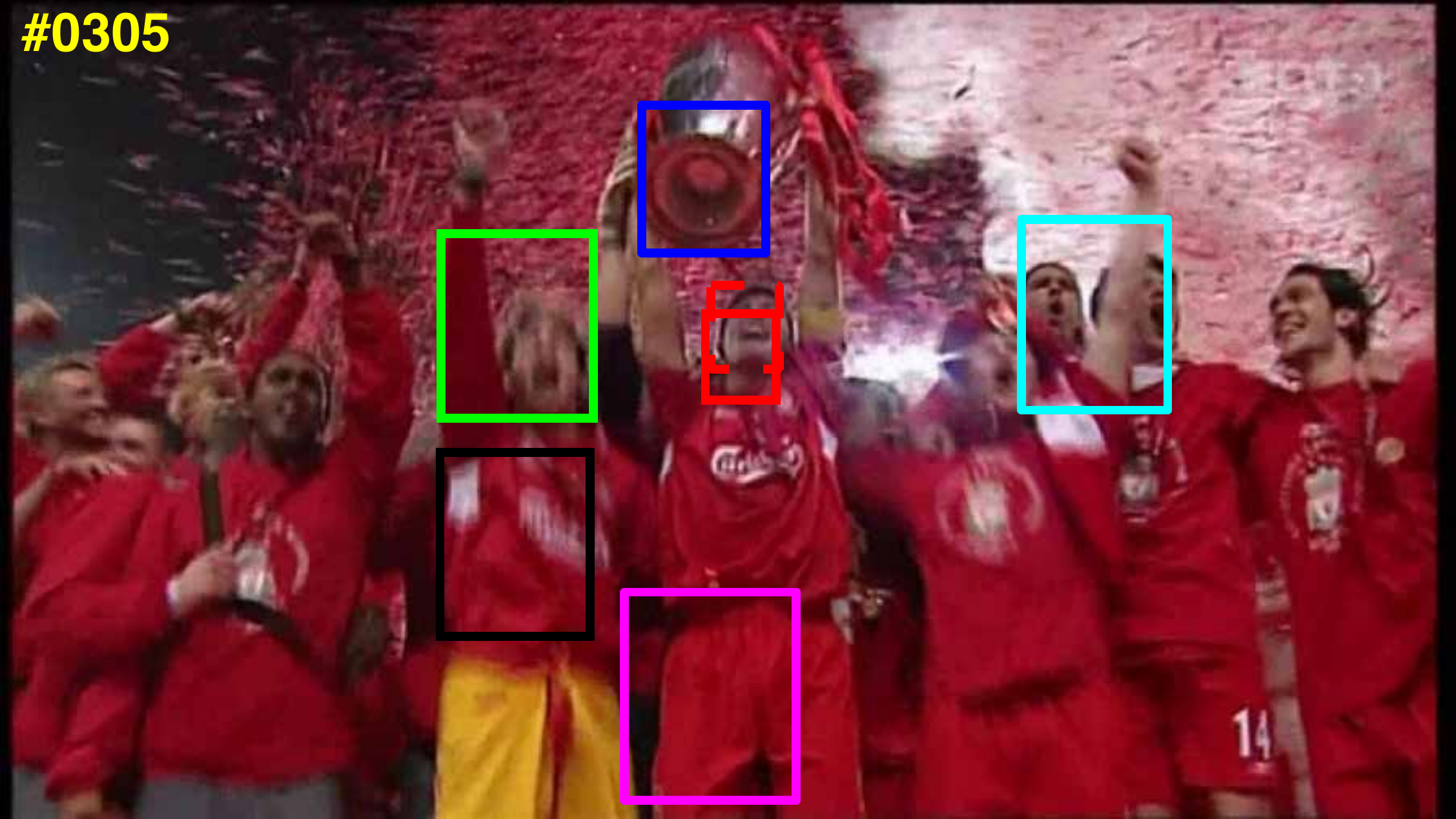}
\includegraphics[width=2.5cm,height=2cm]{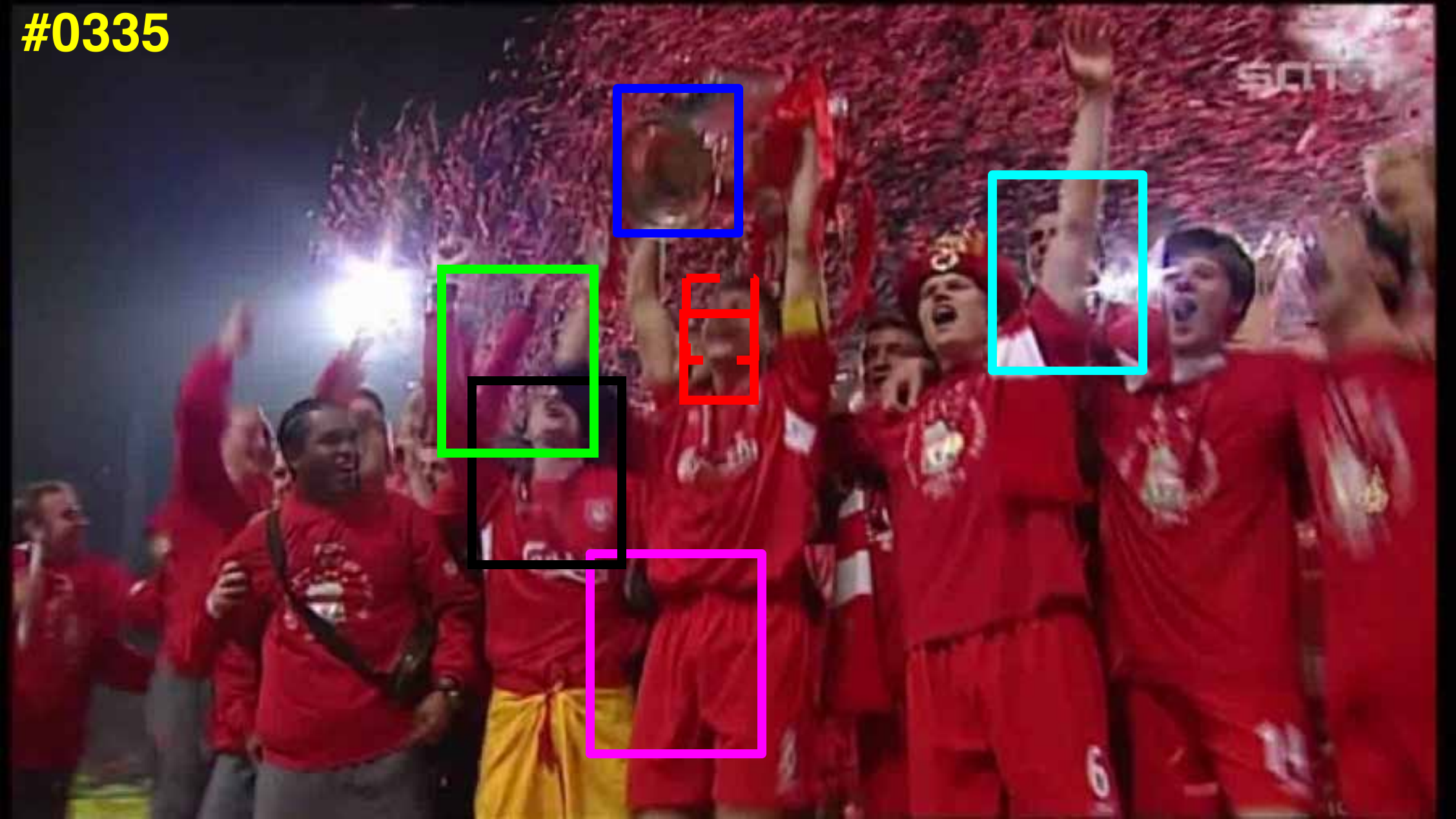}
\includegraphics[width=2.5cm,height=2cm]{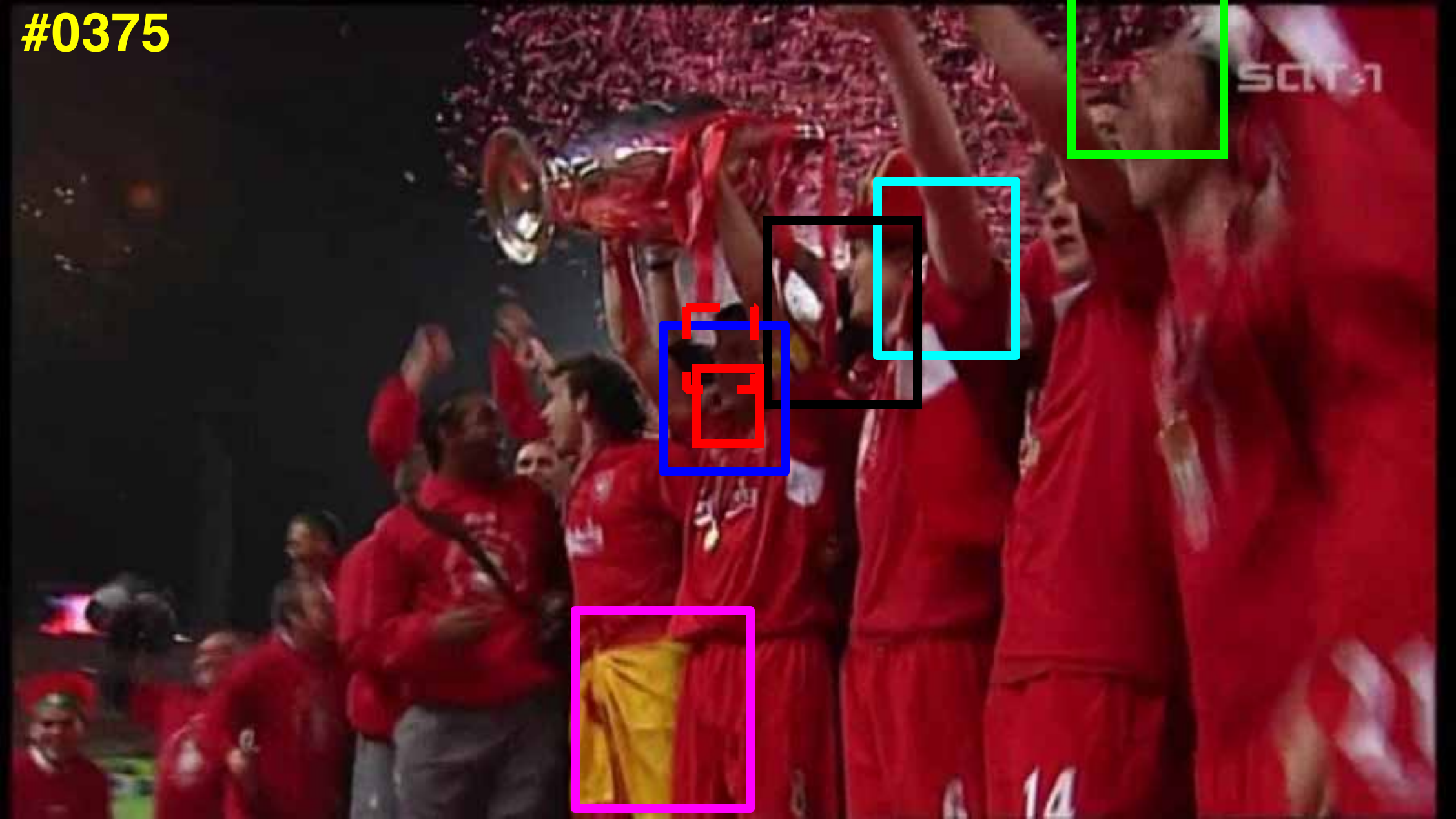}
\end{tabular}
}\\
\subfigure[Shot changes in \textit{Singer1}: there are shot changes due to the gradual changes of both camera point-of-view and camera-subject distances (see frame $2$ to frame $102$ and finally to frame $302$).] {
\begin{tabular}{ccc}
\includegraphics[width=2.5cm,height=2cm]{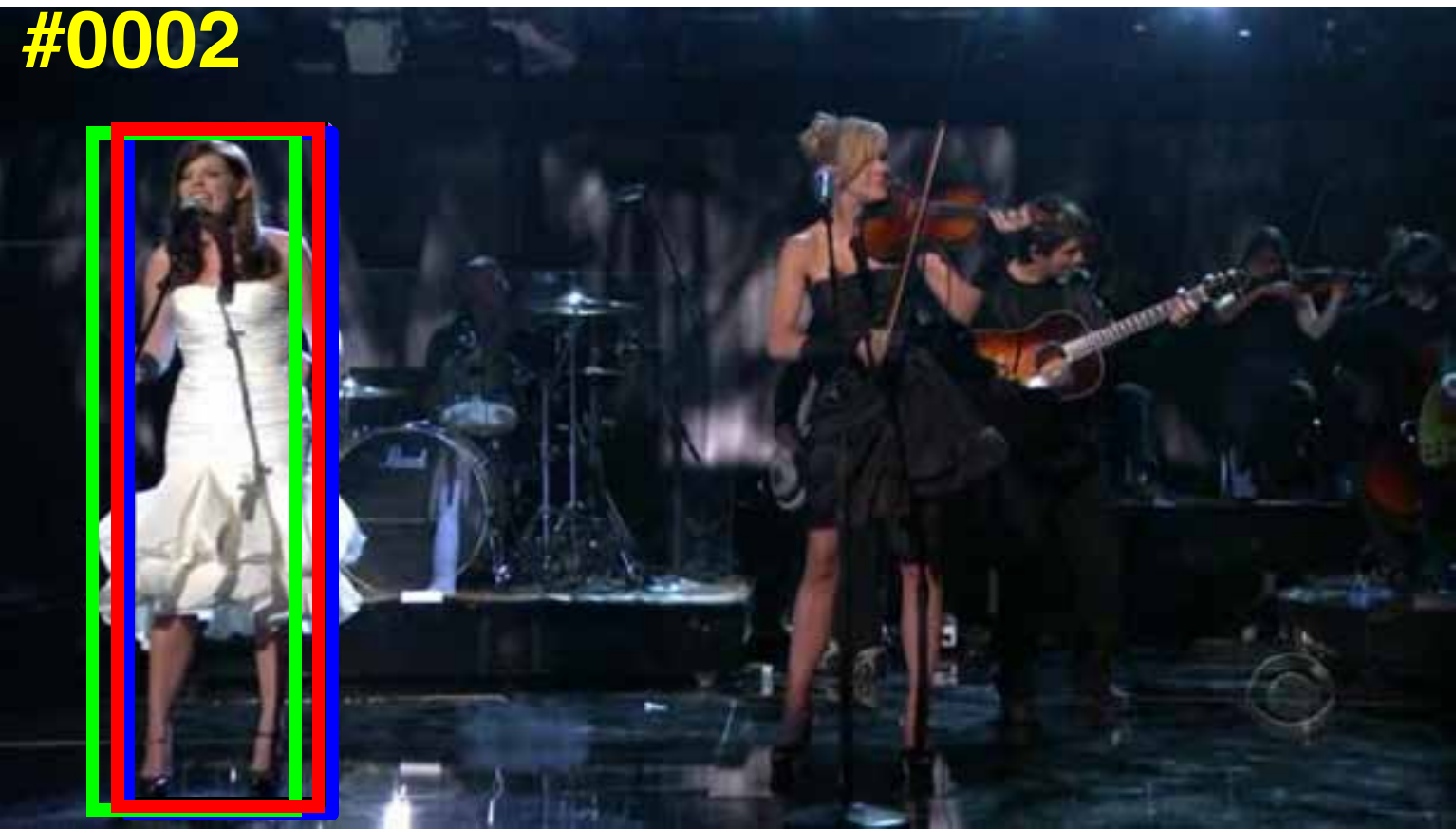}
\includegraphics[width=2.5cm,height=2cm]{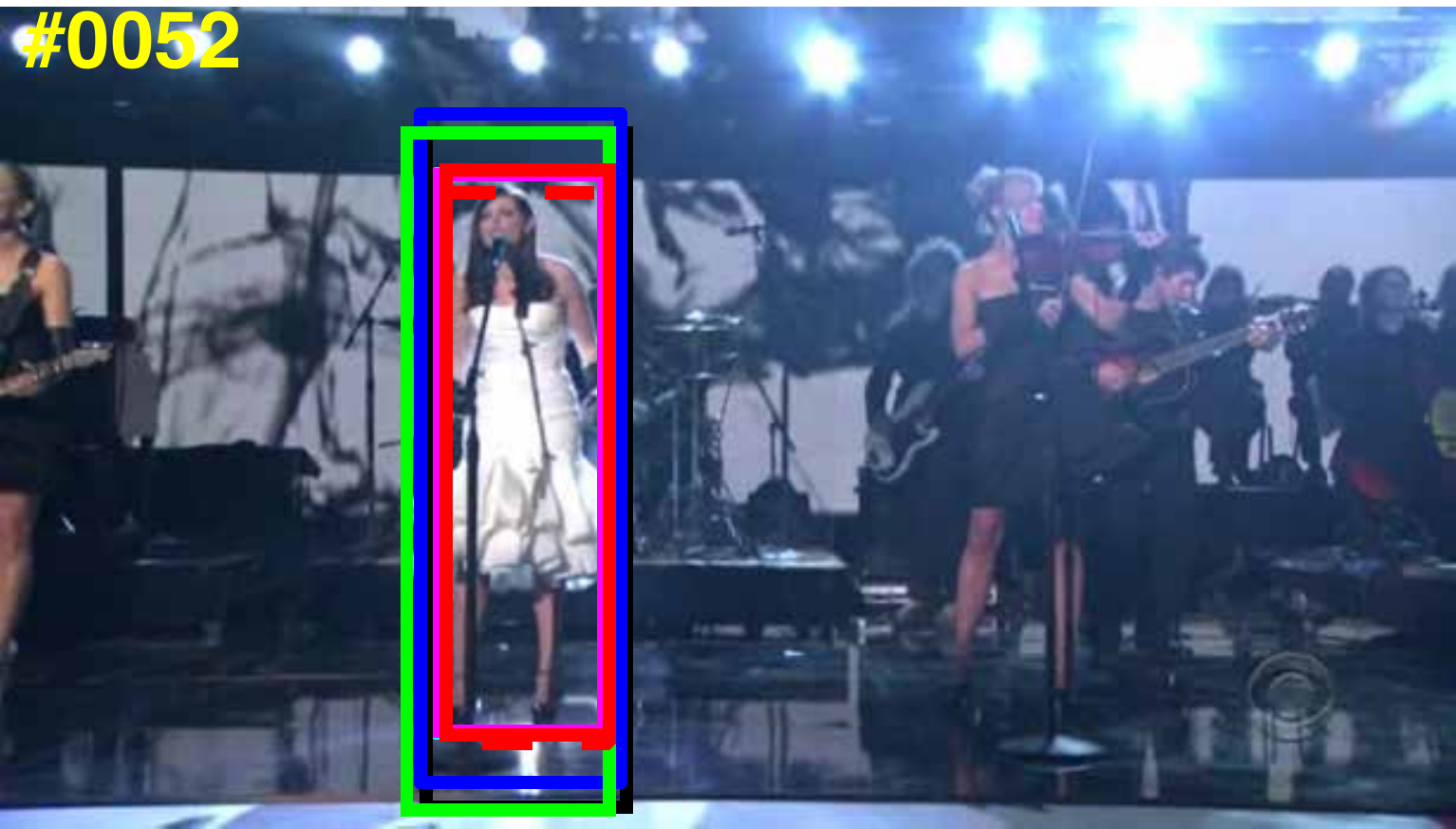}
\includegraphics[width=2.5cm,height=2cm]{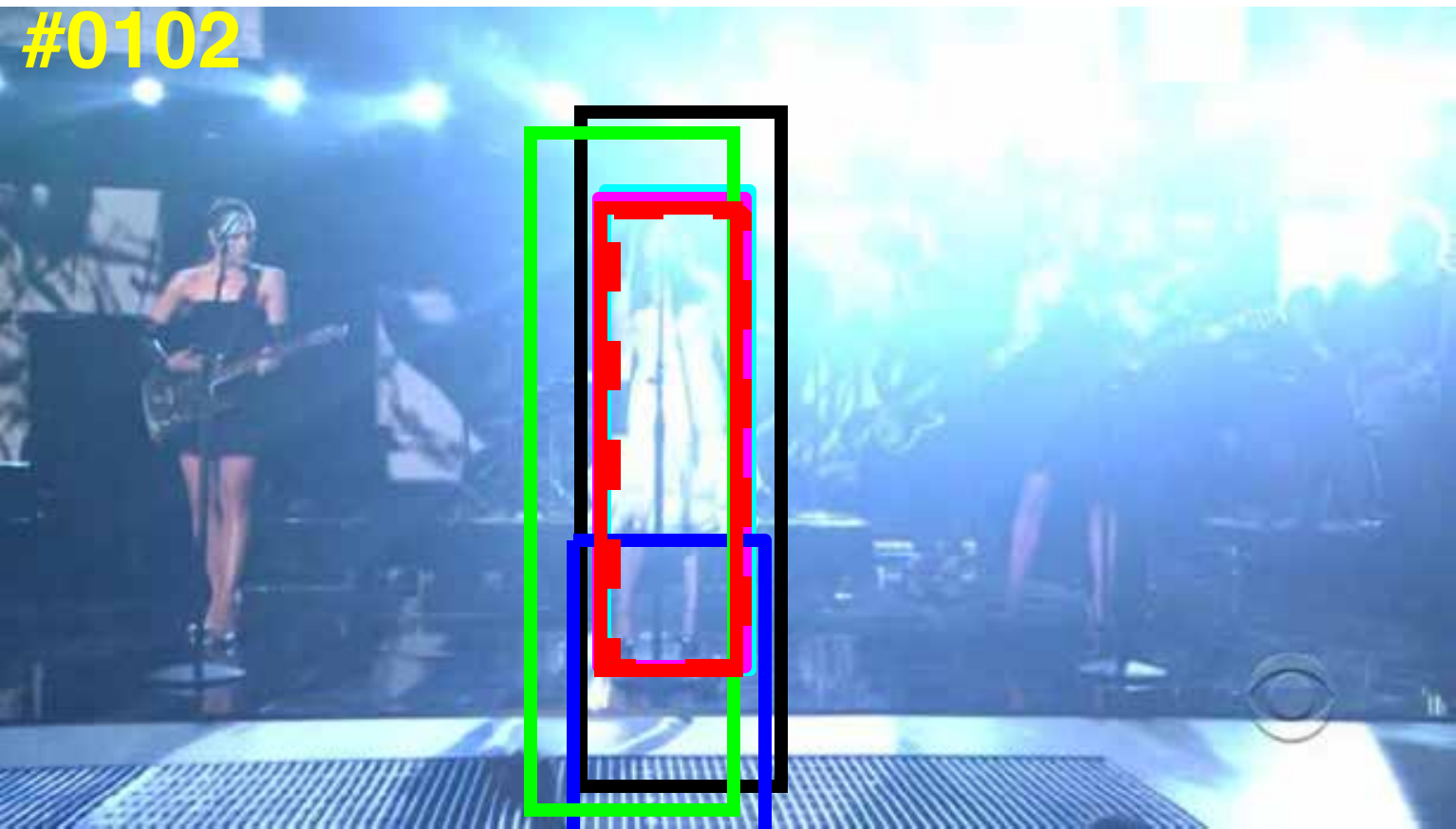}
\includegraphics[width=2.5cm,height=2cm]{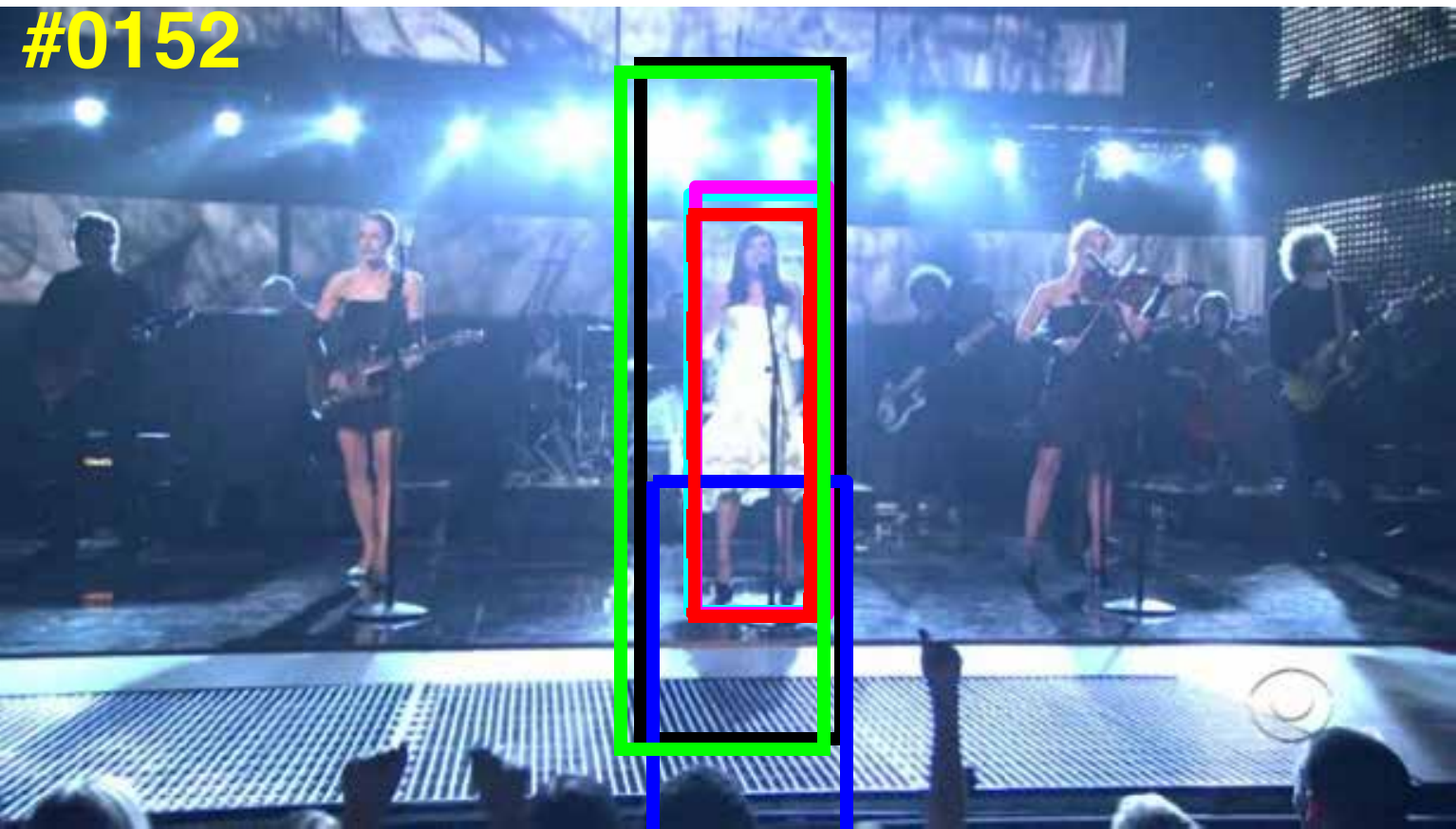}
\includegraphics[width=2.5cm,height=2cm]{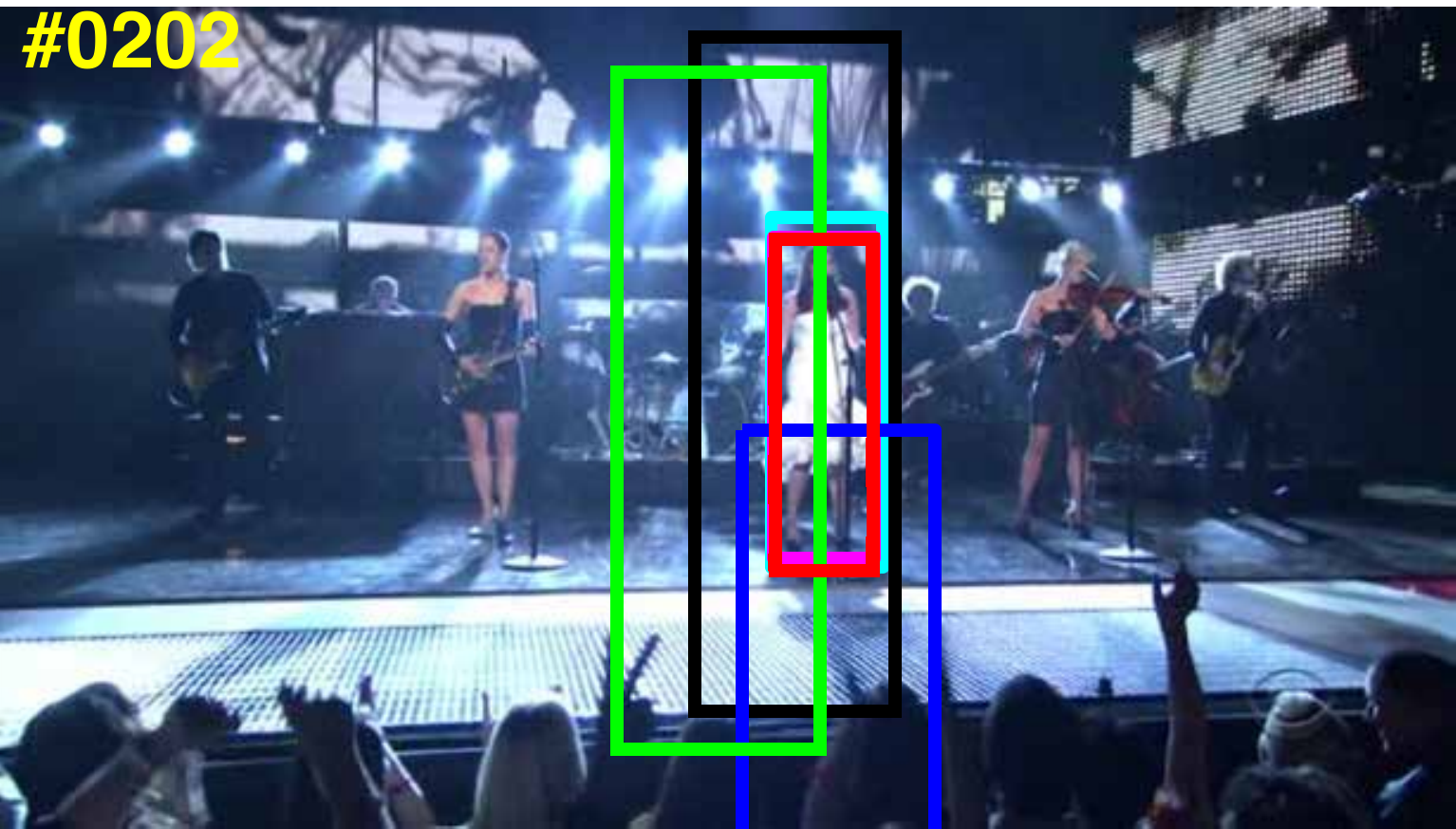}
\includegraphics[width=2.5cm,height=2cm]{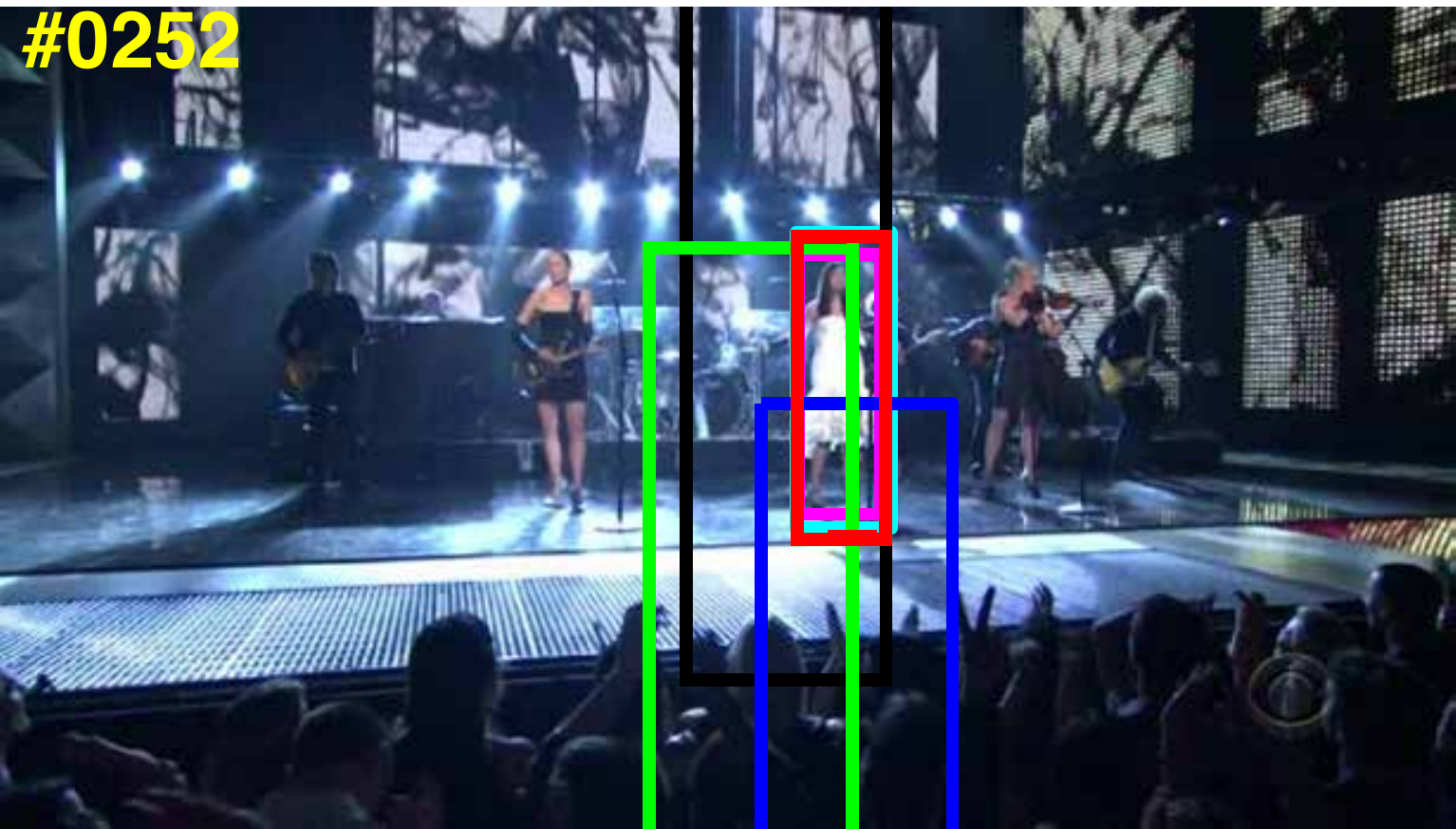}
\includegraphics[width=2.5cm,height=2cm]{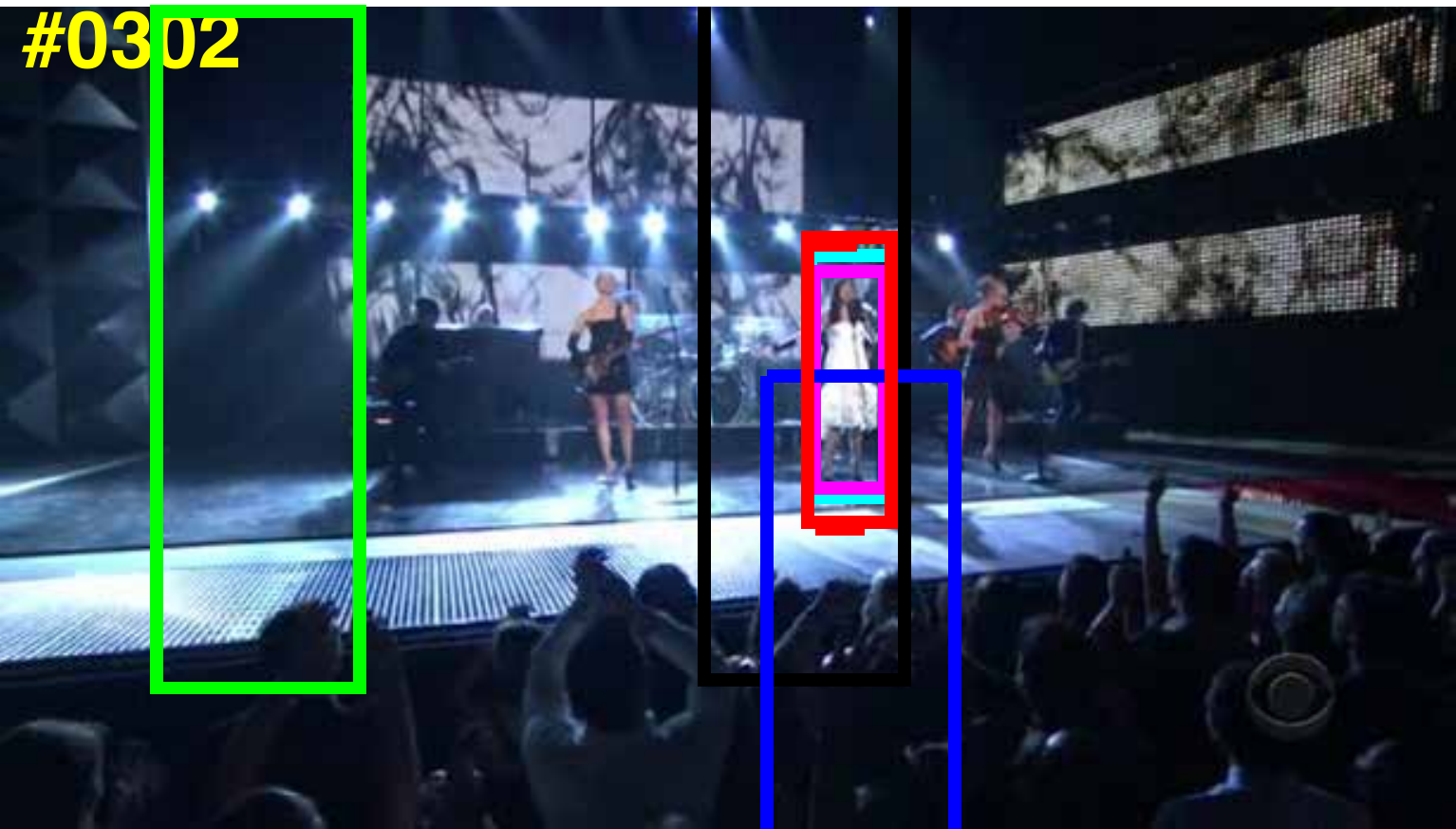}
\end{tabular}
}\\
\subfigure[Shot changes in \textit{Singer3}: there are shot changes due to the gradual (and rapid) changes of both camera point-of-view and camera-subject distances (e.g., frame $26$ to frame $36$). Our modification $\text{K}(\text{MF})^2\text{JMT}$-$\text{M}1$ may underestimate the target size due to the rapid changes, but it still provides the most accurate estimation among others.] {
\begin{tabular}{ccc}
\includegraphics[width=2.5cm,height=2cm]{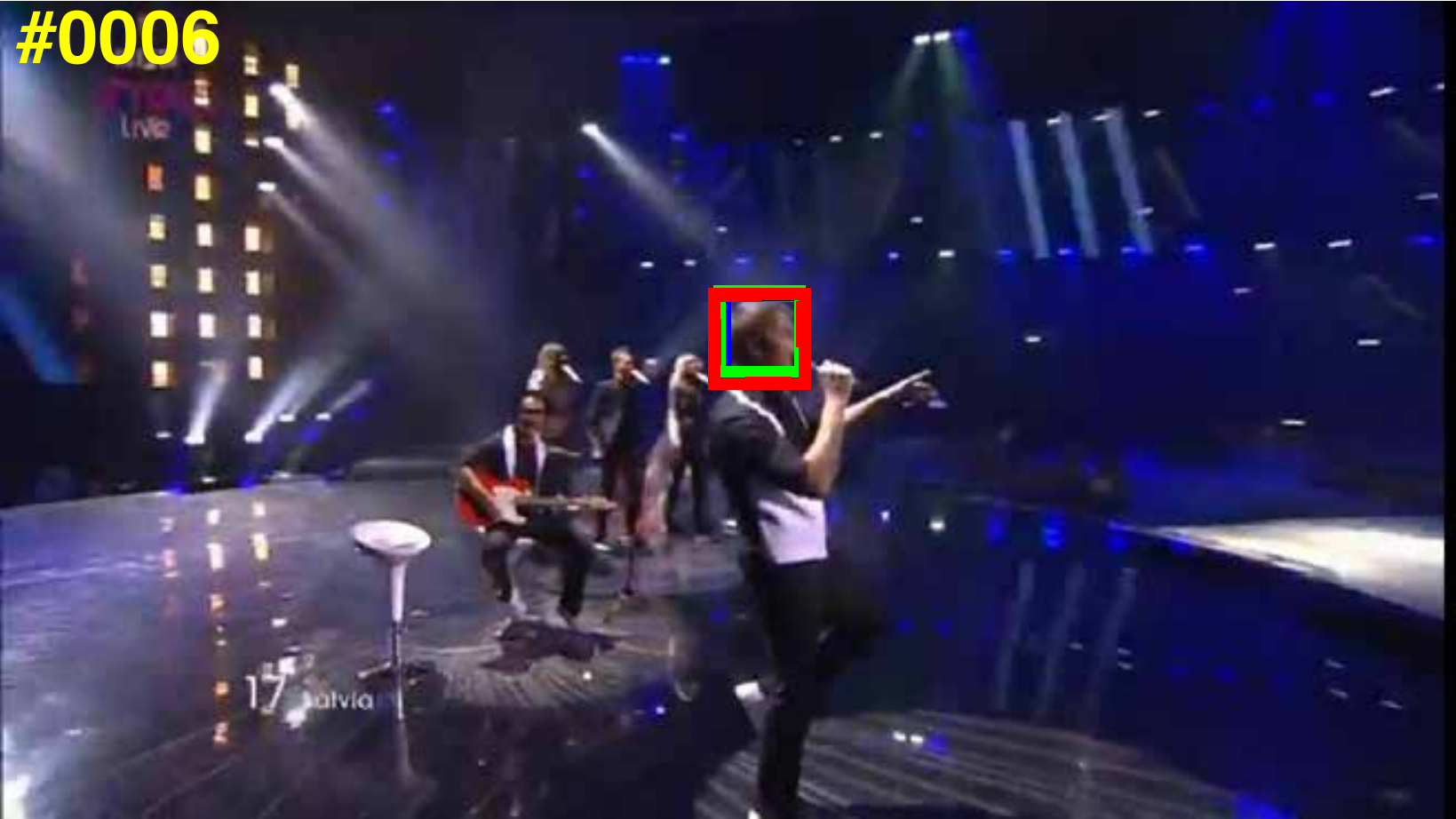}
\includegraphics[width=2.5cm,height=2cm]{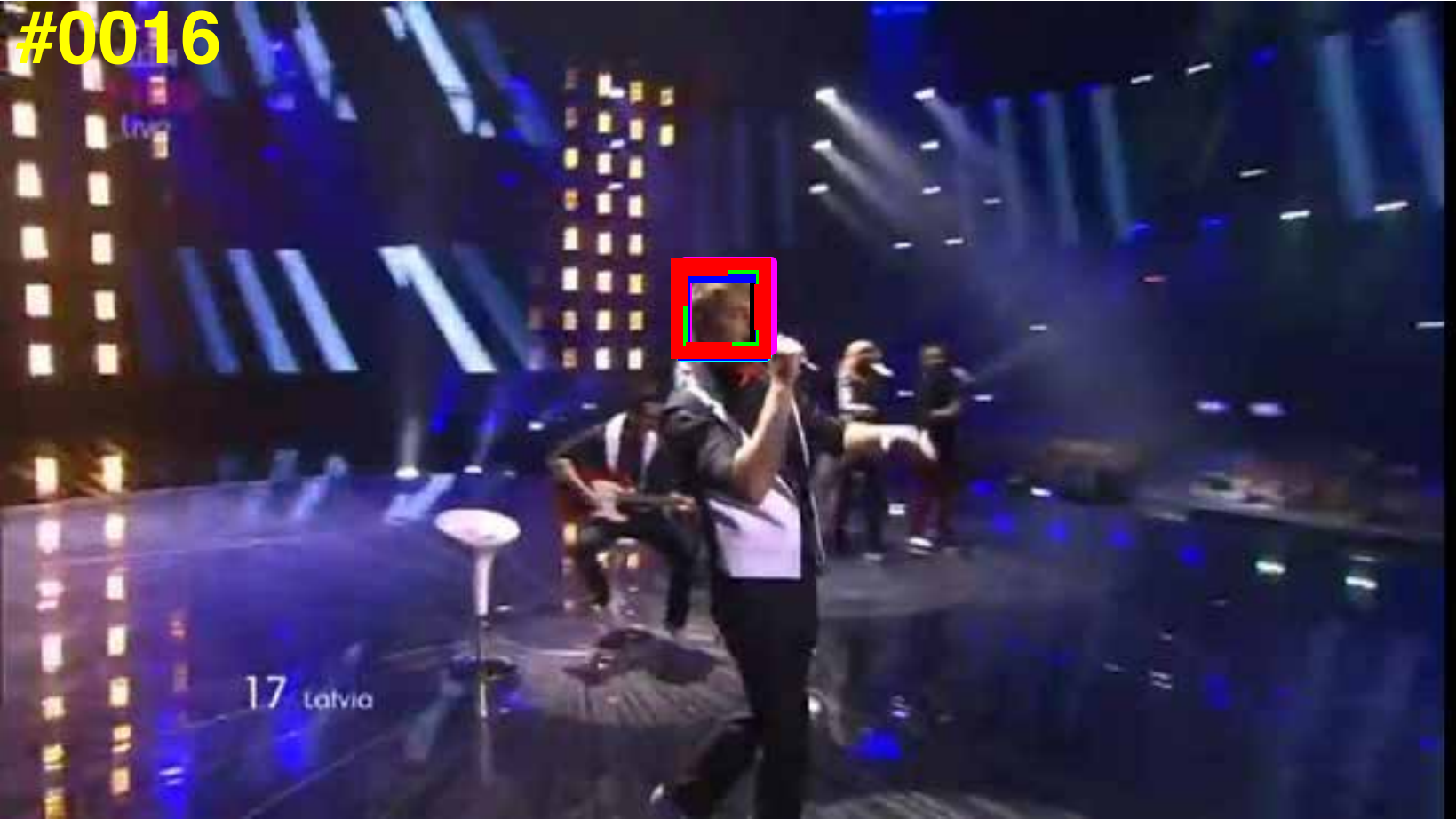}
\includegraphics[width=2.5cm,height=2cm]{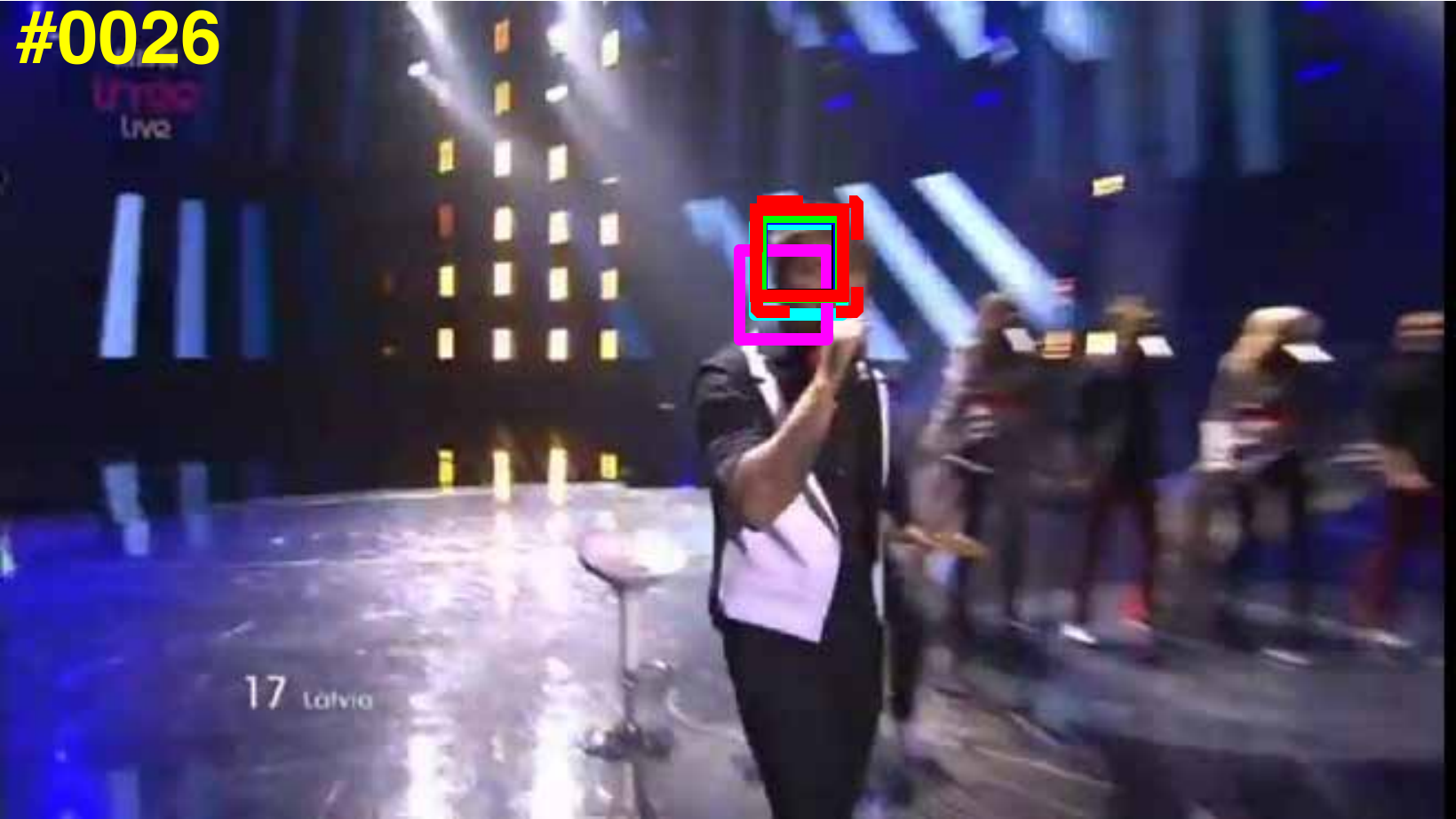}
\includegraphics[width=2.5cm,height=2cm]{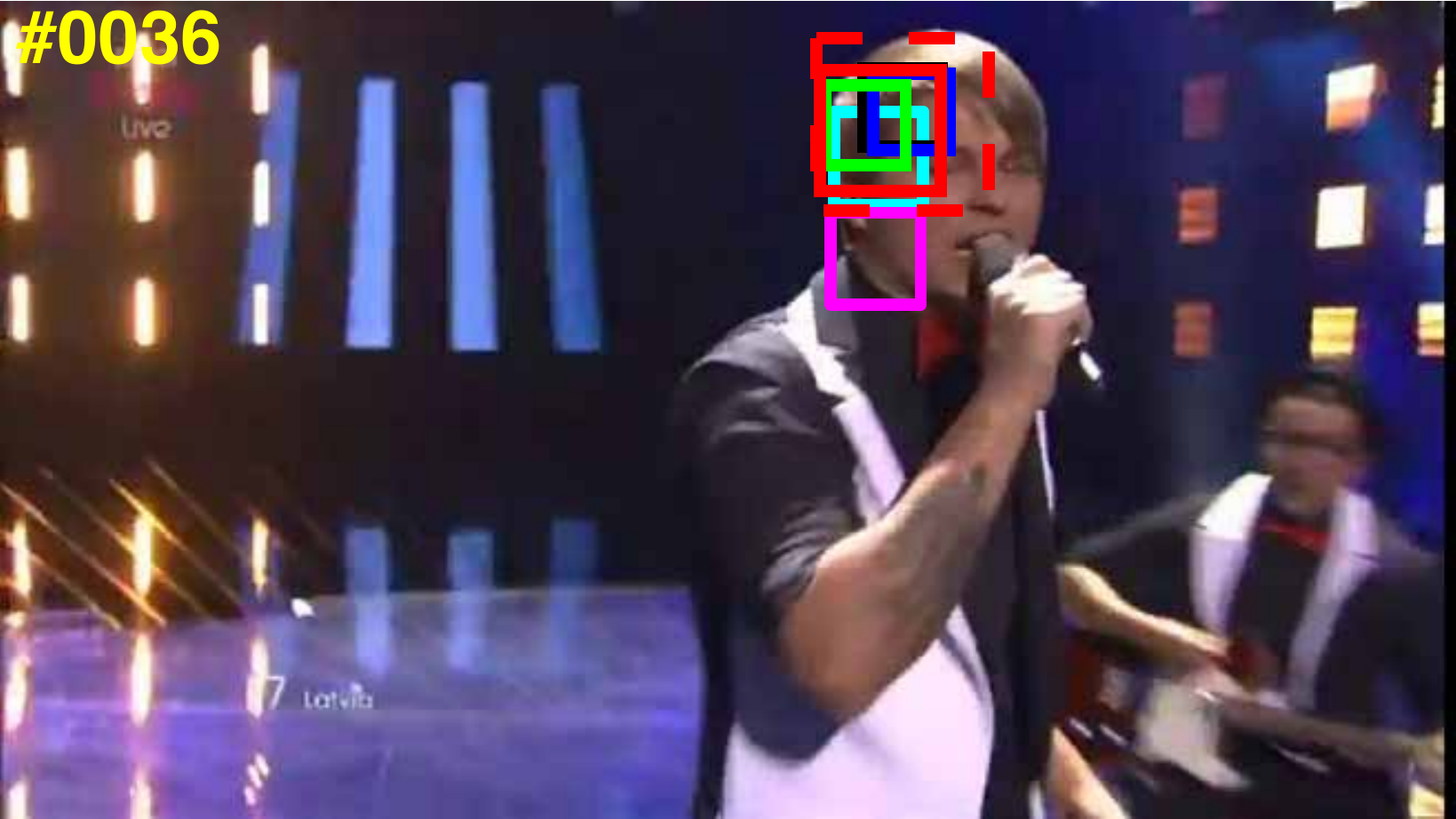}
\includegraphics[width=2.5cm,height=2cm]{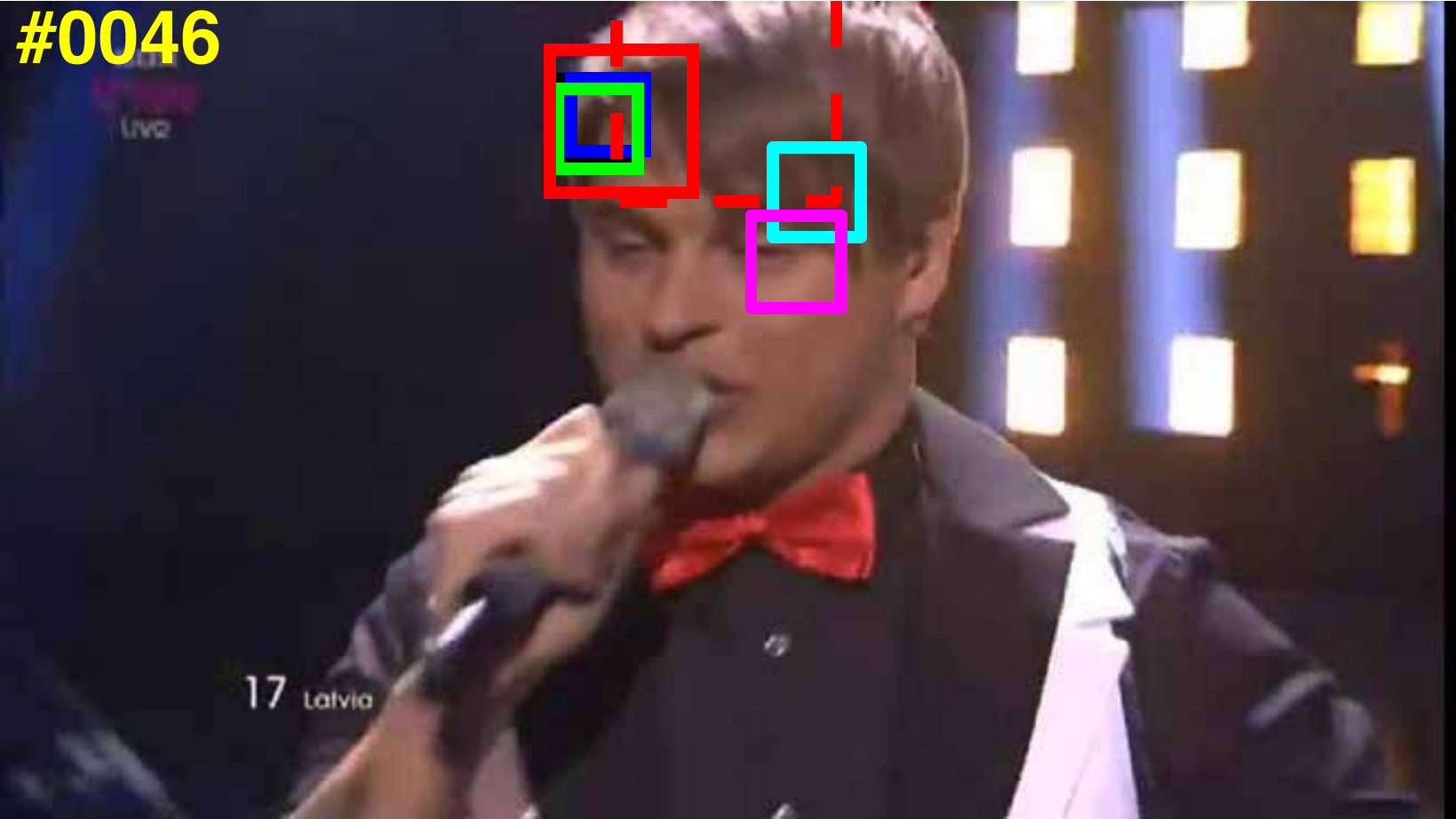}
\includegraphics[width=2.5cm,height=2cm]{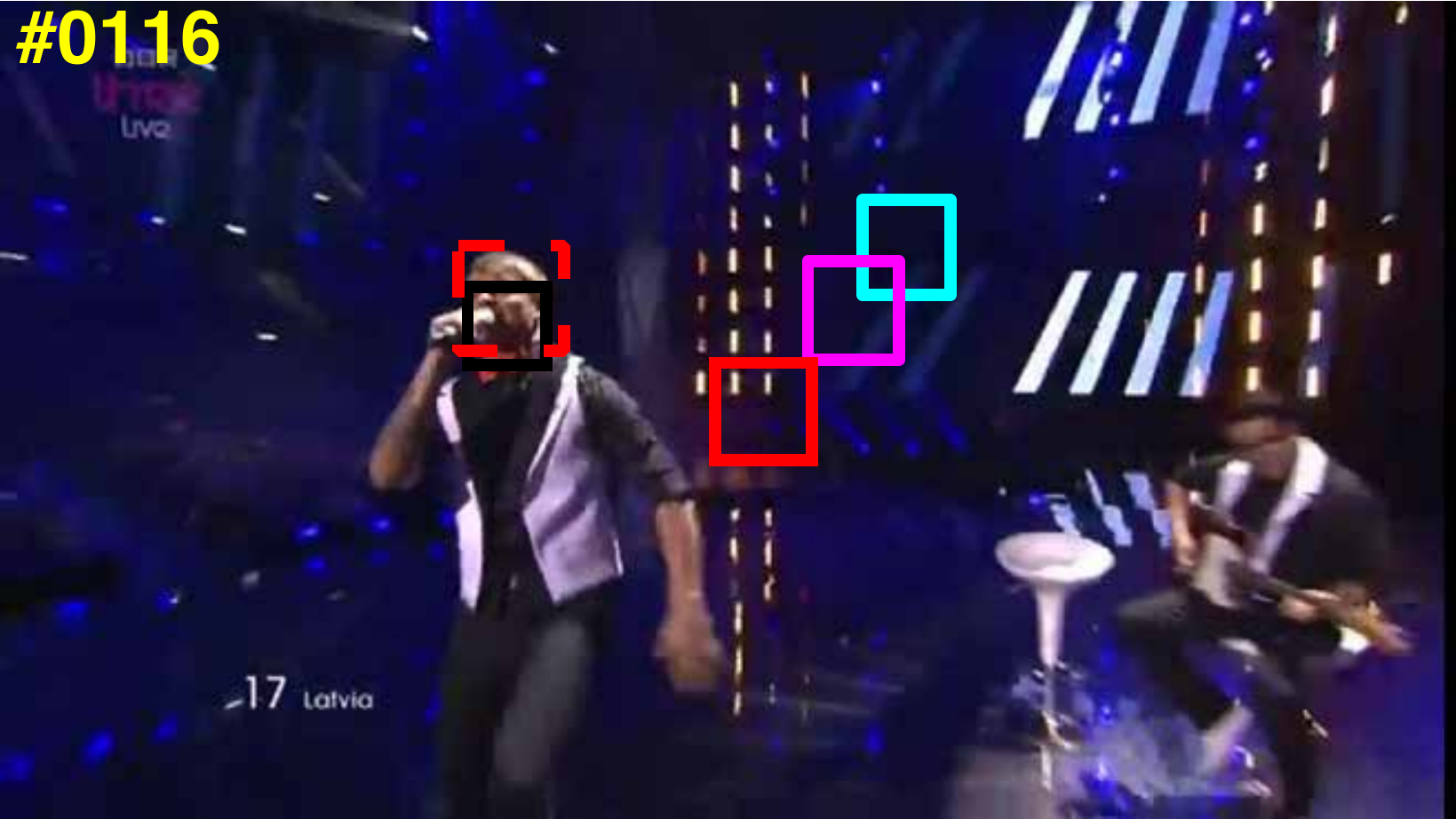}
\includegraphics[width=2.5cm,height=2cm]{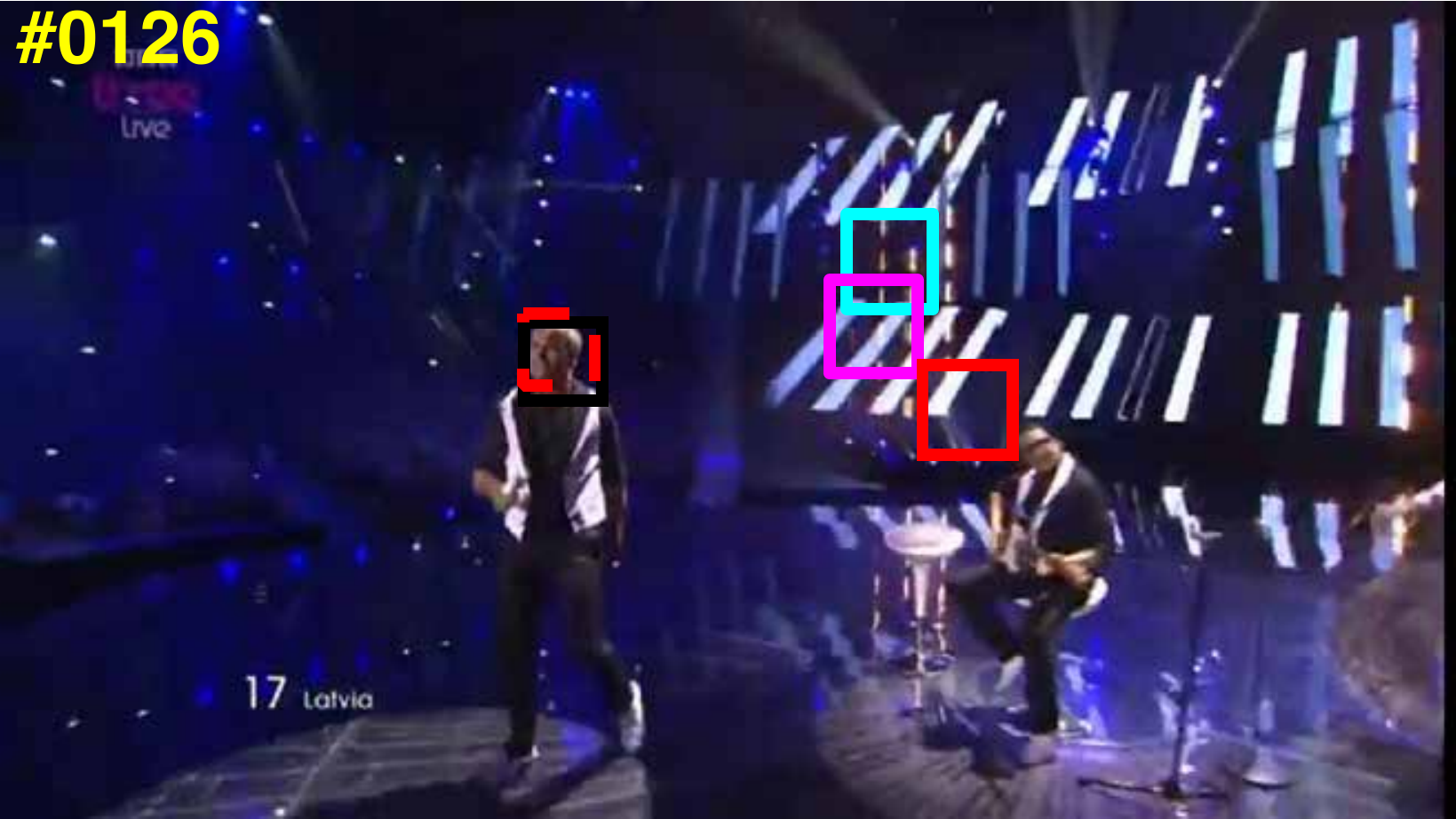}
\end{tabular}
}\\
\subfigure[Tracker legend] {\includegraphics[width=0.48\textwidth]{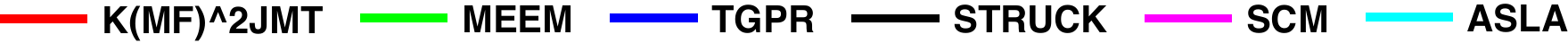}}
\caption{A qualitative comparison of our method and its modification with five state-of-the-art trackers. Tracking results are shown on five videos contain scene cuts or shot transitions. \textit{DragonBaby}, \textit{BlurOwl} and \textit{Soccer} are from OTB $2015$, whereas \textit{Singer1} and \textit{Singer3} are from VOT$2015$ benchmark. The basic $\text{K}(\text{MF})^2\text{JMT}$ performs favorably in these videos. Our modification $\text{K}(\text{MF})^2\text{JMT}$-$\text{M}1$ offers the best performance. (f) shows tracker legend.}
\label{visual_comparison_scene_cuts}
\end{figure*}

We implement the first modification to validate its effectiveness due to its simplicity. Specifically, given $M$ training frames in the overall objective, the weight in the current frame is $A_0$, then the weights in previous frames are decayed inversely proportional to the square of the distance from the current frame (i.e., the weight in the most adjacent frame is $A_0/4$, the weight in the second most adjacent frame is $A_0/9$, and the weight in the farthest frame is $A_0/M^2$). We term this modification $\text{K}(\text{MF})^2\text{JMT}$-$\text{M}1$ and set $A_0=5$ in the following proof-of-concept experiment\footnote{The parameter $A_0$ is selected, from the range $[1,10]$ with interval $1$, as the one that achieves the highest mean success rate among all selected video sequences with scene cuts.}. Fig.~\ref{visual_comparison_scene_cuts} plots the tracking results of our $\text{K}(\text{MF})^2\text{JMT}$ and $\text{K}(\text{MF})^2\text{JMT}$-$\text{M}1$ as well as their five competitors. Table~\ref{lab:modification_improvement} summarizes the overlap precision ($\%$) at threshold $0.5$ for all competing trackers.

As can be seen, our basic $\text{K}(\text{MF})^2\text{JMT}$ performs favorably in these videos, but it may miss the target or overestimate the target size due to unconstrained shot changes. The simple modification can effectively alleviate the negative effects incurred by these changes, thus further improving the performance of $\text{K}(\text{MF})^2\text{JMT}$. This result suggests that the precise utilization of temporal information (coupled with a careful weighting strategy) is preferred in (unconstrained) videos containing shot changes or scene cuts. At the same time, it also suggests the (possible) existence of the room for performance improvement with an advanced strategy to address shot changes. We leave the implementation of the second modification as future work.

\begin{table}[!htbp]
\centering
\caption{A comparison of $\text{K}(\text{MF})^2\text{JMT}$ and $\text{K}(\text{MF})^2\text{JMT}$-$\text{M}1$ with five state-of-the-art trackers. For each tracker, the overlap precision ($\%$) at threshold $0.5$ is presented. The best two results are marked with red and blue respectively.}\label{lab:modification_improvement}
\begin{tabular}{cccccccc}\hline
 & MEEM & TGPR & STRUCK & SCM & ASLA & $\text{K}(\text{MF})^2\text{JMT}$ & $\text{K}(\text{MF})^2\text{JMT}$-$\text{M}1$ \\\hline
$\text{DragonBaby}$ & 65.5 & \color{red}{73.5} & 8.8 & 23.0 & 15.0 & 46.0 & \color{blue}{66.4} \\
$\text{BlurOwl}$ & \color{red}{98.6} & 51.2 & \color{red}{98.6} & 21.6 & 17.6 & 55.9 & 90.2 \\
$\text{Soccer}$ & 36.0 & 13.0 & 15.6 & 23.7 & 12.5 & \color{blue}{56.6} & \color{red}{78.6} \\
$\text{Singer1}$ & 25.1 & 22.8 & 29.9 & \color{red}{100} & \color{red}{100} & 93.7 & 98.6 \\
$\text{Singer3}$ & 15.3 & 15.3 & 24.4 & 15.3 & 16.0 & \color{blue}{17.6} & \color{red}{37.4} \\\hline
$\text{Mean}$ & 48.1 & 35.2 & 35.5 & 36.7 & 32.2 & \color{blue}{54.0} & \color{red}{74.2} \\\hline
\end{tabular}
\end{table}

\end{document}